\documentclass[]{fairmeta}

\usepackage{longtable}
\usepackage{amsmath,amssymb}
\usepackage{comment}
\usepackage{graphicx,booktabs,array,multirow,makecell,needspace,placeins,tcolorbox,adjustbox}
\usepackage{tikz}

\usetikzlibrary{arrows.meta,shapes.geometric}

\newtcolorbox{promptbox}[1]
{breakable,colback=black!2,colframe=black!55,colbacktitle=black!8,coltitle=black,boxrule=0.5pt,arc=1mm,left=6pt,right=6pt,top=5pt,bottom=5pt,fonttitle=\small\bfseries,fontupper=\small,title={#1},before skip=8pt,after skip=8pt}
\graphicspath{{figures/}}

\newcommand{\ourname}{SII-Decision}
\newcommand{\oursmall}{\ourname-4B}
\newcommand{\ourlarge}{\ourname-27B}
\newcommand{\ourmethod}{Reasoning-to-Readout Self-Distillation}

\usepackage[most]{tcolorbox}
\newtcolorbox[auto counter]{finding}[1][]{
  enhanced, breakable,
  colback=blue!3, colframe=blue!35!black, coltitle=black, boxrule=0.4pt, arc=2pt,
  left=6pt, right=6pt, top=4pt, bottom=4pt,
  fontupper=\small,
  title={Finding~\thetcbcounter}, detach title,
  before upper={\textbf{\tcbtitle.}\enspace},
  #1}

\title{General Decision Models: Benchmarking and Insights Beyond Jev}

\author[1,2,*]{Feiyu Duan}
\author[1,2,*]{Jiayu Lin}
\author[1,3,*]{Jia Wang}
\author[2,*]{Jun Xiang}
\author[2]{Jialiang Wu}
\author[1,2]{Xinnong Zhang}
\author[4]{Hanqi Yan}
\author[5]{Siyuan Wang}
\author[1,2,\dagger]{Zhongyu Wei}

\affiliation[1]{Shanghai Innovation Institute}
\affiliation[2]{Fudan University}
\affiliation[3]{Tongji University}
\affiliation[4]{King's College London}
\affiliation[5]{Chinese University of Hong Kong}
 
\contribution[*]{\textbf{Equal Contribution, names are listed in alphabetical order.}}
\contribution[\dagger]{\textbf{Corresponding author.}}

\abstract{General decision models, such as Jev, have recently emerged as efficient alternatives to LLMs for structured judgment and selection.
But what kinds of decisions can these models reliably make, and how does their behavior change when individual decisions are composed into larger systems? 
To study this, we introduce \textbf{JEVal}, a bilingual benchmark comprising 11,257 instances from 36 datasets across 10 application domains, and evaluate 25 model configurations spanning general decision models and generative LLMs. 
Our results show that 
(1) general decision models are most competitive when decisions can be resolved from available evidence, but weaken when they require specialist knowledge or faithful uncertainty estimation: they can often identify the most likely outcome while substantially overstating its probability. 
(2) In more dynamic and realistic systems involving long-horizon, multi-step interactions, the advantages of fast local decision making are offset by reliability failures at the system level. on $\tau$-bench, faster local decisions reduce median episode time but lower task success as decision errors accumulate over long trajectories. (3) In large-scale social simulation, decision models approach strong generative LLMs on individual response prediction at substantially lower inference cost, yet remain weaker in user profiling and exhibit larger aggregate estimation errors and systematic bias.
Finally, we propose \textbf{\oursmall{}} and \textbf{\ourlarge{}}, which internalize an open-weight LLM's own reasoning into a single-pass first-token decision through \textbf{\ourmethod{}}, with \ourlarge{} performing on par with Jev on JEVal while answering a typical query in about 0.1\,s.
}

\metadata[Emails]{
    \email{fyduan25@m.fudan.edu.cn}; 
    \email{zywei@fudan.edu.cn}
}
\metadata[Project Page]{
    \url{https://sii-research.github.io/sii-decision/}
}
\metadata[JEVal Data]{
    \url{https://huggingface.co/datasets/carlosxiang/JEVal}
}
\metadata[\ourname{} Code]{
    \url{https://github.com/amazingljy1206/SII-Decision}
}
\metadata[\ourname{} Models]{
    \href{https://huggingface.co/jylin001206/SII-Decision-27B-Full}{\ourlarge{} full-rank},
    \href{https://huggingface.co/jylin001206/SII-Decision-27B-LoRA}{\ourlarge{} LoRA},
    \href{https://huggingface.co/jylin001206/SII-Decision-4B-Full}{\oursmall{} full-rank},
    \href{https://huggingface.co/jylin001206/SII-Decision-4B-LoRA}{\oursmall{} LoRA}
}
\metadata[\ourname{} Training Data]{
    \href{https://huggingface.co/datasets/jylin001206/SII-Decision-27B-Training-Data}{\ourlarge{}},
    \href{https://huggingface.co/datasets/jylin001206/SII-Decision-4B-Training-Data}{\oursmall{}}
}

\begin{document}

\maketitle

\begin{figure}[h]
  \centering
  \includegraphics[width=1.0\linewidth]{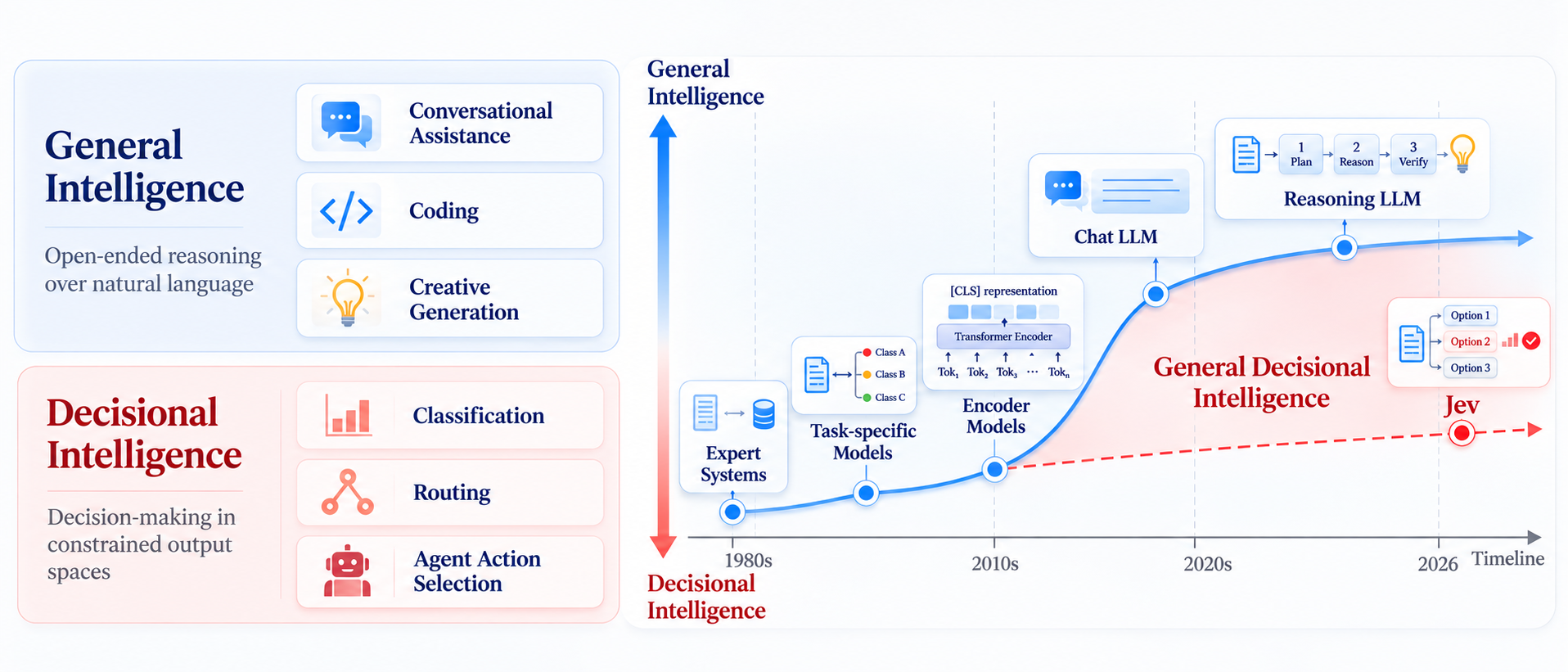}
  \caption{A conceptual view of general and decisional intelligence. The left panel contrasts open-ended language capabilities with decision-making in constrained output spaces. The right panel illustrates the evolution from expert systems and task-specific models to encoder models, chat and reasoning LLMs, and general decision models represented by Jev.}
  \label{fig:intro}
\end{figure}

\section{Introduction}


Large language models (LLMs) have increasingly exhibited \textbf{general intelligence: the ability to adapt across diverse tasks and domains without task-specific retraining}~\citep{OpenAI2023GPT4, grattafiori2024llama}. Their capabilities have expanded from language generation to reasoning, code generation, and agentic workflows~\citep{DeepSeekAI2025R1, hui2024qwen2, wang2025openhands}. 
However, modern AI systems also involve another broad class of computational demands: making structured judgments or selections directly from a given context, such as preference modeling, importance attribution, and action selection~\citep{lambert2025rewardbench,Zhou2024Attribution,wang2024executable}. We refer to this capability as \textbf{decisional intelligence: the ability to make effective structured decisions over a constrained outcome space.}

Existing models occupy different regions along these two capability dimensions, and Figure~\ref{fig:intro} provides a conceptual summary of this evolution. Early expert systems and subsequent task-specific models were primarily designed for predefined decision problems~\citep{shortliffe1977mycin, lecun1998gradient}. With the development of pre-trained representation learning, encoder models became capable of handling more complex and diverse natural-language judgment tasks~\citep{devlin2019bert}. In parallel, ChatLLMs and reasoning LLMs substantially expanded task coverage along the direction of general intelligence~\citep{team2024gemini,team2026kimi}. Recently, models represented by Jev~\citep{TypeSafe2026Jev} have begun to explore a new region: rather than targeting open-ended text generation, they directly map natural-language contexts to structured decisions while attempting to retain applicability across tasks. We define \textbf{general decisional intelligence as the ability to perform structured decision making across diverse tasks and domains without task-specific retraining.} Models that exhibit this capability are referred to as \textbf{general decision models}.

However, although decision models operate over finite output spaces, the capabilities required by different tasks can vary substantially. Knowledge-intensive decisions require access to factual knowledge~\citep{Fang2025KiRAG}. Symbolic decision tasks involving logic and mathematics further require compositional reasoning~\citep{Yu2025FormalMATH}, while long-context decisions require retrieving information distributed across extended inputs~\citep{Yang2025HundredLongBench}. In agentic settings, action selection may additionally depend on jointly reasoning over the current state, interaction history and future planning~\citep{Barres2025Tau2Bench,Abhyankar2025OSWorldHuman}. \textbf{We therefore aim to characterize the current frontier of general decisional intelligence~\footnote{We examine this frontier through Jev as a representative general decision model. Unless otherwise specified, the term ``general decision models'' refers to Jev throughout the remainder of this paper.}, from the reliability of local decisions to their behavior when composed in long-horizon workflows and aggregated in large-scale systems.} Our analysis reveals several key insights:

\paragraph{\textbf{Finding 1: General decision models perform best with explicit evidence in local decision tasks, but weaken when decisions require latent knowledge or calibrated uncertainty.}}
We introduce \textbf{JEVal}, a bilingual benchmark comprising 11,257 instances from 36 datasets across 10 application domains (Section~\ref{sec:jeval_benchmark}), and evaluate 20+ models spanning general decision models and LLMs. On JEVal, the strongest decision models approach leading thinking LLMs in overall accuracy and are particularly competitive when the evidence is contained in the input, but fall behind on specialist-knowledge domains such as medicine, finance, and law (Section~\ref{sec:main-results}). Controlled probability experiments reveal a parallel limitation: decision models can often identify the most likely outcome correctly while substantially overstating its probability (Section~\ref{sec:math-probability}).

\paragraph{\textbf{Finding 2: Strong local decision performance does not directly translate into reliable long-horizon agent behavior.}}
On $\tau$-bench~\citep{yao2024tau}, using Jev for next-action selection reduces median episode time, but also lowers task success and increases tail latency. We find that this gap mainly arises from accumulation of local decision errors over multi-turn trajectories (Sections~\ref{sec:agentic-workflows}). 

\paragraph{\textbf{Finding 3: General decision models make large-scale social simulation more scalable, but still face limitations in individual profiling, aggregate estimation, and systematic bias.}}
Across user profiling, individual survey prediction, and population-level simulation, we find that current general decision models approach strong generative LLMs on individual response prediction at substantially lower inference cost and can drive thousands of simulated agents while preserving most state-level outcomes. However, they remain less effective in user-interest profiling, exhibit substantially larger errors in aggregate vote-share estimation, and retain systematic directional biases also observed in LLMs (Sections~\ref{sec:social-simulation}).

Jev is accessible only through a hosted API. We therefore ask: \textbf{can a general decision model like Jev be built from an open-weight LLM?} Human cognition is often described as two systems: System One decides by intuition, System Two deliberates~\citep{Kahneman2011Thinking}, and intuition grows out of practiced deliberation~\citep{Kahneman2009Conditions}. In these terms, Jev is a System One model, whereas an LLM has both: its next-token distribution is already an immediate decision, and it can also think before answering, repairing wrong immediate decisions. Therefore, \textbf{the model can build its intuition as humans do, teaching its System One with its own System Two.} Concretely, we restrict the logits of the first answer token to the labels of the valid options, which gives a decision head that covers all three JEVal decision formats in a single forward pass. We then propose \textit{\ourmethod{}}, which distills the answer distribution that the model reaches at the end of its reasoning chain into this head, without labeled answers or a stronger teacher. Applied to Qwen3.5-4B and Qwen3.8-27B, it yields two open-source decision models, \texttt{\oursmall{}} and \texttt{\ourlarge{}}. \ourlarge{} performs comparably to Jev on JEVal, and \oursmall{} surpasses decision models of the same size, while answering a typical query in about 60\,ms and 110\,ms, respectively, on a single H100 GPU.

Our contributions are summarized as follows:
\begin{itemize}
    \item \textbf{A comprehensive decision-model benchmark spanning 10 domains and over 10K bilingual instances.} We introduce JEVal, which formulates diverse tasks as decision problems. JEVal contains 11,257 bilingual instances from 36 datasets across 10 application domains.

    \item \textbf{A systematic analysis of the capability boundaries and failure modes of general decision models.} Across JEVal and targeted diagnostic tasks, we find that general decision models are strongest when decisions can be resolved from available evidence, but weaken when they require specialist knowledge or faithful uncertainty estimation.

    \item \textbf{Evaluation in dynamic and realistic decision systems.} We move beyond static benchmarks to evaluate general decision models in interactive agent workflows and large-scale social simulations, revealing system-level behaviors that cannot be observed from isolated decision accuracy alone.

    \item \textbf{Open general decision models.} We open-source \ourname{}, a general decision model at both 4B and 27B scales with competitive performance.
    
\end{itemize}

\section{Related Work}


\paragraph{From task-specific predictors to general decision models.}
Decision-making is a fundamental capability of intelligent systems. Early discriminative models rely on task-specific predictors, such as classification heads for pretrained encoders~\citep{devlin2019bert} or candidate scoring in rerankers~\citep{Zhang2025Qwen3Embedding}. Later work relaxes fixed label spaces through natural-language task and label descriptions, including entailment-based zero-shot classification~\citep{Yin2019ZeroShot}, the Universal Discriminator~\citep{Xu2023UniversalDiscriminator}, and GLiClass~\citep{Stepanov2025GLiClass}. Jev further extends this direction toward general decision models with request-time choices, binary judgments, and scores~\citep{TypeSafe2026Jev}. However, how broadly such models generalize beyond conventional classification remains unclear, motivating the systematic evaluation in JEVal.


\paragraph{Evaluation of decision models.}
A broad range of existing benchmarks can be formulated as decision tasks over finite output spaces, covering capability demands such as knowledge reasoning, long-context understanding, and agentic action selection. MMLU-Pro evaluates knowledge and reasoning~\citep{wang-etal-2024-mmlu-pro}, LongBench focuses on long-context understanding~\citep{bai-etal-2024-longbench}, and $\tau$-bench evaluates interactive tool use~\citep{yao2024tau}, while HELM highlights joint evaluation of accuracy, calibration, and efficiency~\citep{Liang2022HELM}. Recent studies have also evaluated Jev in text annotation~\citep{Ibrahim2026DecisionAnnotation}, response judging~\citep{Li2026JevJudge,Rao2026RubricJudges}, alignment-failure detection~\citep{Guo2026RLCDAlignBench}, legal-document evaluation~\citep{Zhang2026LegalJev}, and agent delegation~\citep{Wu2026REFLEX}, alongside studies of output validity~\citep{Sun2026TypeSafe}, context sensitivity~\citep{Xu2026JevOut}, and confidence calibration~\citep{Guo2017Calibration}. However, these evaluations typically focus on individual tasks, applications, or reliability aspects, leaving the broader capabilities of general decision models underexplored. JEVal therefore unifies diverse decision tasks and evaluation dimensions to systematically characterize their capability frontier.


\section{JEVal Benchmark}
\label{sec:jeval_benchmark}

\begin{figure}
    \centering
    \includegraphics[width=1.0\linewidth]{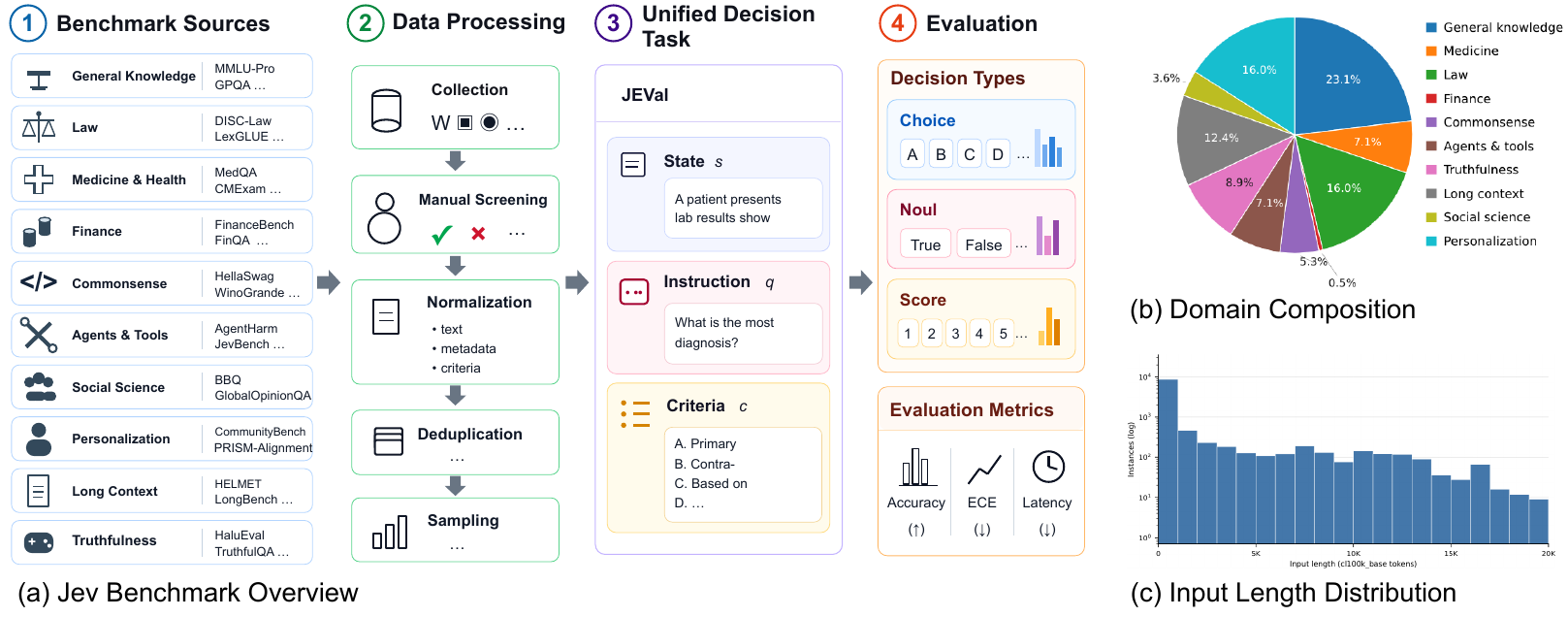}
    \caption{Overview of JEVal. (a) Data processing pipeline. (b) Share of the 11{,}257 JEVal instances contributed by each domain. (c) Input length over JEVal measured over the state, instruction, and candidates in \texttt{cl100k\_base} tokens. The frequency axis is logarithmic.}
    \label{fig:framework}
\end{figure}

In this section, we present \textbf{JEVal}, a benchmark for evaluating general decisional intelligence. Figure~\ref{fig:framework} show the overview of JEVal. We first introduce the task formulation(Section~\ref{sec:task_formulation}), then describe how the benchmark is collected, curated, and normalized (Section~\ref{sec:data-processing}), and finally we report the statistics of our benchmark (Section~\ref{sec:dataset-statistics}).

\subsection{Task formulation}
\label{sec:task_formulation}

JEVal formulates each evaluation sample as a general decision instance

\begin{equation}
    \mathcal{D} = (x, y^{*}), \qquad x = (s, i, c)
\end{equation}

where \(s\), \(i\), and \(c\) denote the \textit{state}, \textit{instruction}, and \textit{criteria}, respectively, and \(y^{*}\) is the gold label. The state provides the information required for the decision, the instruction specifies the decision objective, and the criteria defines the valid decision space. Let \(\mathcal{Y}(x)\) denote the finite decision space corresponding to input \(x\), whose valid range is determined by \(c\), with \(y^{*} \in \mathcal{Y}(x)\). Given \(x\), the model is required to produce a probability distribution \(p_{\theta}(y \mid x)\) over \(\mathcal{Y}(x)\), subject to

\begin{equation}
    p_{\theta}(y \mid x) \geq 0, \qquad
\sum_{y \in \mathcal{Y}(x)} p_{\theta}(y \mid x) = 1,
\end{equation}

and the final decision is obtained as

\begin{equation}
    \hat{y} = \arg\max_{y \in \mathcal{Y}(x)} p_{\theta}(y \mid x)
\end{equation}

According to the structure of the decision space, JEVal categorizes tasks into three primary formats:
\begin{itemize}
    \item \textbf{choice}: the model selects one outcome from an explicitly specified finite set of candidates;
    \item \textbf{noul}: the model performs a binary judgment over the decision space $\{\texttt{true}, \texttt{false}\}$;
    \item \textbf{score}: the model predicts one value from an ordered set of discrete rating levels.
\end{itemize}

\subsection{Data Processing}
\label{sec:data-processing}

\paragraph{Data Collection.} We first define several common domains, including general knowledge reasoning, commonsense inference, agents, mathematics, medicine, among others. We collect datasets from each domain that are compatible with the decision tasks defined in Section~\ref{sec:task_formulation}. Specifically, we prioritize datasets with explicit decision objectives and well-defined outcome spaces. Tasks that cannot be reliably converted into a finite decision space, or whose semantics would be altered by such conversion, are excluded.

\paragraph{Data Curation and Normalization.} For the selected datasets, two human experts further inspect the task definitions and sample content, remove instances that are unsuitable for decision-task formulation, and retain only datasets whose reference answers have been manually verified. We then normalize the samples into the unified JEVal format: background text, dialogue history, or environment states required for decision making are organized as the \texttt{state}; the original questions or task descriptions are used as the \texttt{instructions}; and the meanings of candidate options in the \texttt{criteria} are supplemented based on the original dataset documentation. For tasks that do not explicitly provide candidate options, we construct the candidate space and randomly shuffle the option order to mitigate positional bias. We control input length against a 20K-token budget (20{,}480 tokens) and cap the number of candidate options at 255; the retained upper tail above this length budget is quantified in Appendix~\ref{app:dataset-statistics}.

\paragraph{Deduplication and Sampling.} After normalization, we remove duplicate questions using exact string matching, resulting in more than 202K candidate samples. As the source datasets vary substantially in size, we further resample the data to prevent a small number of large datasets from dominating the overall evaluation and to improve balance across domains and subtasks. For datasets with an official test split, we sample at most 200 instances from the test set for each subtask. For datasets without an official test split, we apply the same sampling rule to the available validation or train split.

Further dataset details are provided in Appendix~\ref{app:datasets-by-domain}.



\subsection{Benchmark Statistics}
\label{sec:dataset-statistics}


The resulting benchmark covers 10 application domains, 36 datasets, and 58 subtasks, comprising 11{,}257 instances in total. As shown in Figure~\ref{fig:framework}:, general knowledge and reasoning is the largest domain (23.1\%), followed by law and personalization and preference (16.0\% each). JEVal contains 9{,}457 English and 1{,}800 Chinese instances. Most examples use the \texttt{choice} format (85.1\%), followed by \texttt{noul} (12.0\%) and \texttt{score} (2.8\%). For \texttt{choice} tasks, the number of candidates ranges from 2 to 255, with a median of five. Figure~\ref{fig:framework} shows the input-length distribution up to 20K tokens, measured over the joint length of state, instruction, and candidates. Additional format, candidate-count, and domain-level length statistics are provided in Appendix~\ref{app:dataset-statistics}.

\section{Experiments}
\label{sec:experimental-setup}
\subsection{Models and Evaluation Protocol}
\label{sec:models-evaluation-protocol}

\paragraph{Models.}
We evaluate 25 configurations in three groups. Appendix~\ref{sec:Model_Configurations} lists the settings of each.
\begin{itemize}
  \item \textbf{Decision models.} Jev~\citep{TypeSafe2026Jev}; the open-weight Kev\footnote{\url{https://github.com/jaredpalmer/kev}} at 0.8B, 4B, 9B, and 27B; Intern-Decision\footnote{\url{https://github.com/InternLM/Intern-Decision}} at 0.8B, 2B, and 4B; Laya\footnote{\url{https://github.com/NandhaKishorM/laya}}, OpenJev-4B-v5\footnote{\url{https://huggingface.co/AlexWortega/openjev}}, Nimble-9B\footnote{\url{https://github.com/bespokelabsai/nimble}}, NanoJev\footnote{\url{https://github.com/TianyuCodings/NanoJev}}, LightJev\footnote{\url{https://github.com/rongxinzy/LightJev}}, and JevForge-0.8B\footnote{\url{https://github.com/zwliJay/jev-forge}}; and our \oursmall{} and \ourlarge{} (Section~\ref{sec:self-distillation}).
  \item \textbf{Non-thinking generative LLMs.} Qwen3.5-4B~\citep{QwenTeam2026Qwen35}, Qwen3.8-27B~\citep{QwenTeam2026Qwen38}, DeepSeek-V4-Flash~\citep{DeepSeekAI2026V4}, GPT-6-Luna, and GPT-6-Sol~\citep{OpenAI2026GPT6SolLuna}. For the two GPT-6 models, we request no reasoning through the API.
  \item \textbf{Thinking generative LLMs.} Qwen3.5-4B at its default effort, Qwen3.8-27B at \texttt{xhigh} effort, and DeepSeek-V4-Flash and GLM-5.3-Flash~\citep{ZAI2026GLM53FlashBlog} at \texttt{high} effort.
\end{itemize}

\paragraph{Implementation details.}
We deploy open-weight models locally on NVIDIA H100 GPUs using vLLM and access Jev and the GPT-6 models through hosted APIs. Generative LLMs use a sampling temperature of 1.0, with reasoning and final-answer tokens counted toward a shared output budget. Open-weight decision baselines retain their released numerical precision and calibration parameters. Appendix~\ref{sec:Model_Configurations} provides detailed hardware configurations, numerical precision settings, and token budgets.

\paragraph{Evaluation metrics.}
We evaluate predictive performance and response efficiency using the following metrics. Let $\mathcal C$ denote all 10,938 classification instances and $\mathcal V_m\subseteq\mathcal C$ the subset with valid responses from configuration $m$.
\begin{itemize}
    \item \textbf{Accuracy.}
    We measure decision correctness as
    \begin{equation}
    \mathrm{Accuracy}_m
    =
    \frac{100}{|\mathcal C|}
    \sum_{i\in\mathcal V_m}
    \mathbf{1}[\hat y_{mi}=y_i],
    \label{eq:accuracy}
    \end{equation}
    where $\hat y_{mi}$ and $y_i$ are the predicted and reference labels.

    \item \textbf{Expected calibration error (ECE).}
    We partition valid responses into ten equal-width confidence bins $B_{mj}$ and compute
    \begin{equation}
    \mathrm{ECE}_m
    =
    \sum_{\substack{j=0\\|B_{mj}|>0}}^{9}
    \frac{|B_{mj}|}{|\mathcal V_m|}
    \left|
    \operatorname{acc}(B_{mj})
    -
    \operatorname{conf}(B_{mj})
    \right|,
    \label{eq:ece}
    \end{equation}
    where $\operatorname{acc}$ and $\operatorname{conf}$ denote the fraction of correct predictions and mean confidence in each bin.

    \item \textbf{Raw probability-vector squared error ($\mathrm{Brier}_{\mathrm{raw}}$).}
    We evaluate candidate probabilities against one-hot labels:
    \begin{equation}
    \mathrm{Brier}_{\mathrm{raw},m}
    =
    \frac{1}{|\mathcal V_m|}
    \sum_{i\in\mathcal V_m}
    \sum_{k=1}^{K_i}
    (p_{mik}-y_{ik})^2,
    \label{eq:brier}
    \end{equation}
    where $K_i$ is the candidate count, $p_{mik}\in[0,1]$ is the reported probability, and $y_{ik}$ is the one-hot reference. 

    \item \textbf{Response latency.}
    We report mean latency over valid responses on a separate 200-question test, stratified into six input-length bins up to 16K Jev-reported tokens. 
\end{itemize}
MAE on \texttt{score} tasks and benchmark-run latencies from a separate setup are reported in Appendix~\ref{app:benchmark-results}.

\begin{table}[t]
\centering
\caption{\textbf{The strongest decision models match thinking LLMs in accuracy and are better calibrated than every generative LLM.} We evaluate 25 configurations on the 10,938 \texttt{choice} and \texttt{noul} questions of JEVal. Accuracy counts invalid responses as incorrect, while ECE and $\mathrm{Brier}_{\mathrm{raw}}$ are computed over valid responses.}
\label{tab:overall-results}
\begingroup\fontsize{9}{10.8}\selectfont\setlength{\tabcolsep}{4pt}\renewcommand{\arraystretch}{1.07}
\begin{tabular*}{0.90\linewidth}{@{\extracolsep{\fill}}lrrr@{}}
\toprule
\textbf{Model} & \textbf{ACC (\%)} $\uparrow$ & \textbf{ECE} $\downarrow$ & $\mathbf{Brier}_{\mathbf{raw}}$ $\downarrow$\\
\midrule
\multicolumn{4}{@{}l}{\textit{General Decision models}}\\*
\midrule
Jev & 78.38 & 0.0532 & \underline{0.287}\\
Kev-0.8b & 49.63 & 0.0432 & 0.617\\
Kev-4b & 68.90 & 0.0843 & 0.435\\
Kev-9b & 70.87 & 0.0579 & 0.404\\
Kev-27b & 77.17 & 0.0356 & 0.304\\
Intern-Decision-0.8B & 47.71 & 0.0635 & 0.643\\
Intern-Decision-2B & 55.72 & 0.0214 & 0.553\\
Intern-Decision-4B & 70.74 & \underline{0.0152} & 0.399\\
Laya & 26.30 & 0.2485 & 0.833\\
Openjev-4b-v5 & 70.20 & 0.1472 & 0.481\\
Nimble-9b & 73.87 & 0.1045 & 0.369\\
Nanojev & 25.34 & 0.1300 & 0.806\\
Lightjev & 37.20 & 0.1145 & 0.742\\
Jevforge-0.8b & 31.04 & 0.1164 & 0.794\\
\textbf{SII-Decision-4B (ours)} & 71.92 & 0.0843 & 0.383\\
\textbf{SII-Decision-27B (ours)} & 79.24 & \textbf{0.0149} & \textbf{0.279}\\
\addlinespace[2pt]
\midrule
\multicolumn{4}{@{}l}{\textit{Non-thinking generative LLMs}}\\*
\midrule
Qwen3.5-4B & 61.37 & 0.2526 & 0.781\\
Qwen3.8-27B & 73.81 & 0.1697 & 0.446\\
DeepSeek-V4-Flash & 72.83 & 0.1923 & 0.494\\
GPT-6-Luna & \textbf{80.93} & 0.1159 & 0.297\\
GPT-6-Sol & 79.92 & 0.1309 & 0.322\\
\addlinespace[2pt]
\midrule
\multicolumn{4}{@{}l}{\textit{Thinking generative LLMs}}\\*
\midrule
Qwen3.5-4B \texttt{default think effort} & 70.73 & 0.1716 & 0.489\\
Qwen3.8-27B \texttt{xhigh} & 79.49 & 0.1005 & 0.319\\
DeepSeek-V4-Flash \texttt{high} & 79.69 & 0.1449 & 0.382\\
GLM-5.3-Flash \texttt{high} & \underline{80.37} & 0.0784 & 0.316\\
\addlinespace[2pt]
\bottomrule
\end{tabular*}
\par\endgroup
\end{table}

\begin{figure}[tp]
\centering
\includegraphics[width=\textwidth]{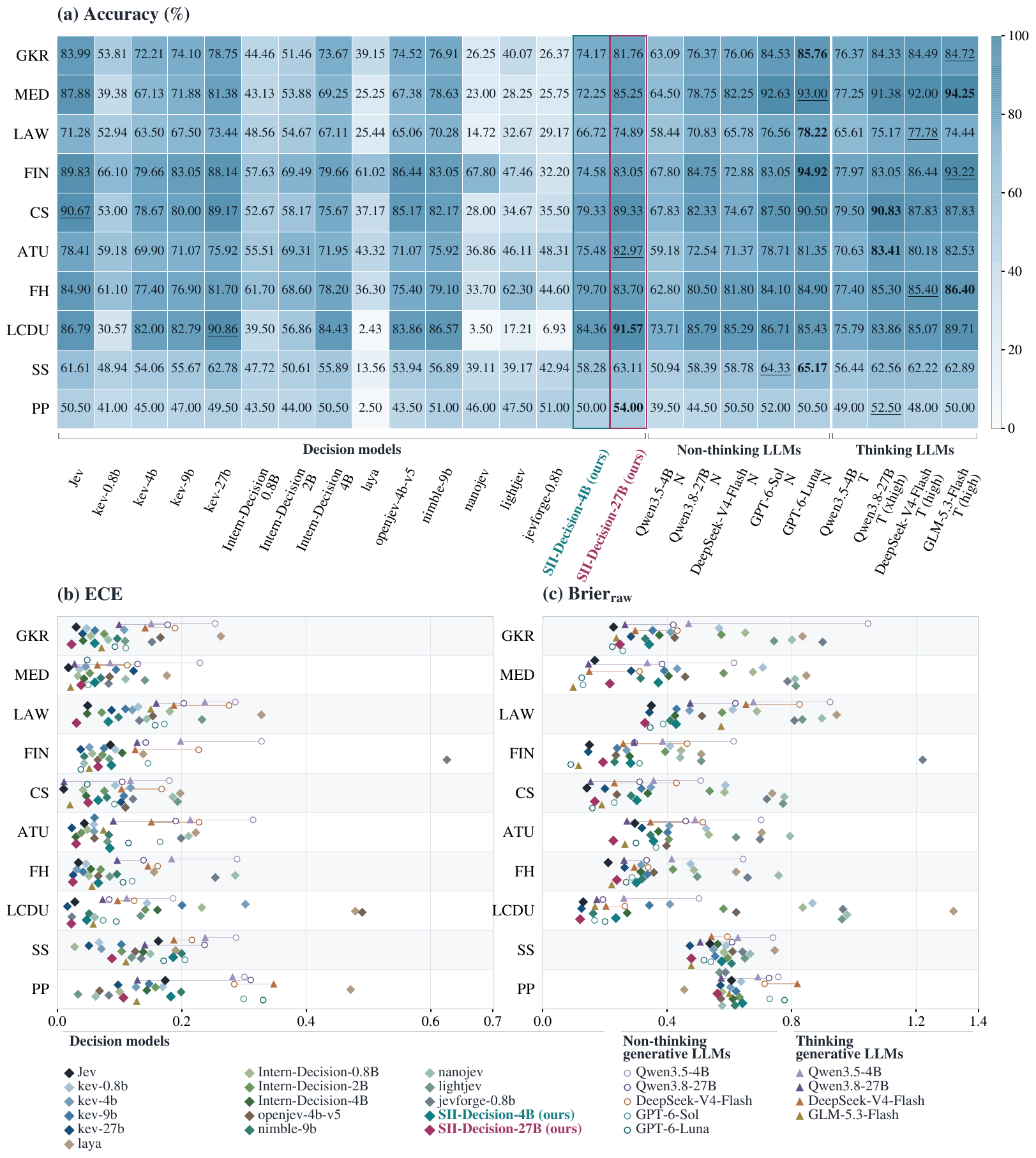}
\caption{\textbf{Decision models lead when the evidence is in the input and trail generative LLMs in medicine, finance, and law.} We break the JEVal results of all 25 configurations down by the ten application domains. (a) Accuracy, (b) ECE, and (c) $\mathrm{Brier}_{\mathrm{raw}}$; N and T denote non-thinking and thinking. GKR: general knowledge and reasoning; MED: medicine; LAW: law; FIN: finance; CS: commonsense; ATU: agent and tool use; FH: factuality and hallucination; LCDU: long-context and document understanding; SS: social science; PP: personalization and preferences.}
\label{fig:domain-results}
\end{figure}

\begin{figure}[t]
\centering
\includegraphics[width=\textwidth]{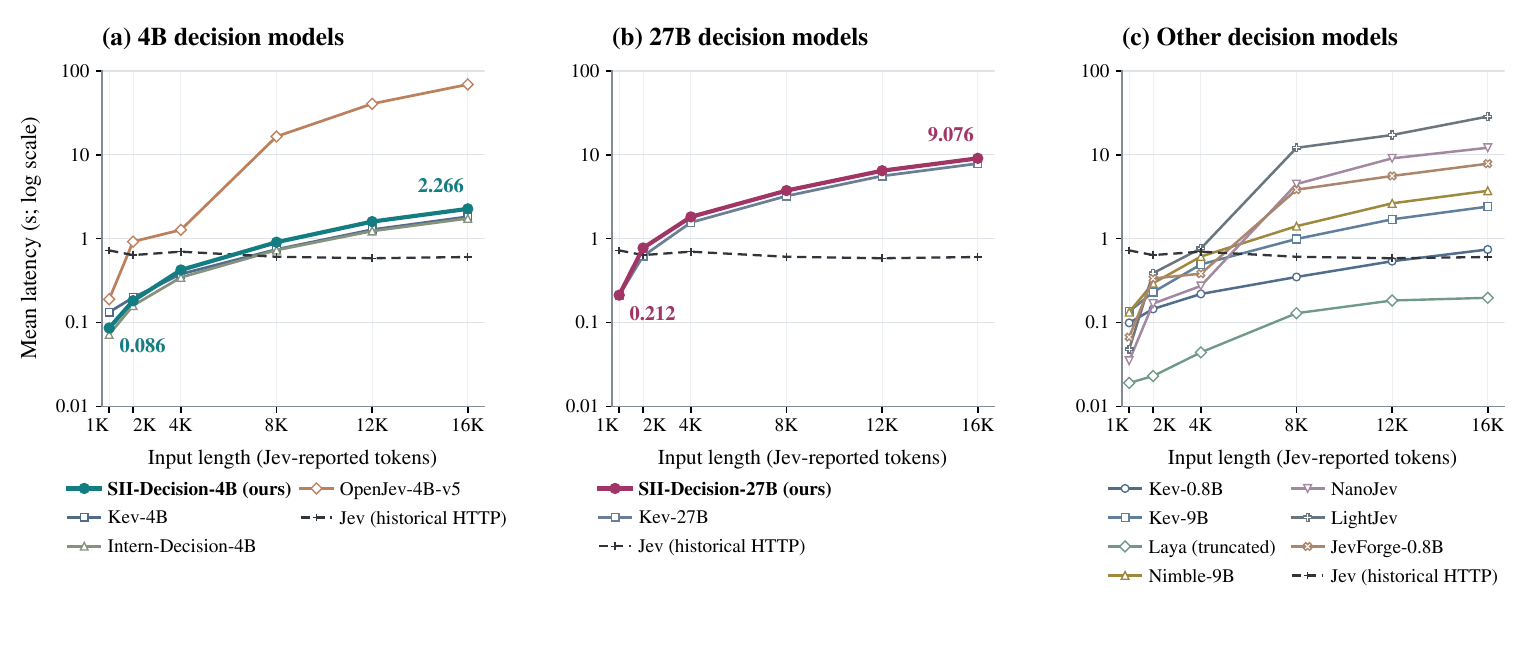}
\caption{\textbf{Local decision models are faster than hosted Jev on short inputs and slower on long ones.} We measure mean latency on a 200-question test stratified into six input-length bins of up to 16K tokens. (a) 4B decision models, (b) 27B decision models, and (c) the other decision models. Jev's latency is that of its hosted API and includes network transport, whereas local models run on RTX 4090 GPUs and exclude queueing and transport.}
\label{fig:latency-scaling}
\end{figure}

\subsection{Main Results}
\label{sec:main-results}

\begin{finding}
The strongest decision models \textbf{match thinking LLMs in accuracy} with \textbf{a single forward pass}.
\end{finding}

\paragraph{Overall accuracy.}
The seven most accurate configurations in Table~\ref{tab:overall-results} fall within three points of one another. They are the two GPT-6 models, the thinking modes of GLM-5.3-Flash, DeepSeek-V4-Flash, and Qwen3.8-27B, and two decision models, \ourlarge{} and Jev.
Open-weight generative LLMs reach this group only with thinking, which adds six to nine points to their accuracy. Without thinking, the best of them is more than four points below Jev.
\ourlarge{} matches the thinking mode of its backbone, Qwen3.8-27B, and \oursmall{} is more accurate than thinking Qwen3.5-4B and every other 4B decision model (Section~\ref{sec:self-distillation}).
Among open decision models, accuracy grows with model size: Kev rises from about 50\% at 0.8B to 77\% at 27B. Laya, NanoJev, LightJev, and JevForge-0.8B stay below 40\%.

\begin{finding}
Decision models \textbf{lead when the evidence is in the input} and \textbf{trail when the decision requires specialist knowledge}.
\end{finding}

\paragraph{Accuracy by domain.}
Figure~\ref{fig:domain-results}(a) reports accuracy in each of the ten domains.
On long-context and document understanding, \ourlarge{} and Kev-27B are the two most accurate of all 25 configurations, and thinking gives the generative LLMs almost no gain. On agent and tool use, \ourlarge{} is within half a point of the best configuration.
The order reverses in medicine, finance, and law, where the best decision model is three to six points below the best generative LLM. Thinking does not explain this gap, since GPT-6-Luna ranks first or second in all three domains without thinking.
Among decision models, Jev leads in general knowledge, medicine, finance, commonsense, and factuality, and \ourlarge{} leads in the other five domains.
Social science and personalization are the hardest domains for every configuration, and no configuration exceeds 54\% on personalization. Appendix~\ref{app:benchmark-results} reports every benchmark.

\begin{finding}
Decision models give \textbf{confidence that tracks their accuracy} more closely than the probabilities generative LLMs \textbf{write out}.
\end{finding}

\paragraph{Calibration.}
Jev, Kev-27B, Intern-Decision-4B, and \ourlarge{} all have lower ECE than every generative configuration, including GPT-6-Luna, the most accurate one. The ECE of \ourlarge{} is the lowest in Table~\ref{tab:overall-results}, about one fifth of that of the best-calibrated generative LLM.
Thinking lowers the ECE of every LLM run in both modes, but not to the level of these decision models.
\ourlarge{} and Jev also have the lowest $\mathrm{Brier}_{\mathrm{raw}}$.
ECE covers only the confidence of the returned option: Kev-0.8B is well calibrated while answering half of the questions correctly. Section~\ref{sec:math-probability} examines the full probability distribution, where Jev overstates the probability of its top choice.

\begin{finding}
Local decision models are \textbf{faster than hosted Jev on short inputs} and \textbf{slower on long ones}.
\end{finding}

\paragraph{Latency.}
Figure~\ref{fig:latency-scaling} reports mean latency by input length. Jev's hosted latency, network transport included, stays below one second at every length, while the latency of local models grows with input length.
On inputs of up to 1K tokens, which make up most of JEVal, every local model is faster than Jev: \oursmall{} answers in under 0.1\,s and \ourlarge{} in about 0.2\,s. \ourlarge{} is as fast as Kev-27B and two points more accurate.
The 27B models become slower than Jev at about 2K tokens, and most 4B models between 4K and 8K tokens. On the longest inputs, the 27B models take several seconds per question.

\section{Analysis: From Local Decisions to Systems}
\label{sec:analysis}

This section goes beyond overall accuracy to examine the capability boundaries and system-level behavior of decision models. We first probe \textbf{local decision boundaries}: exact numerical and probabilistic reasoning (Section~\ref{sec:math-probability}) and long inputs (Section~\ref{sec:long-context}). Choosing the most likely outcome is not the same as estimating its probability. We then examine \textbf{long-horizon decision making} in a multi-step tool-using agent (Section~\ref{sec:agentic-workflows}), where faster action selection reduces median episode time but does not improve task success. Finally, we examine \textbf{large-scale decision systems} through social simulation from individuals to populations (Section~\ref{sec:social-simulation}). Low-cost inference supports scalable simulation with competitive individual predictions, while profiling limitations, aggregate errors, and systematic bias remain.


\subsection{Struggles with Exact Computation and Probability Estimation}
\label{sec:math-probability}

\begin{finding}
Decision models \textbf{choose well but estimate poorly}: they identify the most likely outcome while \textbf{overstating its probability}, and they land \textbf{near exact numerical answers} but rarely on them.
\end{finding}

\begin{table}[b]
\centering
\caption{\textbf{Jev lands near the exact answer and picks the most likely label, while thinking LLMs recover both the exact value and the probabilities.} We construct four tasks of 100 items each, two with a unique numerical answer and two with an analytically determined distribution, and compare Jev with four LLMs, three of them in both inference modes. Acc and Acc$_{\pm 1}$ are the share of answers that are exact and within one integer of the gold value; $E_{\mathrm{mode}}$ and $E_{\mathrm{all}}$ are the mean absolute errors, in percentage points, of the probability of the gold option and of all four options.}
\label{tab:math-probability}
\footnotesize
\setlength{\tabcolsep}{3pt}
\renewcommand{\arraystretch}{1.1}
\resizebox{\textwidth}{!}{%
\begin{tabular}{@{}llcccccccccc@{}}
\toprule
& & \multicolumn{2}{c}{\textbf{Repeating division}} & \multicolumn{2}{c}{\textbf{Area inversion}} & \multicolumn{3}{c}{\textbf{Prior probability}} & \multicolumn{3}{c}{\textbf{Posterior probability}} \\
\cmidrule(lr){3-4}\cmidrule(lr){5-6}\cmidrule(lr){7-9}\cmidrule(lr){10-12}
\textbf{Model} & \textbf{Inference mode} & \textbf{Acc (\%)}$\uparrow$ & \textbf{Acc}$_{\pm1}$ \textbf{(\%)}$\uparrow$ & \textbf{Acc (\%)}$\uparrow$ & \textbf{Acc}$_{\pm1}$ \textbf{(\%)}$\uparrow$ & \textbf{Acc (\%)}$\uparrow$ & $\boldsymbol{E}_{\mathrm{mode}}$ \textbf{(pp)}$\downarrow$ & $\boldsymbol{E}_{\mathrm{all}}$ \textbf{(pp)}$\downarrow$ & \textbf{Acc (\%)}$\uparrow$ & $\boldsymbol{E}_{\mathrm{mode}}$ \textbf{(pp)}$\downarrow$ & $\boldsymbol{E}_{\mathrm{all}}$ \textbf{(pp)}$\downarrow$ \\
\midrule
\textbf{Jev} & Native & 44 & 97 & 39 & 73 & \underline{\textbf{100}} & 53.75 & 26.91 & 67 & 28.12 & 18.13 \\
\textbf{Qwen3.5-4B} & Non-thinking & 34 & 68 & 24 & 45 & 48 & 18.66 & 13.12 & 38 & 21.84 & 17.40 \\
\textbf{Qwen3.8-27B} & Non-thinking & 53 & 99 & 45 & 95 & 98 & 2.85 & 2.06 & 46 & 12.37 & 9.87 \\
\textbf{DeepSeek-V4-Flash} & Non-thinking & 60 & \underline{\textbf{100}} & 69 & 98 & 66 & 9.90 & 6.54 & 43 & 18.67 & 14.99 \\
\midrule
\textbf{Qwen3.5-4B} & default think & 72 & 73 & 82 & 84 & 81 & 0.41 & 0.38 & 82 & 0.74 & 0.63 \\
\textbf{Qwen3.8-27B} & Thinking (xhigh) & 98 & \underline{\textbf{100}} & \underline{\textbf{99}} & \underline{\textbf{100}} & \underline{\textbf{100}} & 0.10 & 0.06 & \underline{\textbf{100}} & 0.79 & 0.48 \\
\textbf{DeepSeek-V4-Flash} & Thinking (high) & \underline{\textbf{100}} & \underline{\textbf{100}} & 96 & 97 & 98 & \underline{\textbf{0.00}} & \underline{\textbf{0.00}} & 93 & \underline{\textbf{0.0011}} & \underline{\textbf{0.0014}} \\
\textbf{GLM-5.3-Flash} & Thinking (high) & \underline{\textbf{100}} & \underline{\textbf{100}} & \underline{\textbf{99}} & \underline{\textbf{100}} & 97 & \underline{\textbf{0.00}} & \underline{\textbf{0.00}} & 99 & 0.15 & 0.23 \\
\bottomrule
\end{tabular}%
}
\end{table}

\begin{figure}[t]
\centering
\includegraphics[width=0.6\columnwidth]{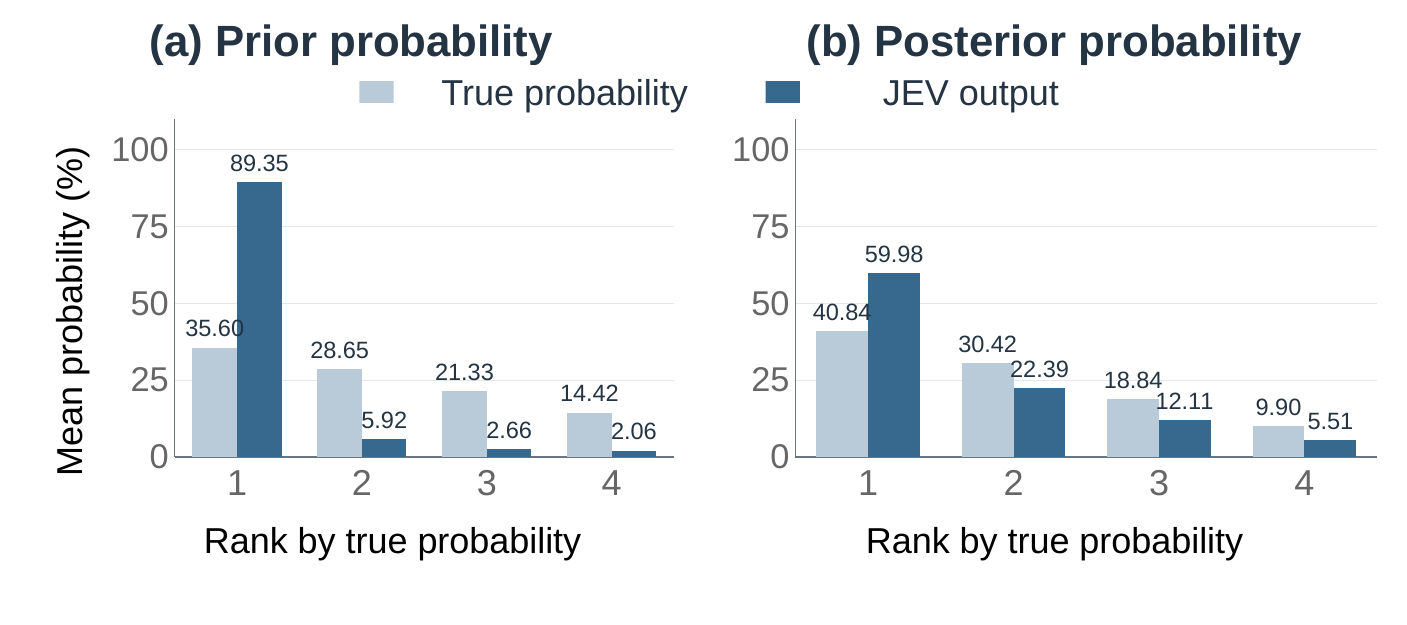}
\caption{\textbf{Jev ranks the options correctly but concentrates the probability on its top choice.} We compare Jev's output probabilities with analytically determined distributions over four options, using 100 items per task. Options are ordered by true probability. (a) Prior items, where no clue is observed. (b) Posterior items, where the probabilities follow from an observed clue.}
\label{fig:probability-concentration}
\end{figure}

To assess decision models' mathematical ability and decision-making under uncertainty, we construct four tasks with 100 items each. Examples of these tasks are shown in Figure~\ref{fig:math-probability-benchmark} in Appendix~\ref{app:math-prompts}. Every mathematical item has a unique correct answer; every probability item has an analytically determined distribution. We compare Jev with four LLMs under different inference modes, following the input/output format requirements of Section~\ref{sec:experimental-setup}; prompts are also provided in Appendix~\ref{app:math-prompts}.

\paragraph{Tasks and metrics.}
For \emph{repeating-decimal division}, the target quantity is $z=a/b\in[0,100]$, with a non-terminating decimal expansion near a rounding boundary. For \emph{area inversion}, $z=\sqrt{A}$ for 50 square sides and $z=\sqrt{A/\pi}$ for 50 circle radii. Both use $\mathcal{C}=\{0,\ldots,100\}$ and gold answer $y^*=\arg\min_{n\in\mathcal{C}}|z-n|$, excluding $z\in\mathbb{Z}+\tfrac12$. With returned choice $\hat y_i$ and $N=100$, we report
\begin{equation}
\mathrm{Acc}_{\pm d}=\frac{100}{N}\sum_i\mathbf{1}\{|\hat y_i-y_i^*|\le d\},
\label{eq:math-accuracy}
\end{equation}
for $d\in\{0,1\}$; $d=0$ gives Acc, and missing choices score zero.

For \emph{prior} and \emph{posterior probability}, let $Y\in\{A,B,C,D\}$ be a hidden label, with prior $\pi_k=n_k/\sum_j n_j$ and likelihood $\ell_k=P(X\mid Y=k)$. Prior items provide no observation, giving $p_k=\pi_k$. Posterior items observe $e=X$ in 50 cases and $e=\bar X$ in the other 50, giving
\begin{equation}
 p_k=P(Y=k\mid e)=\frac{\pi_k L_k(e)}{\sum_j\pi_jL_j(e)},
\label{eq:math-bayes}
\end{equation}
where $L_k(X)=\ell_k$ and $L_k(\bar X)=1-\ell_k$. The gold choice is $k_i^*=\arg\max_k p_{ik}$. We report choice Acc and probability errors in percentage points (pp). Let $\mathcal{V}$ denote the valid responses and $N_{\mathrm{valid}}=|\mathcal{V}|$. For raw returned probabilities $q_{ik}$,
\begin{align}
E_{\mathrm{mode}}&=\frac{100}{N_{\mathrm{valid}}}\sum_{i\in\mathcal{V}}|q_{ik_i^*}-p_{ik_i^*}|,\label{eq:math-mode-error}\\
E_{\mathrm{all}}&=\frac{100}{4N_{\mathrm{valid}}}\sum_{i\in\mathcal{V}}\sum_k|q_{ik}-p_{ik}|.\label{eq:math-all-error}
\end{align}

\paragraph{Jev lands near the exact answer but rarely on it.}
Jev answers 44\% of division items and 39\% of area items exactly, above only non-thinking Qwen3.5-4B among the evaluated configurations, yet 97\% and 73\% of its answers fall within one integer of the gold value (Table~\ref{tab:math-probability}). Because every item lies near a rounding boundary, this pattern shows that Jev recovers the magnitude of the quotient or root but not the final digit that decides the rounding. Thinking supplies exactly this step: Qwen3.8-27B rises from 53\% to 98\% on division and from 45\% to 99\% on area inversion once it reasons.

\paragraph{Jev selects the most likely label but overstates its probability.}
Jev picks the mode on all 100 prior items and on 67 posterior items, ahead of every evaluated non-thinking LLM on both tasks. Its probabilities, however, carry the largest errors of all configurations. On prior items, it assigns the true mode 89.35\% of the mass on average, against 35.60\% in the true distribution (Figure~\ref{fig:probability-concentration}); on posterior items, its own two highest-probability options receive 90.30\% of the mass. Jev therefore ranks the options well but compresses the distribution toward its top choice. Bayesian updating adds a second limitation: with thinking enabled, every evaluated LLM surpasses Jev in posterior accuracy, reaching 82 to 100\% against its 67\%.

\paragraph{Accuracy hides this gap wherever probabilities are consumed.}
A decision that only needs the argmax is unaffected by a compressed distribution, but any downstream use that averages or thresholds probabilities inherits the distortion. Section~\ref{sec:electionsim} shows the population-level consequence: aggregated voter decisions call the right states while misestimating their vote shares.

\subsection{Long Contexts Still Degrade Constrained Decisions}
\label{sec:long-context}

\begin{finding}
A \textbf{constrained output space} does not make the input easier to read: decision models \textbf{degrade with context length} and \textbf{lose evidence in the middle}, just as generative LLMs do.
\end{finding}

\begin{figure*}[htbp]
  \centering
  \begin{subfigure}[t]{.223\linewidth}
    \centering
    \includegraphics[width=\linewidth]{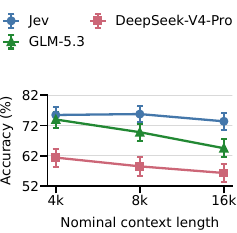}
    \caption{BABILong}
    \label{fig:babilong-context}
  \end{subfigure}\hfill
  \begin{subfigure}[t]{.223\linewidth}
    \centering
    \includegraphics[width=\linewidth]{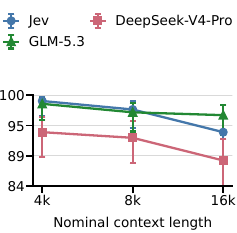}
    \caption{NoLiMa-MC}
    \label{fig:nolima-context}
  \end{subfigure}\hfill
  \begin{subfigure}[t]{.459\linewidth}
    \centering
    \includegraphics[width=\linewidth]{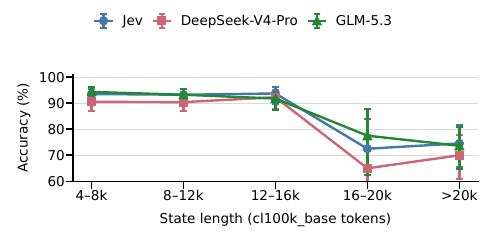}
    \caption{LongBench}
    \label{fig:longbench-context}
  \end{subfigure}
  \caption{\textbf{Accuracy falls as the context grows, for Jev and for generative LLMs alike.} We evaluate Jev, DeepSeek-V4-Pro, and GLM-5.3 on three long-context benchmarks. (a) BABILong and (b) NoLiMa-MC at context lengths of 4k, 8k, and 16k. (c) LongBench, binned by the length of the \texttt{state} field. Error bars are 95\% Wilson intervals.}
  \label{fig:context-length-results}
\end{figure*}

\begin{figure*}[t]
  \centering
  \begin{subfigure}[t]{.32\textwidth}
    \centering
    \includegraphics[width=\linewidth]{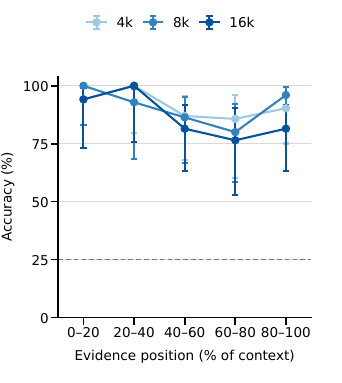}
    \caption{Jev}
    \label{fig:evidence-position-jev}
  \end{subfigure}
  \begin{subfigure}[t]{.32\textwidth}
    \centering
    \includegraphics[width=\linewidth]{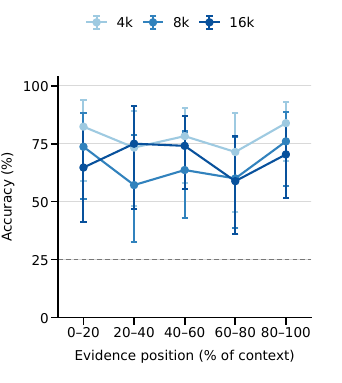}
    \caption{DeepSeek-V4-Pro}
    \label{fig:evidence-position-deepseek}
  \end{subfigure}
  \begin{subfigure}[t]{.32\textwidth}
    \centering
    \includegraphics[width=\linewidth]{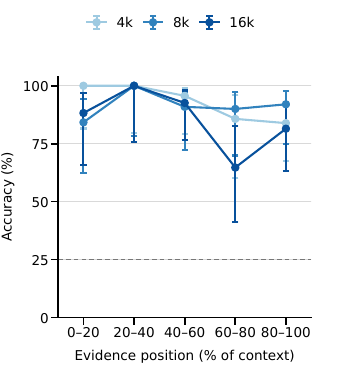}
    \caption{GLM-5.3}
    \label{fig:evidence-position-glm}
  \end{subfigure}
  \caption{\textbf{All three models use evidence in the middle of the context least.} We locate the supporting sentence of each BABILong QA1 example and group accuracy by its relative position in the context, at context lengths of 4k, 8k, and 16k. Error bars are 95\% Wilson intervals.}
  \label{fig:evidence-position-results}
\end{figure*}

We evaluate Jev, DeepSeek-V4-Pro, and GLM-5.3 on LongBench~\citep{bai-etal-2024-longbench}, BABILong~\citep{kuratov-etal-2024-babilong}, and NoLiMa-MC~\citep{modarressi-etal-2025-nolima}. LongBench examples are stratified by the \texttt{cl100k\_base} token length of the \texttt{state} field, while BABILong and NoLiMa-MC are evaluated under 4k, 8k, and 16k context-length settings.

\paragraph{Accuracy falls as the input grows, for decision models and LLMs alike.}
Decision accuracy declines with context length for all three models (Figure~\ref{fig:context-length-results}). On LongBench, Jev's accuracy falls from 93.5\% for examples in the 4--8k token range to 74.5\% for contexts around 20k tokens, and all three models lose accuracy beyond 16k tokens. A constrained output space fixes what the model must produce, not what it must read, so the bottleneck shifts entirely to locating the relevant evidence.

\paragraph{Evidence in the middle of the context is used least.}
Motivated by the lost-in-the-middle phenomenon~\citep{liu-etal-2024-lost}, we identify the sentence containing the gold supporting evidence in each BABILong QA1 example and partition its normalized character position into five equal-width intervals. After aggregating over the three context-length settings, all three models show the U-shaped pattern previously reported for generative LLMs: accuracy reaches its minimum when the supporting evidence appears near the middle of the context (Figure~\ref{fig:evidence-position-results}). Jev follows the same pattern, with its accuracy dropping to 80.4\% in the middle-position bin.

\subsection{Fast Local Decisions in Agentic Workflows}
\label{sec:agentic-workflows}

\begin{finding}
A \textbf{faster decision step} does not make a \textbf{faster agent}: errors in action selection \textbf{compound over an episode}, lowering success and lengthening the slowest runs.
\end{finding}

\begin{table}[htbp]
    \centering
    \caption{\textbf{With Jev selecting actions, the typical episode is faster, but fewer episodes succeed and the slowest ones take longer.} We replace the action-selection step of the $\tau$-bench tool-calling agent with Jev, keep the same base LLM for arguments and responses, and run each test task once. Pass is the episode success rate, Turns is the mean number of agent turns, and P50 and P95 are percentiles of episode wall-clock time.}
    \label{tab:tau-bench-agentic-snapshot}
    \small
    \setlength{\tabcolsep}{3pt}
    \renewcommand{\arraystretch}{1.1}
    \begin{tabular}{@{}llrrrr@{}}
    \toprule
    \textbf{Env.} & \textbf{System} & \textbf{Pass (\%)} & \textbf{Turns} & \textbf{P50 (s)} & \textbf{P95 (s)} \\
    \midrule

    \multirow{2}{*}{\textbf{Retail}}
    & Base (tool call)
    & \textbf{86.1} & \textbf{13.22} & 257.3 & \textbf{661.9} \\
    & Jev + LLM
    & 73.9 & 13.90 & \textbf{147.1} & 702.3 \\

    \midrule

    \multirow{2}{*}{\textbf{Airline}}
    & Base (tool call)
    & \textbf{58.0} & 13.68 & 277.8 & \textbf{795.2} \\
    & Jev + LLM
    & 48.0 & \textbf{12.35} & \textbf{242.6} & 1147.7 \\

    \midrule

    \multirow{2}{*}{\textbf{Overall}}
    & Base (tool call)
    & \textbf{77.6} & \textbf{13.36} & 267.1 & \textbf{695.5} \\
    & Jev + LLM
    & 66.1 & 13.44 & \textbf{160.6} & 816.0 \\

    \bottomrule
    \end{tabular}
\end{table}

To evaluate Jev in realistic agent workflows, we integrate it into the open-source $\tau$-bench tool-calling agent~\citep{yao2024tau} and test it in the Retail and Airline environments. The baseline uses DeepSeek-V4-Flash to jointly select tools and generate arguments. In the Jev condition, Jev selects an available tool or the \texttt{respond} action, while the same base model generates the corresponding arguments or user-facing response. Both systems use the same user simulator, temperature 0, environments, and reward checker. We report episode success (Pass$^1$), mean agent turns, and P50 and P95 wall-clock time. Appendix~\ref{app:tau-bench-agentic} details the matched protocol and per-turn workflows.

\paragraph{Jev shortens typical episodes but lowers task success.}
Jev + LLM succeeds on 66.1\% of episodes against 77.6\% for the baseline, and the drop is similar in Retail, 73.9\% against 86.1\%, and in Airline, 48.0\% against 58.0\% (Table~\ref{tab:tau-bench-agentic-snapshot}). Median episode time falls from 267.1\,s to 160.6\,s while mean turns stay nearly unchanged at 13.44 against 13.36. The tail moves the other way: P95 rises from 695.5\,s to 816.0\,s.

\paragraph{Single-step errors compound into the slow tail.}
The long-running episodes are concentrated among failures. Some failures begin when Jev selects an inappropriate next tool or action in a multi-step task, causing the trajectory to deviate from the goal; others begin after action selection, when the base LLM generates incorrect tool arguments. Both kinds of error trigger repeated lookups, recovery attempts, and retries, and some trajectories eventually time out. A faster decision step therefore speeds up successful trajectories, but each unreliable choice is paid for over the remaining turns of the episode.

\subsection{From Individual Decisions to Population Simulation}
\label{sec:social-simulation}

Social simulation is a natural application for decision models: every simulated person repeatedly makes a constrained choice, and a simulation needs many such choices at low cost. We test decision models at three levels of increasing scope. We first ask whether they can annotate people (Section~\ref{sec:jeval-interest-profile-agreement}), then whether they can simulate an individual's survey responses (Section~\ref{sec:jeval-sociobench}), and finally whether the decisions of many simulated individuals aggregate into population-level outcomes that match reality (Section~\ref{sec:electionsim}). Jev is evaluated at the annotation and individual levels, and an earlier checkpoint of \ourlarge{}, the open decision model built in Section~\ref{sec:self-distillation}, at the individual and population levels.

\subsubsection{Agreement in User Interest Profiling}
\label{sec:jeval-interest-profile-agreement}

\begin{finding}
Jev \textbf{falls short of general-purpose LLMs} in user-interest profiling, so applying it directly to social-science annotation \textbf{calls for caution}.
\end{finding}

Annotating people, for example by labeling the interests expressed in their social-media activity, is one of the most basic tasks in computational social science and the first step in building simulated agents. Each annotation is a choice among predefined categories, which makes it a natural fit for a decision model. We therefore first test whether Jev can serve as such an annotator by comparing the interest profiles it assigns to users with those of a general-purpose LLM and of a reference profiler.

\paragraph{Experimental Setup.}
We profile a held-out cohort of social-media users as multi-label sets over 20 predefined top-level interest categories, using Jev~1.13 and Qwen3.8-27B in non-thinking and thinking modes, and compare each against a frozen Qwen3-Max production profile on the 898 accounts for which all systems return valid outputs. We report the mean per-user Jaccard similarity between interest sets and the rate of exact set matches. None of the outputs is treated as ground truth.

\begin{table}[htbp]
\centering
\small
\caption{\textbf{Jev agrees less with the reference profiler than Qwen3.8-27B does.} We profile 898 social-media accounts as multi-label sets over 20 interest categories and compare each system with a frozen Qwen3-Max reference. Jaccard similarity is computed per user and averaged, and an exact match requires identical sets of categories.}
\label{tab:interest-profile-agreement}
\begin{tabular}{lrr}
\toprule
Compared System & Mean Jaccard $\uparrow$ & Exact Match $\uparrow$ \\
\midrule
Jev~1.13 & 53.88\% & 15.03\% \\
Qwen3.8-27B non-thinking & 65.24\% & 25.06\% \\
Qwen3.8-27B thinking & \textbf{65.86\%} & \textbf{28.84\%} \\
\bottomrule
\end{tabular}
\end{table}

\paragraph{Jev agrees less with a reference profiler than a general-purpose LLM does.}
Jev's interest profiles reach a mean Jaccard similarity of 53.88\% with the reference, against 65.24\% and 65.86\% for non-thinking and thinking Qwen3.8-27B, and match it exactly for 15.03\% of users, against 25.06\% and 28.84\% (Table~\ref{tab:interest-profile-agreement}). Thinking adds less than one Jaccard point, and no system reproduces the reference exactly for most users. Because Qwen3.8-27B and the reference belong to the same model family, part of its advantage may reflect shared preferences rather than more accurate profiles.

\subsubsection{Individual-Level Social Survey Prediction}
\label{sec:jeval-sociobench}

\begin{finding}
Decision models \textbf{predict individual survey responses} on par with generative LLMs at \textbf{a much lower cost}.
\end{finding}

Annotating people is not yet simulating them. We next ask whether a decision model can simulate an individual: given a respondent's demographic profile, it must predict how that person answers a sociological survey question, a standard testbed for LLM-based simulation of individuals. Each prediction is a choice among the answer options, so the task directly measures how well a decision model represents individual opinions.

\begin{table*}[t]
  \centering
  \begingroup
  \small
  \setlength{\tabcolsep}{3pt}
  \renewcommand{\arraystretch}{1.12}
  \caption{\textbf{Decision models predict individual survey responses about as accurately as generative LLMs at a fraction of their latency.} We predict responses from demographic profiles on SocioBench, using the same 277,500 respondent--question pairs across ten domains for all systems. \ourlarge{} is close to thinking DeepSeek-V4-Flash, and Jev to its non-thinking mode. Latency is the mean time in seconds per item.}
  \label{tab:jeval-sociobench}
  \resizebox{\textwidth}{!}{%
    \begin{tabular}{@{}l@{\hspace{8pt}}rr@{\hspace{8pt}}rr@{\hspace{8pt}}rr@{\hspace{8pt}}rr@{\hspace{8pt}}rr@{\hspace{8pt}}rr@{}}
      \toprule
      & \multicolumn{2}{c}{\textbf{Jev}} & \multicolumn{4}{c}{\textbf{Qwen3.5-4B}} & \multicolumn{4}{c}{\textbf{DeepSeek-V4-Flash-0731}} & \multicolumn{2}{c}{\textbf{Ours: \ourlarge{}}} \\
      \cmidrule(lr){2-3}\cmidrule(lr){4-7}\cmidrule(lr){8-11}\cmidrule(lr){12-13}
      & \multicolumn{2}{c}{Native} & \multicolumn{2}{c}{Non-thinking} & \multicolumn{2}{c}{Thinking} & \multicolumn{2}{c}{Non-thinking} & \multicolumn{2}{c}{Thinking (\texttt{high})} & \multicolumn{2}{c}{Native} \\
      \cmidrule(lr){2-3}\cmidrule(lr){4-5}\cmidrule(lr){6-7}\cmidrule(lr){8-9}\cmidrule(lr){10-11}\cmidrule(lr){12-13}
      \textbf{Domain} & \textbf{ACC} $\uparrow$ & \textbf{Latency} $\downarrow$ & \textbf{ACC} $\uparrow$ & \textbf{Latency} $\downarrow$ & \textbf{ACC} $\uparrow$ & \textbf{Latency} $\downarrow$ & \textbf{ACC} $\uparrow$ & \textbf{Latency} $\downarrow$ & \textbf{ACC} $\uparrow$ & \textbf{Latency} $\downarrow$ & \textbf{ACC} $\uparrow$ & \textbf{Latency} $\downarrow$ \\
      \midrule
      Citizenship & 43.00 & 0.478 & 39.46 & 6.999 & 41.84 & 102.333 & 43.60 & 268.818 & \textbf{45.41} & 302.180 & \underline{45.09} & 8.412 \\
      Environment & 34.85 & 0.480 & 30.30 & 6.832 & 33.62 & 101.452 & 35.51 & 234.424 & \textbf{38.29} & 283.856 & \underline{36.21} & 8.919 \\
      Family & 37.23 & 0.493 & 29.17 & 6.728 & 33.55 & 106.760 & 40.31 & 260.248 & \textbf{42.94} & 328.831 & \underline{40.77} & 8.490 \\
      Health & 35.44 & 0.494 & 30.46 & 6.816 & 35.03 & 97.687 & 34.83 & 250.956 & \underline{37.44} & 292.322 & \textbf{39.99} & 8.623 \\
      National Identity & 37.70 & 0.504 & 31.69 & 7.473 & 34.30 & 98.672 & 35.93 & 249.620 & \underline{37.92} & 182.197 & \textbf{38.86} & 8.618 \\
      Religion & 39.69 & 0.494 & 34.77 & 8.319 & 38.39 & 101.512 & 39.90 & 81.061 & \textbf{42.42} & 178.482 & \underline{40.65} & 6.131 \\
      Role of Government & \underline{38.66} & 0.519 & 34.97 & 7.513 & 36.17 & 97.840 & 38.02 & 74.523 & \textbf{39.86} & 189.612 & 38.39 & 8.321 \\
      Social Inequality & 36.03 & 0.486 & 29.05 & 7.115 & 31.34 & 103.303 & 34.83 & 104.599 & \underline{36.75} & 147.013 & \textbf{36.82} & 8.711 \\
      Social Networks & 36.82 & 0.497 & 34.53 & 7.117 & 37.72 & 110.469 & 37.54 & 112.630 & \textbf{39.96} & 115.594 & \underline{38.51} & 8.898 \\
      Work Orientations & 36.27 & 0.506 & 32.36 & 6.164 & 36.70 & 113.625 & 37.33 & 113.121 & \underline{41.93} & 111.529 & \textbf{42.39} & 8.668 \\
      \midrule
      \textbf{Overall} & 37.69 & 0.496 & 32.87 & 7.124 & 36.04 & 103.394 & 37.90 & 173.999 & \textbf{40.41} & 211.649 & \underline{39.87} & 8.357 \\
      \bottomrule
    \end{tabular}%
  }
  \endgroup
\end{table*}

\paragraph{Experimental Setup.}
We predict individual survey responses from respondents' demographic profiles on SocioBench~\citep{jeval_sociobench}, using the same 277,500 respondent--question pairs across ten domains for all systems. We compare Jev~1.13, \ourlarge{}, and two generative LLMs, Qwen3.5-4B and DeepSeek-V4-Flash-0731, each in non-thinking and thinking modes. The LLMs are served with vLLM on eight H100 GPUs at temperature 0.5, with an 8,192-token budget for reasoning and answer, and \ourlarge{} uses its native option-probability readout.

\paragraph{At the individual level, decision models match LLMs at a fraction of their latency.}
\ourlarge{} reaches 39.87\% overall accuracy, 0.54 points below the 40.41\% of the best configuration, thinking DeepSeek-V4-Flash-0731, at a recorded mean latency of 8.36\,s against 211.65\,s (Table~\ref{tab:jeval-sociobench}). It ranks among the top two in nine of the ten domains and leads four of them. Jev performs on par with non-thinking DeepSeek-V4-Flash-0731, at 37.69\% against 37.90\%, with the lowest latency of all systems at 0.50\,s. Thinking improves both LLMs in every domain, but only by 2.5--3.2 points overall, while the thinking configurations take 103--212\,s per item, a cost that a simulation pays once for every member of the population.

\subsubsection{Population-Level Election Simulation}
\label{sec:electionsim}

\begin{finding}
Decision models make \textbf{population-scale simulation affordable} and call \textbf{state-level outcomes} as well as generative voters, but they still exhibit \textbf{LLM-like biases}, such as a partisan lean.
\end{finding}

Finally, we move from individuals to a population. In macro-level simulations such as ElectionSim~\citep{Zhang2024ElectionSim}, the quantity of interest is not any single decision but the aggregate of many: hundreds of thousands of simulated voters whose choices add up to state-level outcomes. We test whether the individual decisions of a decision model, aggregated in this way, agree with the real outcome of the 2024 U.S. presidential election.

\paragraph{Experimental Setup.}
We use \ourlarge{} to drive the voter agents of ElectionSim, which simulates the election state by state with agents built from real social-media users. We sample about 330,000 voters, a ratio of 1/1,000, and describe each by state, five demographic attributes, and recent social-media posts; ElectionSim's area attribute is unavailable to us. Each voter chooses between Harris and Trump under both option orders, and a state's predicted Democratic two-party share is the voter-weighted mean of $P(\text{Harris})$. We compare with the archived output of the original pipeline at the same ratio, in which GPT-4o-mini writes each vote. Following ElectionSim, we count correctly called states and battleground states and measure the error of the predicted vote share; Appendix~\ref{app:electionsim-states} defines these metrics and reports them in full.
\begin{figure*}[t]
  \centering
  \includegraphics[width=\textwidth]{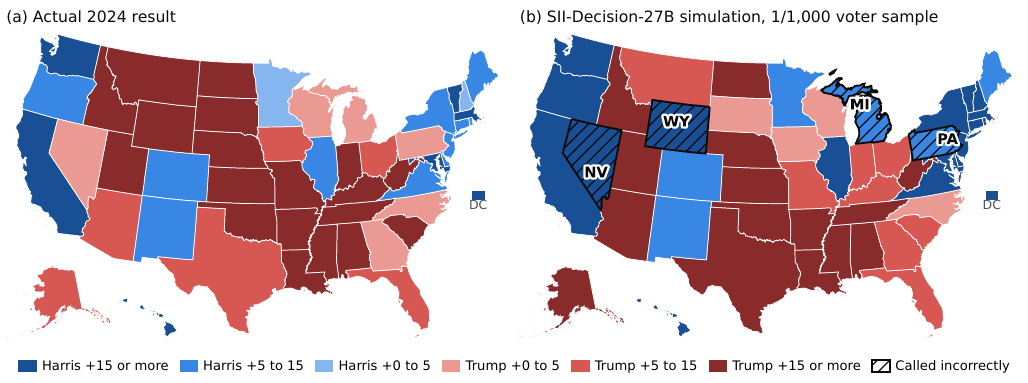}
  \caption{\textbf{SII-Decision-27B calls 47 of 51 states correctly, and all four misses are Trump-won states called for Harris.} We simulate the 2024 U.S. presidential election in ElectionSim, with SII-Decision-27B driving about 330,000 voter agents sampled at a ratio of 1/1,000. (a) Actual two-party margin. (b) Simulated margin. Appendix~\ref{app:electionsim-states} lists the per-state results.}
  \label{fig:electionsim-map}
\end{figure*}

\paragraph{Aggregated decisions call the states but misestimate the shares.}
\ourlarge{} calls 47 of 51 states and 12 of 15 battleground states, one more of each than the original pipeline (Figure~\ref{fig:electionsim-map}), but its vote-share errors are about twice as large (Table~\ref{tab:electionsim} in Appendix~\ref{app:electionsim-states}). A state call depends only on which side of 50\% the averaged $P(\text{Harris})$ falls, whereas the vote share depends on the level of every voter's $P(\text{Harris})$, so the two can diverge in the same way that choosing and estimating diverge in Section~\ref{sec:math-probability}. The largest errors occur in sparsely sampled states such as Wyoming and Alaska, and all four misses are Trump-won states called for Harris.

\paragraph{Decision models inherit the partisan lean of LLM voters.}
Every run overestimates the Democratic share. \ourlarge{}, the original pipeline, and two preliminary runs of Jev through its hosted API all overestimate Harris's share, by two to six points on average; of the Jev runs, one uses the ElectionSim setting at a ratio of 1/100,000, and the other uses demographics only on the SocioVerse voter pool~\citep{Zhang2025SocioVerse}. In the demographics-only run, every misclassified state is a Trump-won state called for Harris despite thousands of simulated voters per state, and liberal Republicans defect to Harris about twice as often as conservative Democrats defect to Trump. ElectionSim reports the same overestimation for generative agents~\citep{Zhang2024ElectionSim}, so moving from generation to decision preserves the lean.

\paragraph{Decisions make population-scale simulation cheap.}
Each voter costs two forward passes and no generated tokens, and $P(\text{Harris})$ yields expected vote shares directly. Reducing the vote-share errors and the inherited lean is what remains before decision models can replace generative voter agents.

\section{Distilling Reasoning into Decision}
\label{sec:self-distillation}

\begin{finding}
An open-weight LLM becomes a general decision model through its \textbf{first-token readout}, and distilling its \textbf{own post-reasoning distribution} into that readout brings it to Jev's level in a single forward pass.
\end{finding}

Jev is available only through a hosted API, and its training recipe is not public. The preceding sections therefore study general decision models from the outside. This section asks the complementary question:
\begin{center}
\textit{Can such a model be rebuilt from an open-weight LLM?}
\end{center}


We answer in two steps. First, an LLM already contains a decision model, and its decisions can be read from the first answer token: the next-token distribution at the answer position, restricted to the labels of the valid options, yields choice, noul, and score decisions from a single forward pass. We call this the \textbf{First-Token Decision Readout} (Section~\ref{sec:first-token-readout}). Second, we distill the answer distribution that the model reaches at the end of its reasoning chain into its first answer token, and call this the \textbf{Reasoning-to-Readout Self-Distillation} (Section~\ref{sec:logit-self-distillation}). Figure~\ref{fig:reproduction-framework} illustrates both steps. Applied to Qwen3.5-4B and Qwen3.8-27B, it yields the open models SII-Decision-4B and SII-Decision-27B, and SII-Decision-27B performs on par with Jev on JEVal at a fraction of its latency (Section~\ref{sec:self-distillation-results}).

\subsection{Reading Decision from the First Token}
\label{sec:first-token-readout}

\begin{figure*}[t]
  \centering
  \includegraphics[width=\textwidth]{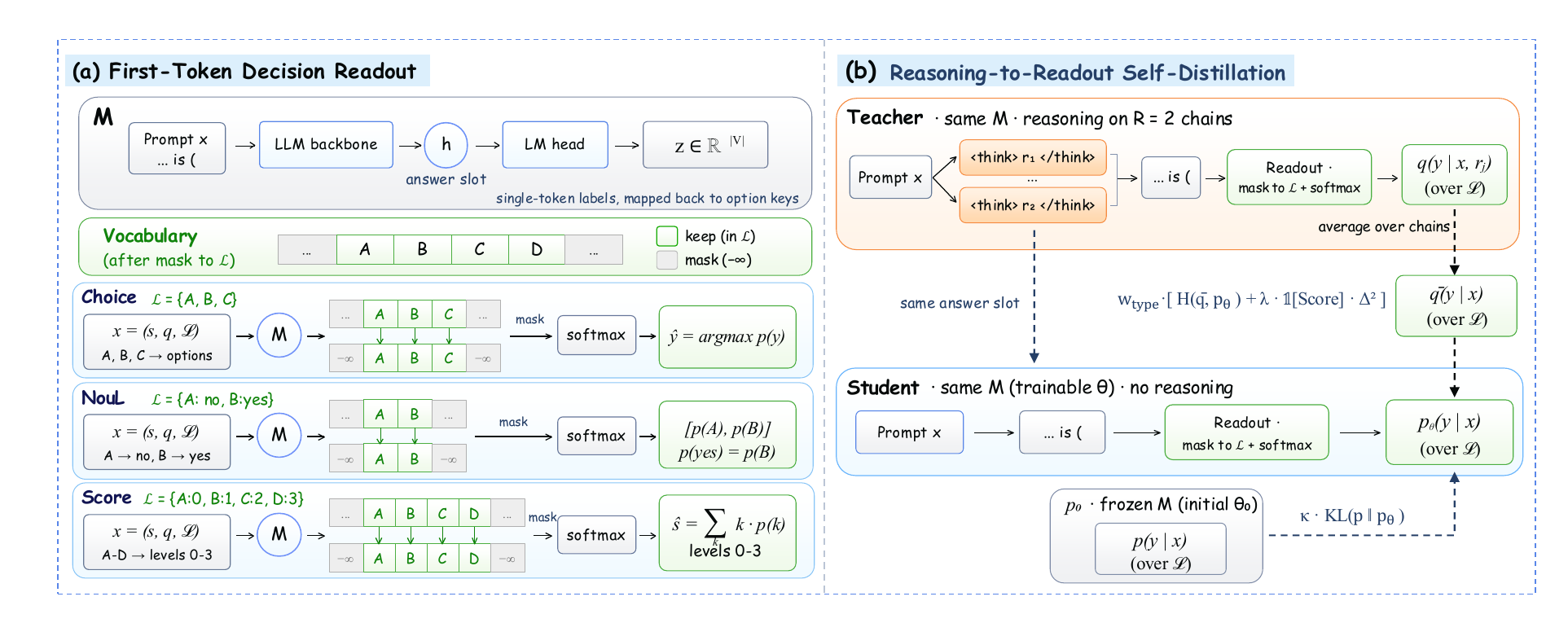}
  \caption{Overview of how we build a decision model from an open-weight LLM. (a) First-Token Decision Readout: the next-token distribution at the answer position is restricted to the labels of the valid options. One forward pass gives the selected option for \texttt{choice}, the probability of true for \texttt{noul}, and the expected level for \texttt{score}. (b) \ourmethod{}: the model first reasons, and its readouts at the end of the reasoning chains are averaged into a target distribution. The same model without reasoning is trained to match this target at its first answer token, with a KL term that keeps it close to the readout of the frozen original model.}
  \label{fig:reproduction-framework}
\end{figure*}

To rebuild a decision model from an LLM, we first need the LLM to produce what a decision model produces: a probability for every valid option, from a single forward pass. An LLM normally gives its decision by writing it out, as the generative baselines of Section~\ref{sec:experimental-setup} do, which costs decoding steps and yields probabilities only as numbers the model writes. The decision is available earlier, however. Once the prompt ends where the answer is due, the model's next-token distribution already assigns a probability to the label of each option. The First-Token Decision Readout takes this distribution as the decision: we read the next-token distribution at the first answer position and restrict it to the labels of the valid options (Figure~\ref{fig:reproduction-framework}a).

Concretely, we render each instance $x=(s,i,c)$ as plain text that lists the state, the instruction, and the candidate options labeled (A), (B), \ldots, and ends with the answer prefix ``\texttt{The correct answer is (}'', without a chat template (Appendix~\ref{app:self-distillation-template}). Options are labeled A to Z and then AA, AB, and so on, keeping only labels that the tokenizer encodes as a single token; the first 255 such labels cover every candidate set in JEVal (Section~\ref{sec:data-processing}) and are single tokens for both backbones, so each option corresponds to one row of the output layer. Let $h_\theta(x)$ denote the final-layer hidden state at the last prefix position, $w_\ell$ the output embedding of token $\ell$, and $\ell(y)$ the label token of option $y$. The readout is the next-token softmax restricted to the labels of the valid options:
\begin{equation}
p_\theta(y \mid x)
=
\frac{\exp\!\big(w_{\ell(y)}^{\top} h_\theta(x)\big)}
{\sum_{y' \in \mathcal{Y}(x)} \exp\!\big(w_{\ell(y')}^{\top} h_\theta(x)\big)},
\qquad y \in \mathcal{Y}(x).
\label{eq:first-token-readout}
\end{equation}
The output-layer rows of the option labels thus act as a decision head over $\mathcal{Y}(x)$, with no new parameters and no generated tokens. All three JEVal formats share this head. For \texttt{choice} questions, the most probable label is mapped back to its option key. For \texttt{noul} questions, the two outcomes are presented as options and the model returns the probability of \texttt{true}. For \texttt{score} questions, every rating level is presented as an option and the model returns the expected level $\hat{s}=\sum_{k} v_k\, p_\theta(k \mid x)$, where $v_k$ is the value of level $k$. These are the same typed outputs that Jev returns, so the readout can replace Jev wherever Jev is used, and because it needs one prefill pass and no decoding, its latency depends only on the backbone and the input length.

Even without training, this readout of Qwen3.8-27B comes within a few points of Jev. The pretrained output layer is also the right decision head to use: in preliminary experiments, training only the label rows of the output layer barely changed accuracy, and replacing them with a newly initialized decision head lowered it by several points. The readout therefore turns an open-weight LLM into a general decision model as it is, with its pretrained head intact.

\subsection{Learning from the Model's Own Reasoning}
\label{sec:logit-self-distillation}

The readout gives the model's decision before it has done any reasoning. The same model can also reason first and then answer, and we observe that this changes its decisions considerably, and almost always for the better. The change is large: on JEVal, thinking adds five to ten points to each open LLM we run in both modes (Section~\ref{sec:main-results}), and on the exact-computation and probability tasks of Section~\ref{sec:math-probability} it lifts Qwen3.8-27B from about half of the items correct to nearly all of them. The change also has a clear direction. Reasoning repairs the readout rather than disrupting it: on the public JevBench questions, thinking fixes all but one readout error of Qwen3.8-27B without changing any correct answer, and on our other development sets it corrects far more answers than it breaks in all three formats.

What the readout lacks is therefore not knowledge but the reasoning that brings it out: after reasoning, the model reaches decisions that its readout misses, and these decisions are its own. The model can therefore act as its own teacher, with no need for labeled answers or a stronger teacher. \ourmethod{} builds on this: it distills the answer distribution that the model reaches at the end of its reasoning chain into its first answer token, so that a single forward pass arrives where the chain would have led (Figure~\ref{fig:reproduction-framework}b).

\paragraph{Teacher: the distribution at the end of reasoning.}
For each training question, the model generates two independent reasoning chains with thinking enabled. For each chain $r$, we append the reasoning segment and the answer prefix to the prompt, discard the answer text written after the reasoning, and apply Eq.~\eqref{eq:first-token-readout} at the end of the chain to obtain $q(y \mid x, r)$; for Qwen3.8-27B, the top option of this readout agrees with the model's written answer in every chain we checked. The teacher distribution $\bar{q}(y \mid x)$ averages $q$ over the two chains. The target is the full distribution rather than the final answer, because downstream uses such as the population simulation of Section~\ref{sec:electionsim} consume probabilities, not only argmax decisions. Gold labels never enter the target: we neither filter questions by whether the teacher is correct nor replace its distribution with an annotation, and a chain without a valid readout falls back to the model's readout without thinking.

\paragraph{Student: the same model without reasoning.}
Teacher and student share the model, the readout, and the label space, and differ only in whether a reasoning chain precedes the answer slot. The student sees only the question and learns to match, at its first answer token, the teacher's distribution at the reasoning endpoint:
\begin{equation}
\mathcal{L}(\theta)
=
\sum_{i} w_{t_i}\Big[
H\big(\bar{q}_i,\, p_\theta(\cdot \mid x_i)\big)
+ \lambda\, \mathbf{1}[t_i=\texttt{score}]\, \Delta_i^{2}
\Big]
+ \sum_{i} \kappa_{t_i}\, \mathrm{KL}\big(p_0(\cdot \mid x_i)\,\big\|\, p_\theta(\cdot \mid x_i)\big),
\label{eq:self-distillation-loss}
\end{equation}
where $H$ is the cross-entropy, $t_i$ is the question format, and $w_t$ are inverse-frequency weights that balance the formats. For \texttt{score} questions, $\Delta_i$ is the difference between the student's and the teacher's expected levels divided by the width of the rating scale, with $\lambda=0.5$. The KL term anchors the student to the frozen readout $p_0$ of the original model to limit forgetting, with $\kappa=0.05$, raised to $0.2$ for hard-labeled \texttt{noul} questions. Because the student must reach the post-reasoning decision without the chain, the computation of the chain has to be absorbed into a single forward pass; this is how the model internalizes its reasoning. The first term is knowledge distillation~\citep{Hinton2015Distilling} restricted to the valid options; unlike prior work that distills explicit reasoning into direct answers~\citep{Deng2023ImplicitCoT,Yu2024System2}, the target is only the decision distribution at the reasoning endpoint, matched at the first answer token.

We update the attention and MLP projections of all language layers at full rank, with LoRA adapters~\citep{Hu2022LoRA} on the same modules as a lighter alternative; the embeddings and the output layer stay frozen, so the label rows used by the readout are unchanged and the trained model has the inference cost of the untrained readout. Training runs for a single epoch, with the checkpoint selected on a frozen development set (Appendix~\ref{app:self-distillation-training}). Qwen3.5-4B and Qwen3.8-27B follow exactly the same recipe, each learning from its own reasoning with its own readout as the anchor $p_0$.

\paragraph{Training data: any question with options.}
Since no gold label enters the target, any question with a well-defined option set can serve as training data, and the design question becomes which questions to use. We select them by the capabilities a general decision model is expected to cover. The base pool draws on public decision datasets in eight groups: language understanding, general knowledge, commonsense, logic, English and Chinese medicine, rule-based decisions and computation, and safety and subjective judgment. Development performance rises steadily with the number of such questions, so for \ourlarge{} we extend the pool with science, quantitative reasoning, professional, and factual-boundary questions from further public sources, for 39,279 questions in total. Beyond this point the gain stalls, because additional questions bring little to learn: questions that the model writes itself tend to be easy for it, and it cannot reliably check whether harder ones are correct (Appendix~\ref{app:self-distillation-scaling}).

\subsection{Results}
\label{sec:self-distillation-results}

Several open decision models have been released to reproduce Jev. We compare \oursmall{} and \ourlarge{} with these reproductions and with Jev itself (Table~\ref{tab:overall-results}). Accuracy is the pooled JEVal accuracy of Section~\ref{sec:main-results}, and latency is the mean over inputs of up to 1K tokens, which cover most of JEVal, measured on a single H100 GPU for the open models; Jev's latency is that of its hosted API and includes network transport.

\begin{figure}[t]
\centering
\includegraphics[width=0.9\textwidth]{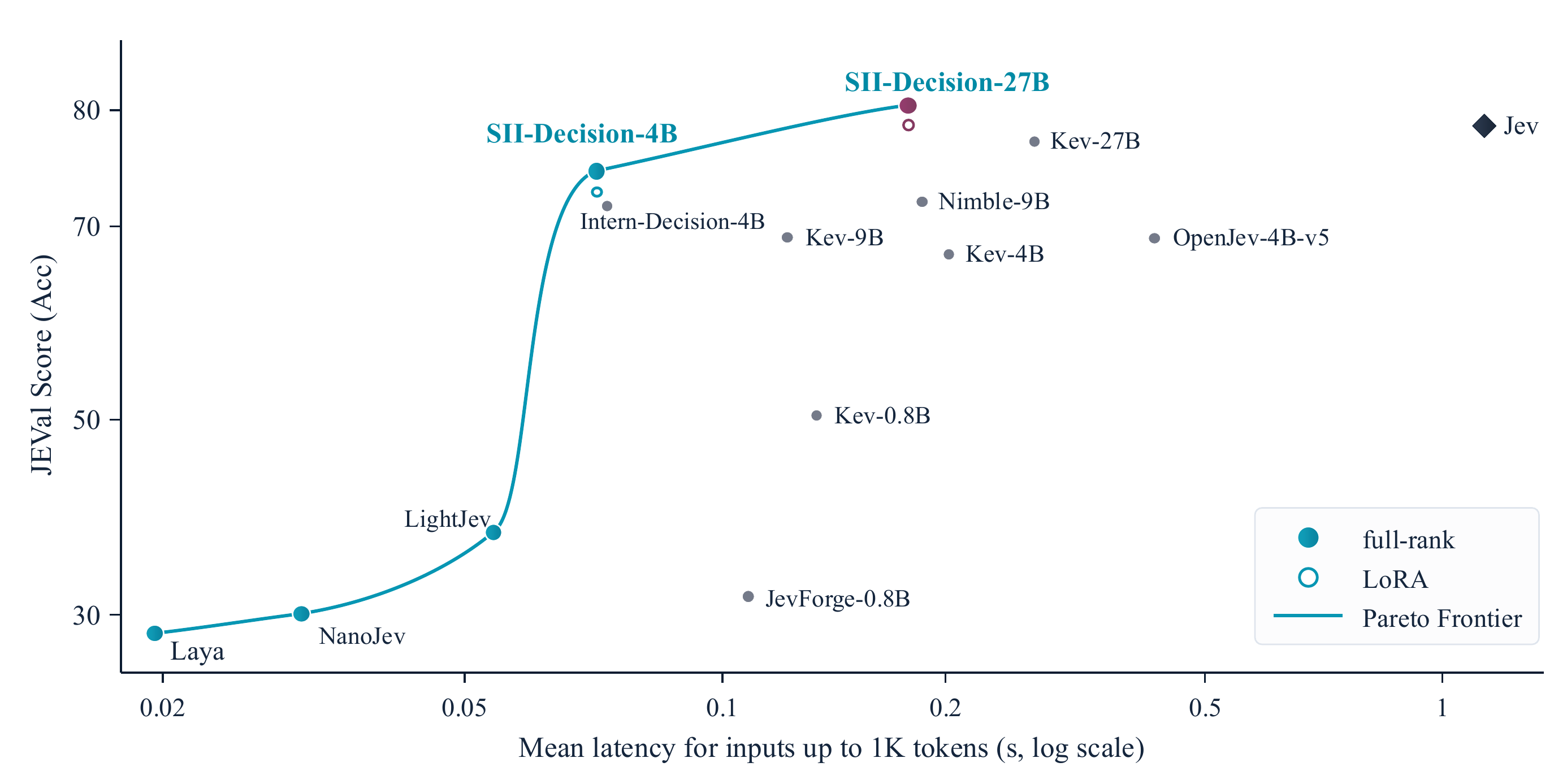}
\caption{\textbf{SII-Decision-4B and SII-Decision-27B lie on the accuracy--latency Pareto frontier, and SII-Decision-27B exceeds Jev in accuracy at a fraction of its latency.} We compare open reproductions of Jev, and Jev itself, on pooled JEVal accuracy and on mean latency for inputs of up to 1K tokens. Open models are measured on a single H100 GPU, and Jev through its hosted API, including network transport.}
\label{fig:self-distillation-frontier}
\end{figure}

\paragraph{\ourname{} sits on the accuracy--latency frontier.}
Both models lie on the Pareto frontier of Figure~\ref{fig:self-distillation-frontier}, and \ourlarge{} sets its upper end: it is the most accurate decision model on JEVal, nearly two points above Jev, and answers in a fraction of Jev's time on typical inputs. \oursmall{} is the most accurate of the models that answer within a tenth of a second, ahead of Intern-Decision-4B at a similar latency. On the same GPU, their median latency is about 60\,ms for \oursmall{} and 110\,ms for \ourlarge{}. The full-rank and LoRA versions land within a point of each other at both scales, so the gain comes from what the model learns, its own post-reasoning distribution, rather than from how its weights are updated. The gain carries over to newly constructed questions that played no role in training or model selection, where \ourlarge{} improves on the untrained readout by about seven points. Calibration follows the strength of the teacher: \ourlarge{} has the lowest ECE and $\mathrm{Brier}_{\mathrm{raw}}$ of all systems in Table~\ref{tab:overall-results}, whereas \oursmall{} is less calibrated than Intern-Decision-4B.

\paragraph{Distillation transfers the weighing of options, not intermediate computation.}
Relative to its untrained readout, \ourlarge{} gains most on label-selection benchmarks such as intent recognition, harmful-request judgment, and emotion classification, and it matches or exceeds Jev on more than half of the benchmarks. On NLU-Eval-68, CLINC150, AgentHarm, and GoEmotions, the distilled readout recovers most of the gain that thinking brings to Qwen3.8-27B. In these tasks, reasoning mainly weighs the evidence for each option, and the result of that weighing fits into the first-token distribution. On reasoning-intensive benchmarks the picture reverses: on GPQA and MMLU-Pro, thinking lifts Qwen3.8-27B beyond Jev's level, yet distillation recovers a quarter of that gain or less. Such reasoning computes intermediate results, as in multi-step arithmetic, solution verification, or aggregation over many records, and a single forward pass does not reproduce that computation even when trained on its outcome. This matches Section~\ref{sec:math-probability}, where Jev lands near exact numerical answers but rarely on them. The teacher sets a second limit: on TruthfulQA, thinking Qwen3.8-27B stays well below Jev, so there is little to transfer, and distillation lowers accuracy.

\section{Future Directions}

Our results show that current decision models offer broad capabilities and substantial efficiency gains, while their limits, scalability, and integration with generative models remain open questions. Therefore, we further discuss several open questions:

\paragraph{Scaling Toward Collective Intelligence.}
General intelligence has largely been provided by generative LLMs, whose high inference latency and computational cost limit large-scale parallel deployment, particularly in settings that require continuous interaction among many agents, such as large-scale social simulation, multi-agent coordination, and collective behavior modeling. Fast general decision models make it possible to compose broadly capable decision-making components at substantially lower cost, opening a new avenue for studying collective behavior and emergent intelligence at scale.

\paragraph{The Capability Frontier of General Decision Models.}
Our results show that finite output spaces do not restrict decision tasks to simple pattern matching; general decision models can already handle knowledge-intensive, reasoning, long-context, and agentic tasks. However, we also observe a remaining capability gap relative to state-of-the-art LLMs, and it remains unclear how far this  can be extended. Characterizing this progression is an important direction for understanding the limits of general decisional intelligence.

\paragraph{System-Level Composition of Heterogeneous Intelligence.}
Future AI systems may benefit from combining components with different capability rather than relying on a single model for all computation. Generative LLMs are well suited for open-ended reasoning, planning, and complex generation, while fast decision models can handle large volumes of frequent and structured decisions. Combining such complementary components offers a natural path toward balancing system capability, inference latency, and computational cost.

\section{Conclusion}

We presented a systematic study of \textbf{general decisional intelligence}, from isolated decisions to their behavior when composed in larger systems. We introduced \textbf{JEVal}, a bilingual benchmark comprising 11,257 instances from 36 datasets across 10 domains, and found that current general decision models can approach strong thinking LLMs on broad decision tasks, especially when decisions can be resolved from available evidence, while remaining weaker in specialist knowledge, faithful uncertainty estimation, exact computation, and long-context evidence use. Moving beyond static benchmarks, we further showed that strong local decision performance does not directly translate into reliable system behavior: in interactive agent workflows, faster local decisions coincided with lower task success, while the trajectory audit identified action errors and integration-level empty responses without isolating a dominant cause. In large-scale social simulation, efficient decision inference enables scalable population modeling but remains vulnerable to profiling errors, inaccurate aggregate estimates, and systematic bias. Finally, we showed that general decision models can be constructed from open-weight LLMs through first-token decision readout and \textbf{Reasoning-to-Readout Self-Distillation}, yielding \textbf{SII-Decision-4B} and \textbf{SII-Decision-27B} with competitive single-pass performance.

\section{Limitations}

\paragraph{Broader and more complex agentic evaluation.}
Our current analysis evaluates general decision models in two representative agentic settings: tool-using agents and social simulation. While these experiments provide initial evidence of how decision models behave within larger agent systems, they cover only a limited range of workflows and interaction patterns. Future work should extend evaluation to broader and more complex agentic settings, including richer tool environments and multi-agent coordination, to better understand how general decision models perform at the system level.

\paragraph{Broader model coverage.}
Our current evaluation covers only a limited set of general decision models, with Jev serving as the primary representative. Future work should evaluate a broader range of open- and closed-source decision models and LLMs across different model families, scales, and training paradigms to examine whether the observed capability and efficiency patterns generalize more broadly.

\paragraph{Training and construction of general decision models.}
Beyond evaluating existing models, we explore an open-source training approach based on \ourmethod{} as an initial attempt to construct general decision models. This represents only one possible training paradigm. Future work should investigate dedicated training objectives, alternative model architectures, broader training data, and more scalable training strategies for improving general decisional intelligence.

\bibliographystyle{assets/plainnat}
\bibliography{ref}

\clearpage
\appendix
\appendix

\section{Ethics}
\label{app:ethics}

JEVal includes tasks that may involve personal preferences, demographic attributes, harmful requests, or stereotyped scenarios, raising concerns about privacy, harmful content, and inherited dataset bias. To mitigate these risks, we do not reproduce individual participant profiles or demographic records, and sensitive examples are included solely for evaluating model behavior rather than endorsing the assumptions or labels in the source data. We also acknowledge that biases may persist through data collection and annotation, and therefore interpret results only within the scope of the sampled tasks rather than as representative evidence about real-world populations or settings.

\section{More Benchmark Details}
\label{app:more-benchmark-details}

\subsection{Datasets by Domain}
\label{app:datasets-by-domain}

We organize the source datasets by application domain. The final evaluation set contains 11,257 examples from 36 source datasets in ten domains; datasets with multiple retained task configurations contribute a total of 58 evaluation subtasks. We describe the source task, its conversion to the JEVal decision interface, and the split and sample count used in each case. Split names below refer to the labeled source partitions used to construct the evaluation set. Each item states its resulting JEVal decision format or formats.

\paragraph{General knowledge and reasoning.}
\begin{itemize}
\item \textbf{AI2 ARC} contains science examination questions collected in Challenge and Easy subsets \citep{jev-arc}. We place the question in the state, ask the model to select an answer, and use the original options and answer key as the criteria and label of a \texttt{choice} task. We sample 200 examples from the test split of each subset.

\item \textbf{AIDA-TestC} consists of news articles with mentions linked to knowledge-base entities \citep{jev-aida}. The article and marked mention form the state, while entity linking becomes a \texttt{choice} decision among the gold entity and sampled distractors. We take 200 examples from Test-C.

\item \textbf{BoolQ} pairs naturally occurring yes/no questions with supporting passages \citep{jev-boolq}. We retain the passage and question and map the answer to a \texttt{noul} judgment. The evaluation set contains 200 examples from the labeled train split.

\item \textbf{C-Eval} evaluates Chinese knowledge across academic and professional subjects using multiple-choice questions \citep{jev-ceval}. We preserve the question stem, options, and verified answer as a \texttt{choice} task and sample 200 examples from its test split.

\item \textbf{CLINC150} contains user utterances labeled with 150 in-scope intents and an out-of-scope class \citep{jev-clinc}. We present the utterance as state and the 151 intent labels as the decision space of a \texttt{choice} task, with 200 examples sampled from the test split.

\item \textbf{CMMLU} provides Chinese multiple-choice questions spanning many knowledge domains \citep{jev-cmmlu}. Each question becomes a state with its answer options as criteria and the supplied answer as the label of a \texttt{choice} task; we sample 200 test examples.

\item \textbf{FEVER} contains claims and Wikipedia evidence annotated for factual verification \citep{jev-fever}. We use the evidence as state and ask whether it supports the claim, retaining the supported and refuted cases that admit a \texttt{noul} judgment. We sample 200 examples from the test split.

\item \textbf{GoEmotions} annotates Reddit comments with fine-grained emotion labels \citep{jev-goemotions}. We retain single-label examples and ask the model to make a \texttt{choice} decision over the 28-label emotion inventory. We sample 200 examples from the test split.

\item \textbf{GPQA} contains expert-written questions designed to distinguish domain expertise from plausible incorrect answers \citep{jev-gpqa}. We convert the Diamond subset to four-option \texttt{choice} decisions, retaining its question, candidate answers, and gold answer. All 198 available test examples are included.

\item \textbf{MMLU-Pro} extends multidisciplinary knowledge evaluation with more challenging, often reasoning-intensive multiple-choice questions \citep{wang-etal-2024-mmlu-pro}. We retain the question and options in the \texttt{choice} format and sample 200 examples from the labeled train split.

\item \textbf{NLU Evaluation Data} collects utterances across 68 intents in 18 scenarios \citep{jev-nlu68}. We use each utterance as state and the intent inventory as criteria for a \texttt{choice} decision, selecting 200 examples from the source collection, which is represented as a test split in our converted data.

\item \textbf{SciFact} pairs scientific claims with evidence abstracts and veracity annotations \citep{jev-scifact}. The abstract is the state and the claim is assessed through a \texttt{noul} support judgment; we retain examples with support or contradiction labels. We sample 200 labeled dev examples because the released test questions do not provide gold labels.
\end{itemize}

\paragraph{Medicine.}
\begin{itemize}
\item \textbf{CMB} collects Chinese medical examination questions across clinical specialties \citep{jev-cmb}. We retain single-answer questions, map their stems and answer options to the \texttt{choice} interface, and join the released answer key where needed. We sample 200 examples from the labeled test split.

\item \textbf{CMExam} consists of Chinese medical licensing and professional examination questions \citep{jev-cmexam}. We retain questions with one unambiguous answer among the supplied options and convert them to \texttt{choice} decisions. We sample 200 test examples.

\item \textbf{MedQA} draws medical board-style questions from Chinese and English examination sources \citep{jev-medqa}. We preserve the original question and four candidates as a \texttt{choice} task, constructing separate Chinese and English subtasks. Each contributes 200 test examples.
\end{itemize}

\paragraph{Law.}
\begin{itemize}
\item \textbf{CUAD} annotates commercial contracts with clause types that legal reviewers may need to identify \citep{jev-cuad}. We ask whether a contract contains a specified review-relevant clause, using the contract as state and clause presence as the label of a \texttt{noul} decision. We sample 200 test examples.

\item \textbf{DISC-Law} provides Chinese legal examination questions, including single-answer and multiple-answer items \citep{jev-disc-law}. We keep the stem and options and represent a multiple-answer selection as one \texttt{choice} decision over the valid nonempty option combinations. We sample 200 test examples for each of the single-answer and multiple-answer subtasks.

\item \textbf{LexGLUE} brings together legal understanding tasks from case law, human-rights decisions, contracts, and terms of service \citep{jev-lexglue}. We convert CaseHOLD, ECtHR-A, ECtHR-B, LEDGAR, SCOTUS, and Unfair-ToS to \texttt{choice} decisions, retaining only items with a single valid class where required. Each of the six test subtasks contributes 200 examples.
\end{itemize}

\paragraph{Finance.}
\begin{itemize}
\item \textbf{FinanceBench} asks questions grounded in company filings and supplies document evidence and reference answers \citep{jev-financebench}. We retain questions whose reference answer supports an unambiguous \texttt{noul} judgment, placing the evidence in state and the question in instruction. The test subset contains 37 such examples.

\item \textbf{FinQA} pairs financial-report passages and tables with questions and annotated reasoning programs \citep{jev-finqa}. We retain questions with a verifiable yes/no answer and convert the report context and question to a \texttt{noul} decision. The test subset contains 22 examples.
\end{itemize}

\paragraph{Commonsense.}
\begin{itemize}
\item \textbf{CommonsenseQA} contains multiple-choice questions whose distractors reflect related commonsense concepts \citep{jev-commonsenseqa}. We use the question as state and its five answers as the criteria of a \texttt{choice} task, sampling 200 examples from the labeled dev split.

\item \textbf{HellaSwag} asks a model to select the plausible continuation of an everyday or instructional scenario \citep{jev-hellaswag}. The context becomes state and four endings define the criteria of a \texttt{choice} task; we sample 200 dev examples.

\item \textbf{WinoGrande} tests commonsense coreference through paired sentence-completion alternatives \citep{jev-winogrande}. We preserve the sentence and its two fillers as a \texttt{choice} decision and sample 200 labeled dev examples.
\end{itemize}

\paragraph{Agents and tools.}
\begin{itemize}
\item \textbf{AgentHarm} evaluates agent behavior on harmful and benign requests in tool-use settings \citep{jev-agentharm}. We turn its behavior annotations into \texttt{noul} harmfulness judgments and \texttt{score} severity ratings over the same request context. We sample 200 test records across these decision formats.

\item \textbf{Decider}\footnote{\url{https://github.com/Mapika/decider}} contains decision prompts for structured judgments. We retain their scenario text as state and encode candidate outcomes as criteria for \texttt{choice} tasks and binary outcomes for \texttt{noul} tasks. We sample 200 records from the labeled train split.

\item \textbf{JevBench}\footnote{\url{https://github.com/fstandhartinger/jevbench}} collects requests and responses with judgments about correctness, constraint satisfaction, and related decision properties. We preserve each request--response context and formulate its judgment as a \texttt{noul}, \texttt{choice}, or \texttt{score} decision according to the supplied label. We sample 200 test records.

\item \textbf{JevForge}\footnote{\url{https://github.com/zwliJay/jev-forge}} collects structured interaction states and next-step decisions for agent tasks. We use the task and available interface elements as state and criteria for \texttt{choice} action selection, and express binary checks as \texttt{noul} decisions. We sample 200 records from the labeled train split.
\end{itemize}

\begin{figure*}[t]
\centering
\begin{subfigure}[t]{0.48\textwidth}
\centering
\includegraphics[width=\linewidth]{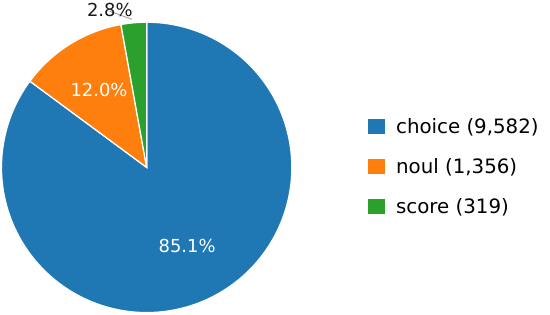}
\caption{Decision formats across all 11{,}257 instances. Slice labels show percentages; the legend gives counts.}
\label{fig:jeval-decision-formats}
\end{subfigure}\hfill
\begin{subfigure}[t]{0.48\textwidth}
\centering
\includegraphics[width=\linewidth]{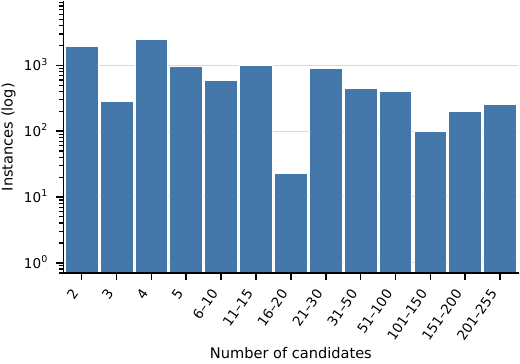}
\caption{Candidate counts among 9{,}582 \texttt{choice} instances. The vertical axis is logarithmic.}
\label{fig:jeval-candidate-distribution}
\end{subfigure}
\caption{Decision-format and candidate-space composition of the final JEVal evaluation set.}
\end{figure*}

\begin{table*}[t]
\centering
\caption{Domain-level composition of the final evaluation set. Language
follows each source subtask's designation, rather than a separate label for
every input field. Empty planned domains are omitted.}
\label{tab:jeval-domain-composition}
\small
\setlength{\tabcolsep}{3pt}
\renewcommand{\arraystretch}{1.1}
\begin{tabular}{@{}lrrrrrr@{}}
\toprule
\textbf{Domain} & \textbf{Datasets} & \textbf{Subtasks} & \textbf{Instances} & \textbf{Share (\%)} & \textbf{English} & \textbf{Chinese} \\
\midrule
\textbf{General knowledge} & 12 & 13 & 2{,}598 & 23.1 & 2{,}198 & 400 \\
\textbf{Medicine} & 3 & 4 & 800 & 7.1 & 200 & 600 \\
\textbf{Law} & 3 & 9 & 1{,}800 & 16.0 & 1{,}400 & 400 \\
\textbf{Finance} & 2 & 2 & 59 & 0.5 & 59 & 0 \\
\textbf{Commonsense} & 3 & 3 & 600 & 5.3 & 600 & 0 \\
\textbf{Agents and tools} & 4 & 4 & 800 & 7.1 & 800 & 0 \\
\textbf{Truthfulness} & 2 & 5 & 1{,}000 & 8.9 & 1{,}000 & 0 \\
\textbf{Long context} & 2 & 7 & 1{,}400 & 12.4 & 1{,}000 & 400 \\
\textbf{Social science} & 2 & 2 & 400 & 3.6 & 400 & 0 \\
\textbf{Personalization} & 3 & 9 & 1{,}800 & 16.0 & 1{,}800 & 0 \\
\midrule
\textbf{Total} & 36 & 58 & 11{,}257 & 100.0 & 9{,}457 & 1{,}800 \\
\bottomrule
\end{tabular}
\end{table*}

\paragraph{Truthfulness and hallucination.}
\begin{itemize}
\item \textbf{HaluEval} pairs grounded and hallucinated responses across dialogue, question answering, and summarization, and includes direct hallucination judgments \citep{jev-halueval}. We turn the paired responses into two-option \texttt{choice} tasks and the general judgments into \texttt{noul} decisions. Dialogue, general, QA, and summarization each contribute 200 examples from the labeled train split.

\item \textbf{TruthfulQA} asks questions chosen to elicit common falsehoods and supplies answers with truthfulness annotations \citep{jev-truthfulqa}. We use its MC1 candidates as the decision space of a \texttt{choice} task and their annotated correct answer as label, sampling 200 examples from the labeled dev split.
\end{itemize}

\paragraph{Long context and document understanding.}
\begin{itemize}
\item \textbf{HELMET} evaluates long-context models with controlled tasks at different context lengths \citep{jev-helmet}. We use its key--value retrieval instances: the long dictionary is state, the query identifies a key, and the listed values form the \texttt{choice} decision space. We sample 200 test examples.

\item \textbf{LongBench} aggregates bilingual long-context tasks, including classification and passage retrieval \citep{bai-etal-2024-longbench}. We retain LSHT, Retrieval-en, Retrieval-en-e, Retrieval-zh, TREC, and TREC-e, using the original class inventory or paragraph identifiers as criteria of \texttt{choice} tasks. Each of the six test subtasks contributes 200 examples.
\end{itemize}

\paragraph{Social science.}
\begin{itemize}
\item \textbf{BBQ} probes social bias with questions whose answers depend on whether the provided context resolves an ambiguity \citep{jev-bbq}. We preserve the context, question, three candidate answers, and annotated answer in a \texttt{choice} task, sampling 200 test examples.

\item \textbf{GlobalOpinionQA} organizes survey questions and response distributions across countries \citep{jev-globalopinions}. We combine a question with its country context and ask for a \texttt{choice} decision selecting the most common response among its original answer options, excluding tied modes. We sample 200 test examples.
\end{itemize}

\paragraph{Personalization and preference.}
\begin{itemize}
\item \textbf{CommunityBench} evaluates whether a model can infer and follow the preferences of an online community \citep{jev-communitybench}. We convert community prediction and response preference identification into \texttt{choice} tasks conditioned on community descriptions or examples. Each test subtask contributes 200 examples.

\item \textbf{CompRed}\footnote{\url{https://huggingface.co/datasets/allenai/compred}} contains community-grounded response preferences from discussions in several topical domains. We present the discussion history as state and ask which candidate response the community prefers in a \texttt{choice} task. Finance, gender and sexuality, history, politics, and science each contribute 200 test examples.

\item \textbf{PRISM-Alignment} records participant preferences and ratings of model responses in conversational settings \citep{jev-prism}. We use the participant context to predict the preferred response as a \texttt{choice} task or an ordered rating band as a \texttt{score} task. Both the alignment and score subtasks contribute 200 examples from the labeled train split.
\end{itemize}

\subsection{Benchmark Composition}
\label{app:dataset-statistics}

\begin{table}[t]
\centering
\caption{Input lengths by domain in \texttt{cl100k\_base} tokens. Each
instance is capped at 20,480 tokens before calculating the mean and maximum
for this descriptive table. Stored inputs above the cap are not claimed to
have been truncated during evaluation.}
\label{tab:jeval-domain-lengths}
\small
\setlength{\tabcolsep}{3pt}
\renewcommand{\arraystretch}{1.1}
\begin{tabular}{@{}lrr@{}}
\toprule
\textbf{Domain} & \textbf{Mean} & \textbf{Maximum} \\
\midrule
\textbf{General knowledge} & 348 & 3{,}447 \\
\textbf{Medicine} & 136 & 653 \\
\textbf{Law} & 2{,}831 & 20{,}480 \\
\textbf{Finance} & 565 & 1{,}591 \\
\textbf{Commonsense} & 85 & 276 \\
\textbf{Agents and tools} & 267 & 3{,}523 \\
\textbf{Truthfulness} & 327 & 2{,}312 \\
\textbf{Long context} & 10{,}433 & 20{,}480 \\
\textbf{Social science} & 71 & 221 \\
\textbf{Personalization} & 474 & 6{,}769 \\
\midrule
\textbf{All instances} & 1{,}974 & 20{,}480 \\
\bottomrule
\end{tabular}
\end{table}

\paragraph{Coverage.}
The final JEVal evaluation set comprises 36 datasets, 58 subtasks, and 11,257 instances, after sampling from the larger pool of normalized candidate examples. We treat each upstream source as a single dataset, each distinct evaluation configuration as a subtask, and each retained decision record as an instance. Table~\ref{tab:jeval-domain-composition} summarizes the number of datasets, subtasks, and instances by domain, together with the source-designated language distribution.

\onecolumn

\begingroup
\small
\setlength{\tabcolsep}{3pt}
\renewcommand{\arraystretch}{1.1}
\begin{longtable}{@{}p{0.18\textwidth}p{0.77\textwidth}@{}}
\caption{Representative JEVal test cases. Each row shows one evaluation instance in the shared decision format.}
\label{tab:benchmark-test-cases}\\
\toprule
\textbf{Dataset and format} & \textbf{Evaluation instance} \\
\midrule
\endfirsthead
\toprule
\textbf{Dataset and format} & \textbf{Evaluation instance} \\
\midrule
\endhead

\textbf{AI2 ARC}\par \textbf{Science question}\par \textbf{\texttt{choice}} &
\textbf{State:} It was once thought that living organisms could come from non-living matter. For example, people believed that flies would develop from rotting meat.\par
\textbf{Instruction:} This idea was later disproved primarily because of\par
\textbf{Criteria:} \texttt{opt\_1}: the discovery of the atom; \texttt{opt\_2}: better surgical techniques; \texttt{opt\_3}: continued experimentation; \texttt{opt\_4}: the invention of the microscope.\par
\textbf{Gold label:} \texttt{opt\_3} (continued experimentation). \\
\midrule

\textbf{MedQA}\par \textbf{Medical examination}\par \textbf{\texttt{choice}} &
\textbf{State:} A 72-year-old anthropologist with long-standing hypertension visits your office for a routine exam. You notice an abnormality on his laboratory results caused by his regimen of captopril and triamterene. What abnormality did you most likely find?\par
\textbf{Instruction:} Based on the question, choose the correct option.\par
\textbf{Criteria:} A: Hyperkalemia; B: Hypernatremia; C: Thrombocytopenia; D: Anemia.\par
\textbf{Gold label:} A (Hyperkalemia). \\
\midrule

\textbf{LexGLUE}\par \textbf{Unfair contract clause}\par \textbf{\texttt{choice}} &
\textbf{State:} We may modify this contract, our privacy policy and our cookies policies from time to time.\par
\textbf{Instruction:} Which type of unfair clause does this terms-of-service excerpt contain? Choose None of the 8 if it contains none.\par
\textbf{Criteria:} Limitation of liability; Unilateral termination; Unilateral change; Content removal; Contract by using; Choice of law; Jurisdiction; Arbitration; None of the 8.\par
\textbf{Gold label:} Unilateral change. \\
\midrule

\textbf{HaluEval}\par \textbf{Response grounding}\par \textbf{\texttt{noul}} &
\textbf{State:} User query: ``Sketch a logo for the company International Business Company.'' Response: ``Unfortunately, as an AI language model, I do not have the capability to generate visual content. Can I assist you with anything else?''\par
\textbf{Instruction:} Does this response contain hallucinated content not supported by the query?\par
\textbf{Criteria:} \texttt{false}: The response is faithful to the source and contains no hallucination; \texttt{true}: The response contains content not supported by the source (hallucinated).\par
\textbf{Gold label:} \texttt{false}. \\
\midrule

\textbf{LongBench}\par \textbf{Paragraph retrieval}\par \textbf{\texttt{choice}} &
\textbf{State (excerpt):} Paragraph 16: ``This species of lizard has a large head that is elongated and depressed, with the cheeks swollen in adult males.'' Paragraph 17: ``Biometrika was established in 1901 by Francis Galton, Karl Pearson, and Raphael Weldon to promote the study of biometrics.'' The other 28 paragraphs are omitted here.\par
\textbf{Instruction (excerpt):} Which paragraph in the passage below matches this description? The text describes the physical characteristics of a species of lizard. It focuses on details such as the large head with swollen cheeks in adult males, the length of the snout, \ldots\par
\textbf{Criteria:} Paragraph 1 through Paragraph 30 (30 candidate identifiers).\par
\textbf{Gold label:} Paragraph 16. \\
\midrule

\textbf{JevBench}\par \textbf{Incident impact}\par \textbf{\texttt{score}} &
\textbf{State:} The icon is misaligned. Every function works.\par
\textbf{Instruction:} Rate incident impact using only reported facts. Use the highest fully supported level.\par
\textbf{Criteria:} 0: No function impaired; cosmetic only. 1: One user or a nonessential function impaired, with a workaround. 2: Many users blocked from a core function, no data loss. 3: Confirmed irreversible data loss or physical harm.\par
\textbf{Gold label:} 0 (cosmetic only). \\
\bottomrule
\end{longtable}
\endgroup

\paragraph{Decision formats and candidate spaces.}
The relative frequencies of the three decision formats are shown in Figure~\ref{fig:jeval-decision-formats}. Choice decisions dominate the final set, while binary and ordered-score decisions provide distinct output structures. For the 9{,}582 choice instances, the number of candidates ranges from two to 255. Figure~\ref{fig:jeval-candidate-distribution} shows the candidate-count distribution using the same bar-chart style as the input-length distribution in the main text. The bins cover the entire observed range.

\paragraph{Input length by domain.}
Table~\ref{tab:jeval-domain-lengths} reports the mean and maximum input length for each domain, measured over state, instruction, and criteria with \texttt{cl100k\_base}. For this table, we cap each measured length at the Jev reporting limit of 20{,}480 tokens before computing either statistic. This gives the length scale within the Jev budget while retaining all 11{,}257 records in the denominator. In the stored evaluation data, 202 raw inputs exceed 20{,}480 tokens; the cap in this table is a reporting convention rather than evidence that those stored records were truncated before evaluation.

\subsection{Representative Test Cases}
\label{app:benchmark-test-cases}

Table~\ref{tab:benchmark-test-cases} presents six instances drawn from the final JEVal evaluation set: AI2 ARC \citep{jev-arc}, MedQA \citep{jev-medqa}, LexGLUE \citep{jev-lexglue}, HaluEval \citep{jev-halueval}, LongBench \citep{bai-etal-2024-longbench}, and JevBench \citep{TypeSafe2026Jev}. They illustrate how different source tasks appear under the common decision interface. We show the state, instruction, criteria, and gold label for each instance.

\FloatBarrier
\begingroup
\def\JEValNA{\textemdash}
\def\JEValBrierRaw{\mathrm{Brier}_{\mathrm{raw}}}
\long\def\JEValCode#1{{\ttfamily\mdseries\upshape #1}}
\long\def\JEValChange#1{\ifdefined\JEValShowChanges{\bfseries\boldmath #1}\else#1\fi}
\setlength{\LTpre}{4pt}\setlength{\LTpost}{5pt}\setlength{\LTcapwidth}{\textwidth}
\newlength{\JEValCompactCol}
\def\JEValModeKey{N: non-thinking; T: thinking. Parentheses specify reasoning effort.}
\def\JEValSetTableWidths#1{\setlength{\tabcolsep}{1.8pt}\setlength{\JEValCompactCol}{\dimexpr(\linewidth-2\tabcolsep*(#1-1))/#1\relax}}
\long\def\JEValHeader#1#2{\parbox[c][27pt][c]{\linewidth}{\centering\normalfont\fontsize{7.3}{8.1}\selectfont #1\par#2\par}}

\section{More JEVal experimental settings}

\subsection{Input and Output Protocol}
All configurations answer the same 11,257 instances from 58 subtasks, and all but 319 \texttt{score} questions are \texttt{choice} or \texttt{noul} classification questions. Every model receives the same task content in the state--instruction--criteria form of Section~\ref{sec:task_formulation}, but the two kinds of model produce probabilities differently. Decision models read them from a prediction head over the candidates: Jev through its native interface, the other open decision models through their own encoders and heads with the candidate order unchanged, and \oursmall{} and \ourlarge{} through a softmax over the candidate labels at the first answer position. Generative LLMs instead write their answer and a probability for every candidate as JSON that follows a question-specific schema, whether or not they think first. Appendix~\ref{app:prompts} gives the prompt templates, schemas, and validation rules.

\subsection{Prompt Templates and Response Protocol}
\label{app:prompts}

\paragraph{Input representation and prompt construction.}
Jev receives the benchmark's standardized \JEValCode{state} and \JEValCode{questions} fields through its native interface. The seven generative LLM configurations receive the same task content through a shared system message and task-specific user templates, shown in Figure~\ref{fig:prompt-protocol}. Each user message contains the task state, instructions, and the available options or score levels. Candidate identifiers, descriptions, order, and language are preserved. Strings remain unchanged, while objects and arrays are serialized as compact JSON with \JEValCode{ensure\_ascii=False}. Missing or null descriptions are omitted together with the preceding colon. Score levels follow their original order and are indexed from $0$ to $K-1$, where $K$ is the number of levels.

\paragraph{Response specification.}
Each LLM request includes a question-specific JSON Schema through \JEValCode{response\_format}, with \JEValCode{type=json\_schema} and \JEValCode{strict=true}. Responses are organized under \JEValCode{answers}, keyed by the original question identifier. Each answer specifies its \JEValCode{type}: \JEValCode{choice} responses return an original option identifier, \JEValCode{noul} responses return the probability that the answer is true, and \JEValCode{score} responses return a numerical value in $[0,K-1]$. Choice and score responses additionally include \JEValCode{confidence} and a \JEValCode{probabilities} mapping containing every option or level exactly once, including zero-valued entries. Score responses also include a \JEValCode{legend} mapping level identifiers to their original descriptions. Confidence and probability values must lie in $[0,1]$. All specified fields are required, and every schema object sets \JEValCode{additionalProperties=false}.

\paragraph{Output validation.}
Only the final \JEValCode{message.content} is parsed for LLM answers; reasoning text is excluded. Incomplete or malformed JSON, duplicate keys, non-finite numbers, missing or additional fields, invalid candidate identifiers, and out-of-range values are rejected without repair. The returned choice is not replaced by the probability argmax, and the returned score is not recomputed as a probability-weighted expectation. The independent \JEValCode{confidence} field is not required to equal the selected-option probability. Probability vectors are neither normalized nor rejected because their entries do not sum to one.

\begin{figure}[!htb]
\centering
\includegraphics[width=.98\textwidth,height=.6748823529\textheight,keepaspectratio]{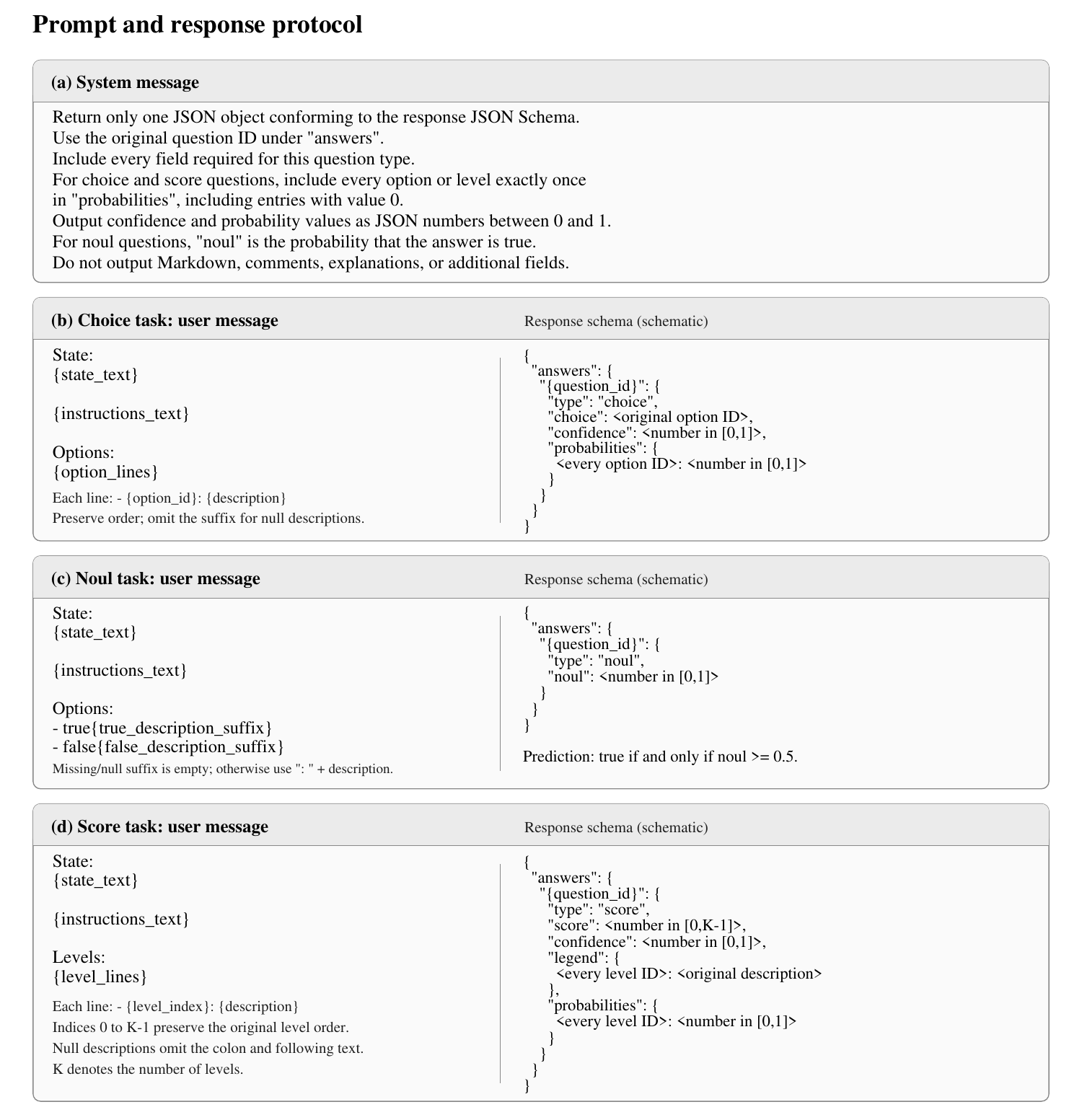}
\caption{Prompt templates and response structures for the \JEValChange{nine} LLM configurations. Panel (a) presents the shared system message; panels (b)--(d) present the user templates and schematic responses for \JEValCode{choice}, \JEValCode{noul}, and \JEValCode{score}, respectively. Placeholders are instantiated per question. Formatting notes, value constraints, and the binary prediction rule are explanatory annotations, not additional prompt text.}
\label{fig:prompt-protocol}
\end{figure}

\subsection{Model Configurations}
\label{sec:Model_Configurations}

Table~\ref{tab:inference-configurations} summarizes the inference configurations used in our main experiments, covering decision models and both non-thinking and thinking configurations of generative LLMs. For each configuration, we report the applicable reasoning controls, token budgets, and temperature. The reported context windows and token limits represent experimental budgets rather than the models' native context capacities. Temperature is used for candidate-probability scaling in decision models and token sampling in generative LLMs.

Local benchmark runs use eight NVIDIA H100 80\,GB GPUs with a development build of vLLM~\JEValCode{0.29.1rc1}. The generative Qwen models use BF16 weights, and DeepSeek-V4-Flash and GLM-5.3-Flash use FP8-quantized weights. LLM requests are non-streaming. Non-thinking requests use output budgets of 256 to 8,192 tokens depending on the response schema, and thinking requests use an 8,192-token output budget that includes a 4,096-token thinking budget. SII-Decision runs with BF16 base weights and unmerged BF16 LoRA adapters, with tensor parallelism 1 and data parallelism 8. The other local decision models keep their checkpoint-specific precision and calibration parameters. Only unsuccessful requests are retried, and the first valid response is kept.

\begin{table}[!htbp]
\centering
\caption{Inference configurations for JEVal. \textbf{Context.} Token columns report maximum experimental budgets; input limits apply to complete encoded paths. Relative to released encoders, we raise Laya's limit from 512 to 8,192 tokens and the limits of LightJev (256), JevForge (768), OpenJev (4,096), and NanoJev, Nimble, and Intern-Decision (8,192) to 24,576 tokens; Kev accepts an explicit encoding limit. Backbone positional configurations remain unchanged: 8,192 for Laya, 40,960 for NanoJev and LightJev, and 262,144 for the Qwen3.5/3.8-based decision models, including SII-Decision. Some NanoJev and LightJev runs use shorter context/input limits of 8,192 and 256 tokens, respectively. GPT entries are client-side budgets. Laya retains its 48-token option limit; OpenJev processes all 24,000-character windows with 2,000-character overlap. \textbf{Output.} Thinking and final-answer tokens share the output budget; SII-Decision's one-token allowance supports candidate-logit readout. \textbf{Temperature.} Values specify candidate-probability scaling for decision models and token sampling for generative LLMs. Laya uses task-specific temperatures of 1.6369 (choice), 1.2514 (score), and 1.9834 (noul), with choice overrides of 1.9064, 1.7602, 1.0000, and 0.1006 for 2, 3--5, 6--10, and $\geq$11 options, respectively. \JEValNA{} denotes an inapplicable or unreported scalar.}
\label{tab:inference-configurations}
\begingroup\fontsize{8}{10.2}\selectfont
\setlength{\tabcolsep}{3pt}\renewcommand{\arraystretch}{1.07}
\begin{adjustbox}{max width=0.97\linewidth}
\begin{tabular}{@{}lccccccc@{}}

\toprule
\textbf{Model} & \textbf{Mode} & \shortstack{\textbf{Reasoning}\\\textbf{effort}} & \shortstack{\textbf{Context}\\\textbf{window}} & \shortstack{\textbf{Input}\\\textbf{budget}} & \shortstack{\textbf{Output}\\\textbf{budget}} & \shortstack{\textbf{Thinking}\\\textbf{budget}} & \textbf{Temperature}\\
\midrule
\multicolumn{8}{@{}l}{\textit{Decision models}}\\*
Jev & \JEValNA & \JEValNA & \JEValNA & \JEValNA & \JEValNA & \JEValNA & \JEValNA\\
Kev-0.8b & \JEValNA & \JEValNA & 32,768 & 24,576 & \JEValNA & \JEValNA & 2.3511\\
Kev-4b & \JEValNA & \JEValNA & 32,768 & 24,576 & \JEValNA & \JEValNA & 2.4061\\
Kev-9b & \JEValNA & \JEValNA & 32,768 & 24,576 & \JEValNA & \JEValNA & 2.2974\\
Kev-27b & \JEValNA & \JEValNA & 32,768 & 24,576 & \JEValNA & \JEValNA & 1.3819\\
Intern-Decision-0.8B & \JEValNA & \JEValNA & 32,768 & 24,576 & \JEValNA & \JEValNA & 2.7478\\
Intern-Decision-2B & \JEValNA & \JEValNA & 32,768 & 24,576 & \JEValNA & \JEValNA & 2.1005\\
Intern-Decision-4B & \JEValNA & \JEValNA & 32,768 & 24,576 & \JEValNA & \JEValNA & 1.9924\\
Laya & \JEValNA & \JEValNA & 8,192 & 8,192 & \JEValNA & \JEValNA & \JEValNA\\
Openjev-4b-v5 & \JEValNA & \JEValNA & 32,768 & 24,576 & \JEValNA & \JEValNA & \JEValNA\\
Nimble-9b & \JEValNA & \JEValNA & 32,768 & 24,576 & \JEValNA & \JEValNA & 1.0\\
Nanojev & \JEValNA & \JEValNA & 32,768 & 24,576 & \JEValNA & \JEValNA & 1.0\\
Lightjev & \JEValNA & \JEValNA & 32,768 & 24,576 & \JEValNA & \JEValNA & 1.0\\
Jevforge-0.8b & \JEValNA & \JEValNA & 32,768 & 24,576 & \JEValNA & \JEValNA & 0.9717\\
\textbf{SII-Decision-4B (ours)} & \JEValNA & \JEValNA & 32,768 & 24,576 & 1 & \JEValNA & 1.0\\
\textbf{SII-Decision-27B (ours)} & \JEValNA & \JEValNA & 32,768 & 24,576 & 1 & \JEValNA & 1.0\\
\addlinespace[2pt]
\multicolumn{8}{@{}l}{\textit{Non-thinking generative LLMs}}\\*
Qwen3.5-4B & Non-thinking & \JEValNA & 32,768 & 24,576 & 8,192 & \JEValNA & 1.0\\
Qwen3.8-27B & Non-thinking & \JEValNA & 32,768 & 24,576 & 8,192 & \JEValNA & 1.0\\
DeepSeek-V4-Flash & Non-thinking & \JEValNA & 32,768 & 24,576 & 8,192 & \JEValNA & 1.0\\
GPT-6-Luna & Non-thinking & \JEValNA & 32,768 & 24,576 & 8,192 & \JEValNA & 1.0\\
GPT-6-Sol & Non-thinking & \JEValNA & 32,768 & 24,576 & 8,192 & \JEValNA & 1.0\\
\addlinespace[2pt]
\multicolumn{8}{@{}l}{\textit{Thinking generative LLMs}}\\*
Qwen3.5-4B & Thinking & \JEValCode{default} & 32,768 & 24,576 & 8,192 & 4,096 & 1.0\\
Qwen3.8-27B & Thinking & \JEValCode{xhigh} & 32,768 & 24,576 & 8,192 & 4,096 & 1.0\\
DeepSeek-V4-Flash & Thinking & \JEValCode{high} & 32,768 & 24,576 & 8,192 & 4,096 & 1.0\\
GLM-5.3-Flash & Thinking & \JEValCode{high} & 32,768 & 24,576 & 8,192 & 4,096 & 1.0\\
\addlinespace[2pt]
\bottomrule
\end{tabular}
\end{adjustbox}
\par\endgroup
\end{table}

\section{Benchmark-Level Results}
\label{app:benchmark-results}

This section presents benchmark-level results for all 25 model configurations across the 58 JEVal benchmarks in ten domains, together with pooled domain summaries. We report accuracy, ECE, and $\mathrm{Brier}_{\mathrm{raw}}$ for classification questions, MAE separately for score questions, and mean latency over valid responses across all question types. Accuracy counts skips and final failures as incorrect, whereas ECE and $\mathrm{Brier}_{\mathrm{raw}}$ are computed only from valid classification responses. Bold and underline indicate the best and second-best distinct displayed values, respectively, across all 25 configurations.

\textbf{Mean absolute error.}
For the valid \texttt{score} responses $\mathcal S_m$, we compute
\begin{equation}
\mathrm{MAE}_m
=
\frac{1}{|\mathcal S_m|}
\sum_{i\in\mathcal S_m}
|\hat s_{mi}-s_i|,
\label{eq:jeval-mae}
\end{equation}
where $\hat s_{mi}$ is the returned score and $s_i$ is its reference value. Each question retains its defined rating scale, and invalid outputs are excluded rather than imputed.

\Needspace{100pt}
\subsection{General knowledge and reasoning}
\label{app:domain-general-knowledge-and-reasoning}
\begingroup\def\rmdefault{lmr}\normalfont\fontsize{8}{9.3}\selectfont\JEValSetTableWidths{9}\renewcommand{\arraystretch}{1.0}
\begin{longtable}{@{}>{\raggedright\arraybackslash}p{\JEValCompactCol}>{\raggedleft\arraybackslash}p{\JEValCompactCol}>{\raggedleft\arraybackslash}p{\JEValCompactCol}>{\raggedleft\arraybackslash}p{\JEValCompactCol}>{\raggedleft\arraybackslash}p{\JEValCompactCol}>{\raggedleft\arraybackslash}p{\JEValCompactCol}>{\raggedleft\arraybackslash}p{\JEValCompactCol}>{\raggedleft\arraybackslash}p{\JEValCompactCol}>{\raggedleft\arraybackslash}p{\JEValCompactCol}@{}}
\caption{General knowledge and reasoning: model band 1/3. ACC (\%) $\uparrow$; benchmark-run latency (s), ECE, and $\JEValBrierRaw$ $\downarrow$. Rankings span all bands.}\label{tab:domain-general-knowledge-and-reasoning}\\
\toprule
\multicolumn{9}{@{}l@{}}{\fontsize{7.3}{8.4}\selectfont\JEValModeKey}\\[1.5pt]
\JEValHeader{\textbf{Metric}}{} & \JEValHeader{Jev}{} & \JEValHeader{SII-Decision-4B}{(ours)} & \JEValHeader{SII-Decision-27B}{(ours)} & \JEValHeader{kev-0.8b}{} & \JEValHeader{kev-4b}{} & \JEValHeader{kev-9b}{} & \JEValHeader{kev-27b}{} & \JEValHeader{Intern-Decision\\0.8B}{}\\\midrule
\endfirsthead
\multicolumn{9}{@{}l}{\textit{Table \thetable{} continued; model band 1/3}}\\\toprule
\multicolumn{9}{@{}l@{}}{\fontsize{7.3}{8.4}\selectfont\JEValModeKey}\\[1.5pt]
\JEValHeader{\textbf{Metric}}{} & \JEValHeader{Jev}{} & \JEValHeader{SII-Decision-4B}{(ours)} & \JEValHeader{SII-Decision-27B}{(ours)} & \JEValHeader{kev-0.8b}{} & \JEValHeader{kev-4b}{} & \JEValHeader{kev-9b}{} & \JEValHeader{kev-27b}{} & \JEValHeader{Intern-Decision\\0.8B}{}\\\midrule
\endhead
\midrule\multicolumn{9}{r@{}}{\fontsize{7.3}{8.4}\selectfont Continued on next page}\\\endfoot
\bottomrule\endlastfoot
\multicolumn{9}{@{}l@{}}{\textbf{aida-testc}\hfill $N=200$}\\*
ACC $\uparrow$ & 95.00 & 86.00 & 97.50 & 38.00 & 80.00 & 96.00 & 98.50 & 6.50\\*
Latency $\downarrow$ & \textbf{2.11} & \underline{5.80} & 14.02 & 18.15 & 57.79 & 94.05 & 55.95 & 6.38\\*
ECE $\downarrow$ & 0.009 & 0.043 & 0.012 & 0.169 & 0.532 & 0.302 & 0.077 & 0.051\\*
$\JEValBrierRaw\,\downarrow$ & 0.032 & 0.193 & 0.040 & 0.807 & 0.600 & 0.191 & 0.037 & 0.980\\[1.5pt]
\multicolumn{9}{@{}l@{}}{\textbf{ai2-arc-challenge}\hfill $N=200$}\\*
ACC $\uparrow$ & 98.00 & 90.50 & 96.00 & 57.00 & 92.50 & 94.00 & 96.00 & 60.00\\*
Latency $\downarrow$ & \textbf{0.87} & 5.59 & 12.41 & 18.10 & 56.66 & 94.98 & 56.51 & 6.38\\
\end{longtable}
\endgroup

\begingroup\def\rmdefault{lmr}\normalfont\fontsize{8}{9.3}\selectfont\JEValSetTableWidths{9}\renewcommand{\arraystretch}{1.0}
\begin{longtable}{@{}>{\raggedright\arraybackslash}p{\JEValCompactCol}>{\raggedleft\arraybackslash}p{\JEValCompactCol}>{\raggedleft\arraybackslash}p{\JEValCompactCol}>{\raggedleft\arraybackslash}p{\JEValCompactCol}>{\raggedleft\arraybackslash}p{\JEValCompactCol}>{\raggedleft\arraybackslash}p{\JEValCompactCol}>{\raggedleft\arraybackslash}p{\JEValCompactCol}>{\raggedleft\arraybackslash}p{\JEValCompactCol}>{\raggedleft\arraybackslash}p{\JEValCompactCol}@{}}
\caption{General knowledge and reasoning: model band 2/3. ACC (\%) $\uparrow$; benchmark-run latency (s), ECE, and $\JEValBrierRaw$ $\downarrow$. Rankings span all bands.}\label{tab:domain-general-knowledge-and-reasoning-band-2}\\
\toprule
\multicolumn{9}{@{}l@{}}{\fontsize{7.3}{8.4}\selectfont\JEValModeKey}\\[1.5pt]
\JEValHeader{\textbf{Metric}}{} & \JEValHeader{Intern-Decision\\2B}{} & \JEValHeader{Intern-Decision\\4B}{} & \JEValHeader{laya}{} & \JEValHeader{openjev-4b-v5}{} & \JEValHeader{nimble-9b}{} & \JEValHeader{nanojev}{} & \JEValHeader{lightjev}{} & \JEValHeader{jevforge-0.8b}{}\\\midrule
\endfirsthead
\multicolumn{9}{@{}l}{\textit{Table \thetable{} continued; model band 2/3}}\\\toprule
\multicolumn{9}{@{}l@{}}{\fontsize{7.3}{8.4}\selectfont\JEValModeKey}\\[1.5pt]
\JEValHeader{\textbf{Metric}}{} & \JEValHeader{Intern-Decision\\2B}{} & \JEValHeader{Intern-Decision\\4B}{} & \JEValHeader{laya}{} & \JEValHeader{openjev-4b-v5}{} & \JEValHeader{nimble-9b}{} & \JEValHeader{nanojev}{} & \JEValHeader{lightjev}{} & \JEValHeader{jevforge-0.8b}{}\\\midrule
\endhead
\midrule\multicolumn{9}{r@{}}{\fontsize{7.3}{8.4}\selectfont Continued on next page}\\\endfoot
\bottomrule\endlastfoot
\multicolumn{9}{@{}l@{}}{\textbf{aida-testc}\hfill $N=200$}\\*
ACC $\uparrow$ & 13.00 & 83.50 & 19.00 & 87.50 & 94.00 & 1.50 & 42.00 & 16.00\\*
Latency $\downarrow$ & 6.32 & 10.07 & 12.40 & 229.93 & 23.87 & 40.14 & 266.04 & 216.20\\*
ECE $\downarrow$ & 0.203 & 0.314 & 0.625 & 0.542 & 0.046 & \textbf{0.001} & 0.406 & 0.125\\*
$\JEValBrierRaw\,\downarrow$ & 1.004 & 0.376 & 1.387 & 0.521 & 0.094 & 1.000 & 0.980 & 0.977\\[1.5pt]
\multicolumn{9}{@{}l@{}}{\textbf{ai2-arc-challenge}\hfill $N=200$}\\*
ACC $\uparrow$ & 75.50 & 93.00 & 33.50 & 93.50 & 96.50 & 27.00 & 29.00 & 30.50\\*
Latency $\downarrow$ & 6.43 & 9.82 & 12.88 & 233.40 & 23.72 & 37.46 & 16.97 & 221.52\\
\end{longtable}
\endgroup

\begingroup\def\rmdefault{lmr}\normalfont\fontsize{8}{9.3}\selectfont\JEValSetTableWidths{10}\renewcommand{\arraystretch}{1.0}
\begin{longtable}{@{}>{\raggedright\arraybackslash}p{\JEValCompactCol}>{\raggedleft\arraybackslash}p{\JEValCompactCol}>{\raggedleft\arraybackslash}p{\JEValCompactCol}>{\raggedleft\arraybackslash}p{\JEValCompactCol}>{\raggedleft\arraybackslash}p{\JEValCompactCol}>{\raggedleft\arraybackslash}p{\JEValCompactCol}>{\raggedleft\arraybackslash}p{\JEValCompactCol}>{\raggedleft\arraybackslash}p{\JEValCompactCol}>{\raggedleft\arraybackslash}p{\JEValCompactCol}>{\raggedleft\arraybackslash}p{\JEValCompactCol}@{}}
\caption{General knowledge and reasoning: model band 3/3. ACC (\%) $\uparrow$; benchmark-run latency (s), ECE, and $\JEValBrierRaw$ $\downarrow$. Rankings span all bands.}\label{tab:domain-general-knowledge-and-reasoning-band-3}\\
\toprule
\multicolumn{10}{@{}l@{}}{\fontsize{7.3}{8.4}\selectfont\JEValModeKey}\\[1.5pt]
\JEValHeader{\textbf{Metric}}{} & \JEValHeader{Qwen3.5-4B}{N} & \JEValHeader{Qwen3.8-27B}{N} & \JEValHeader{DeepSeek\\V4-Flash}{N} & \JEValHeader{GPT-6-Sol}{N} & \JEValHeader{GPT-6-Luna}{N} & \JEValHeader{Qwen3.5-4B}{T} & \JEValHeader{Qwen3.8-27B}{T (xhigh)} & \JEValHeader{DeepSeek\\V4-Flash}{T (high)} & \JEValHeader{GLM-5.3-Flash}{T (high)}\\\midrule
\endfirsthead
\multicolumn{10}{@{}l}{\textit{Table \thetable{} continued; model band 3/3}}\\\toprule
\multicolumn{10}{@{}l@{}}{\fontsize{7.3}{8.4}\selectfont\JEValModeKey}\\[1.5pt]
\JEValHeader{\textbf{Metric}}{} & \JEValHeader{Qwen3.5-4B}{N} & \JEValHeader{Qwen3.8-27B}{N} & \JEValHeader{DeepSeek\\V4-Flash}{N} & \JEValHeader{GPT-6-Sol}{N} & \JEValHeader{GPT-6-Luna}{N} & \JEValHeader{Qwen3.5-4B}{T} & \JEValHeader{Qwen3.8-27B}{T (xhigh)} & \JEValHeader{DeepSeek\\V4-Flash}{T (high)} & \JEValHeader{GLM-5.3-Flash}{T (high)}\\\midrule
\endhead
\midrule\multicolumn{10}{r@{}}{\fontsize{7.3}{8.4}\selectfont Continued on next page}\\\endfoot
\bottomrule\endlastfoot
\multicolumn{10}{@{}l@{}}{\textbf{aida-testc}\hfill $N=200$}\\*
ACC $\uparrow$ & 77.00 & 96.50 & 91.00 & \textbf{99.50} & 96.50 & 89.00 & 97.50 & 98.50 & \underline{99.00}\\*
Latency $\downarrow$ & 25.21 & 90.29 & 54.52 & 45.27 & 13.75 & 74.46 & 248.88 & 206.46 & 211.86\\*
ECE $\downarrow$ & 0.204 & 0.024 & 0.087 & \underline{0.005} & 0.035 & 0.035 & 0.015 & 0.020 & 0.008\\*
$\JEValBrierRaw\,\downarrow$ & 4.291 & 0.064 & 0.180 & \textbf{0.010} & 0.070 & 0.099 & 0.028 & 0.035 & \underline{0.022}\\[1.5pt]
\multicolumn{10}{@{}l@{}}{\textbf{ai2-arc-challenge}\hfill $N=200$}\\*
ACC $\uparrow$ & 83.50 & 93.00 & 94.00 & 97.50 & \textbf{99.00} & 95.00 & \underline{98.50} & 97.00 & 98.00\\*
Latency $\downarrow$ & \underline{1.66} & 3.18 & 3.30 & 10.20 & 8.85 & 43.20 & 96.70 & 61.12 & 8.76\\
\end{longtable}
\endgroup

\Needspace{100pt}
\subsection{Medicine}
\label{app:domain-medicine}
\begingroup\def\rmdefault{lmr}\normalfont\fontsize{8}{9.3}\selectfont\JEValSetTableWidths{9}\renewcommand{\arraystretch}{1.0}
\begin{longtable}{@{}>{\raggedright\arraybackslash}p{\JEValCompactCol}>{\raggedleft\arraybackslash}p{\JEValCompactCol}>{\raggedleft\arraybackslash}p{\JEValCompactCol}>{\raggedleft\arraybackslash}p{\JEValCompactCol}>{\raggedleft\arraybackslash}p{\JEValCompactCol}>{\raggedleft\arraybackslash}p{\JEValCompactCol}>{\raggedleft\arraybackslash}p{\JEValCompactCol}>{\raggedleft\arraybackslash}p{\JEValCompactCol}>{\raggedleft\arraybackslash}p{\JEValCompactCol}@{}}
\caption{Medicine: model band 1/3. ACC (\%) $\uparrow$; benchmark-run latency (s), ECE, and $\JEValBrierRaw$ $\downarrow$. Rankings span all bands.}\label{tab:domain-medicine}\\
\toprule
\multicolumn{9}{@{}l@{}}{\fontsize{7.3}{8.4}\selectfont\JEValModeKey}\\[1.5pt]
\JEValHeader{\textbf{Metric}}{} & \JEValHeader{Jev}{} & \JEValHeader{SII-Decision-4B}{(ours)} & \JEValHeader{SII-Decision-27B}{(ours)} & \JEValHeader{kev-0.8b}{} & \JEValHeader{kev-4b}{} & \JEValHeader{kev-9b}{} & \JEValHeader{kev-27b}{} & \JEValHeader{Intern-Decision\\0.8B}{}\\\midrule
\endfirsthead
\multicolumn{9}{@{}l}{\textit{Table \thetable{} continued; model band 1/3}}\\\toprule
\multicolumn{9}{@{}l@{}}{\fontsize{7.3}{8.4}\selectfont\JEValModeKey}\\[1.5pt]
\JEValHeader{\textbf{Metric}}{} & \JEValHeader{Jev}{} & \JEValHeader{SII-Decision-4B}{(ours)} & \JEValHeader{SII-Decision-27B}{(ours)} & \JEValHeader{kev-0.8b}{} & \JEValHeader{kev-4b}{} & \JEValHeader{kev-9b}{} & \JEValHeader{kev-27b}{} & \JEValHeader{Intern-Decision\\0.8B}{}\\\midrule
\endhead
\midrule\multicolumn{9}{r@{}}{\fontsize{7.3}{8.4}\selectfont Continued on next page}\\\endfoot
\bottomrule\endlastfoot
\multicolumn{9}{@{}l@{}}{\textbf{cmb}\hfill $N=200$}\\*
ACC $\uparrow$ & 84.50 & 67.50 & 80.50 & 33.50 & 63.50 & 68.00 & 76.00 & 36.50\\*
Latency $\downarrow$ & \textbf{0.82} & 5.63 & 12.45 & 17.62 & 57.19 & 94.69 & 54.87 & 6.35\\*
ECE $\downarrow$ & 0.056 & 0.095 & \underline{0.048} & 0.066 & 0.076 & 0.104 & 0.145 & 0.073\\*
$\JEValBrierRaw\,\downarrow$ & 0.198 & 0.425 & 0.269 & 0.740 & 0.484 & 0.424 & 0.366 & 0.714\\[1.5pt]
\multicolumn{9}{@{}l@{}}{\textbf{cmexam}\hfill $N=200$}\\*
ACC $\uparrow$ & 86.50 & 75.00 & 84.00 & 41.00 & 66.00 & 68.50 & 81.00 & 40.50\\*
Latency $\downarrow$ & \textbf{0.65} & 5.64 & 12.38 & 17.51 & 56.83 & 93.76 & 55.06 & 6.37\\
\end{longtable}
\endgroup

\begingroup\def\rmdefault{lmr}\normalfont\fontsize{8}{9.3}\selectfont\JEValSetTableWidths{9}\renewcommand{\arraystretch}{1.0}
\begin{longtable}{@{}>{\raggedright\arraybackslash}p{\JEValCompactCol}>{\raggedleft\arraybackslash}p{\JEValCompactCol}>{\raggedleft\arraybackslash}p{\JEValCompactCol}>{\raggedleft\arraybackslash}p{\JEValCompactCol}>{\raggedleft\arraybackslash}p{\JEValCompactCol}>{\raggedleft\arraybackslash}p{\JEValCompactCol}>{\raggedleft\arraybackslash}p{\JEValCompactCol}>{\raggedleft\arraybackslash}p{\JEValCompactCol}>{\raggedleft\arraybackslash}p{\JEValCompactCol}@{}}
\caption{Medicine: model band 2/3. ACC (\%) $\uparrow$; benchmark-run latency (s), ECE, and $\JEValBrierRaw$ $\downarrow$. Rankings span all bands.}\label{tab:domain-medicine-band-2}\\
\toprule
\multicolumn{9}{@{}l@{}}{\fontsize{7.3}{8.4}\selectfont\JEValModeKey}\\[1.5pt]
\JEValHeader{\textbf{Metric}}{} & \JEValHeader{Intern-Decision\\2B}{} & \JEValHeader{Intern-Decision\\4B}{} & \JEValHeader{laya}{} & \JEValHeader{openjev-4b-v5}{} & \JEValHeader{nimble-9b}{} & \JEValHeader{nanojev}{} & \JEValHeader{lightjev}{} & \JEValHeader{jevforge-0.8b}{}\\\midrule
\endfirsthead
\multicolumn{9}{@{}l}{\textit{Table \thetable{} continued; model band 2/3}}\\\toprule
\multicolumn{9}{@{}l@{}}{\fontsize{7.3}{8.4}\selectfont\JEValModeKey}\\[1.5pt]
\JEValHeader{\textbf{Metric}}{} & \JEValHeader{Intern-Decision\\2B}{} & \JEValHeader{Intern-Decision\\4B}{} & \JEValHeader{laya}{} & \JEValHeader{openjev-4b-v5}{} & \JEValHeader{nimble-9b}{} & \JEValHeader{nanojev}{} & \JEValHeader{lightjev}{} & \JEValHeader{jevforge-0.8b}{}\\\midrule
\endhead
\midrule\multicolumn{9}{r@{}}{\fontsize{7.3}{8.4}\selectfont Continued on next page}\\\endfoot
\bottomrule\endlastfoot
\multicolumn{9}{@{}l@{}}{\textbf{cmb}\hfill $N=200$}\\*
ACC $\uparrow$ & 45.00 & 67.00 & 26.00 & 63.00 & 75.50 & 20.00 & 33.50 & 24.00\\*
Latency $\downarrow$ & 6.35 & 9.72 & 12.62 & 233.39 & 23.79 & 39.32 & 19.22 & 221.44\\*
ECE $\downarrow$ & 0.112 & 0.056 & 0.133 & 0.140 & 0.112 & 0.092 & 0.090 & 0.078\\*
$\JEValBrierRaw\,\downarrow$ & 0.672 & 0.460 & 0.838 & 0.529 & 0.342 & 0.816 & 0.814 & 0.810\\[1.5pt]
\multicolumn{9}{@{}l@{}}{\textbf{cmexam}\hfill $N=200$}\\*
ACC $\uparrow$ & 56.00 & 67.00 & 20.50 & 66.50 & 81.50 & 16.50 & 16.50 & 19.50\\*
Latency $\downarrow$ & 6.31 & 9.77 & 11.89 & 229.51 & 23.44 & 39.29 & 16.92 & 217.34\\
\end{longtable}
\endgroup

\begingroup\def\rmdefault{lmr}\normalfont\fontsize{8}{9.3}\selectfont\JEValSetTableWidths{10}\renewcommand{\arraystretch}{1.0}
\begin{longtable}{@{}>{\raggedright\arraybackslash}p{\JEValCompactCol}>{\raggedleft\arraybackslash}p{\JEValCompactCol}>{\raggedleft\arraybackslash}p{\JEValCompactCol}>{\raggedleft\arraybackslash}p{\JEValCompactCol}>{\raggedleft\arraybackslash}p{\JEValCompactCol}>{\raggedleft\arraybackslash}p{\JEValCompactCol}>{\raggedleft\arraybackslash}p{\JEValCompactCol}>{\raggedleft\arraybackslash}p{\JEValCompactCol}>{\raggedleft\arraybackslash}p{\JEValCompactCol}>{\raggedleft\arraybackslash}p{\JEValCompactCol}@{}}
\caption{Medicine: model band 3/3. ACC (\%) $\uparrow$; benchmark-run latency (s), ECE, and $\JEValBrierRaw$ $\downarrow$. Rankings span all bands.}\label{tab:domain-medicine-band-3}\\
\toprule
\multicolumn{10}{@{}l@{}}{\fontsize{7.3}{8.4}\selectfont\JEValModeKey}\\[1.5pt]
\JEValHeader{\textbf{Metric}}{} & \JEValHeader{Qwen3.5-4B}{N} & \JEValHeader{Qwen3.8-27B}{N} & \JEValHeader{DeepSeek\\V4-Flash}{N} & \JEValHeader{GPT-6-Sol}{N} & \JEValHeader{GPT-6-Luna}{N} & \JEValHeader{Qwen3.5-4B}{T} & \JEValHeader{Qwen3.8-27B}{T (xhigh)} & \JEValHeader{DeepSeek\\V4-Flash}{T (high)} & \JEValHeader{GLM-5.3-Flash}{T (high)}\\\midrule
\endfirsthead
\multicolumn{10}{@{}l}{\textit{Table \thetable{} continued; model band 3/3}}\\\toprule
\multicolumn{10}{@{}l@{}}{\fontsize{7.3}{8.4}\selectfont\JEValModeKey}\\[1.5pt]
\JEValHeader{\textbf{Metric}}{} & \JEValHeader{Qwen3.5-4B}{N} & \JEValHeader{Qwen3.8-27B}{N} & \JEValHeader{DeepSeek\\V4-Flash}{N} & \JEValHeader{GPT-6-Sol}{N} & \JEValHeader{GPT-6-Luna}{N} & \JEValHeader{Qwen3.5-4B}{T} & \JEValHeader{Qwen3.8-27B}{T (xhigh)} & \JEValHeader{DeepSeek\\V4-Flash}{T (high)} & \JEValHeader{GLM-5.3-Flash}{T (high)}\\\midrule
\endhead
\midrule\multicolumn{10}{r@{}}{\fontsize{7.3}{8.4}\selectfont Continued on next page}\\\endfoot
\bottomrule\endlastfoot
\multicolumn{10}{@{}l@{}}{\textbf{cmb}\hfill $N=200$}\\*
ACC $\uparrow$ & 60.50 & 72.50 & 82.00 & \underline{90.50} & \underline{90.50} & 67.00 & 87.00 & \underline{90.50} & \textbf{91.50}\\*
Latency $\downarrow$ & \underline{0.98} & 3.09 & 3.23 & 12.59 & 5.63 & 30.37 & 96.36 & 63.81 & 11.64\\*
ECE $\downarrow$ & 0.273 & 0.184 & 0.106 & 0.077 & 0.075 & 0.170 & 0.066 & 0.088 & \textbf{0.031}\\*
$\JEValBrierRaw\,\downarrow$ & 0.698 & 0.496 & 0.295 & 0.181 & \underline{0.179} & 0.488 & 0.216 & 0.180 & \textbf{0.142}\\[1.5pt]
\multicolumn{10}{@{}l@{}}{\textbf{cmexam}\hfill $N=200$}\\*
ACC $\uparrow$ & 64.50 & 74.50 & 82.00 & 90.50 & \underline{93.00} & 78.50 & 88.00 & 91.00 & \textbf{95.00}\\*
Latency $\downarrow$ & \underline{2.45} & 4.81 & 4.10 & 12.47 & 5.16 & 32.01 & 93.65 & 60.85 & 11.32\\
\end{longtable}
\endgroup

\Needspace{100pt}
\subsection{Law}
\label{app:domain-law}
\begingroup\def\rmdefault{lmr}\normalfont\fontsize{8}{9.3}\selectfont\JEValSetTableWidths{9}\renewcommand{\arraystretch}{1.0}
\begin{longtable}{@{}>{\raggedright\arraybackslash}p{\JEValCompactCol}>{\raggedleft\arraybackslash}p{\JEValCompactCol}>{\raggedleft\arraybackslash}p{\JEValCompactCol}>{\raggedleft\arraybackslash}p{\JEValCompactCol}>{\raggedleft\arraybackslash}p{\JEValCompactCol}>{\raggedleft\arraybackslash}p{\JEValCompactCol}>{\raggedleft\arraybackslash}p{\JEValCompactCol}>{\raggedleft\arraybackslash}p{\JEValCompactCol}>{\raggedleft\arraybackslash}p{\JEValCompactCol}@{}}
\caption{Law: model band 1/3. ACC (\%) $\uparrow$; benchmark-run latency (s), ECE, and $\JEValBrierRaw$ $\downarrow$. Rankings span all bands.}\label{tab:domain-law}\\
\toprule
\multicolumn{9}{@{}l@{}}{\fontsize{7.3}{8.4}\selectfont\JEValModeKey}\\[1.5pt]
\JEValHeader{\textbf{Metric}}{} & \JEValHeader{Jev}{} & \JEValHeader{SII-Decision-4B}{(ours)} & \JEValHeader{SII-Decision-27B}{(ours)} & \JEValHeader{kev-0.8b}{} & \JEValHeader{kev-4b}{} & \JEValHeader{kev-9b}{} & \JEValHeader{kev-27b}{} & \JEValHeader{Intern-Decision\\0.8B}{}\\\midrule
\endfirsthead
\multicolumn{9}{@{}l}{\textit{Table \thetable{} continued; model band 1/3}}\\\toprule
\multicolumn{9}{@{}l@{}}{\fontsize{7.3}{8.4}\selectfont\JEValModeKey}\\[1.5pt]
\JEValHeader{\textbf{Metric}}{} & \JEValHeader{Jev}{} & \JEValHeader{SII-Decision-4B}{(ours)} & \JEValHeader{SII-Decision-27B}{(ours)} & \JEValHeader{kev-0.8b}{} & \JEValHeader{kev-4b}{} & \JEValHeader{kev-9b}{} & \JEValHeader{kev-27b}{} & \JEValHeader{Intern-Decision\\0.8B}{}\\\midrule
\endhead
\midrule\multicolumn{9}{r@{}}{\fontsize{7.3}{8.4}\selectfont Continued on next page}\\\endfoot
\bottomrule\endlastfoot
\multicolumn{9}{@{}l@{}}{\textbf{cuad}\hfill $N=200$}\\*
ACC $\uparrow$ & 55.50 & 81.00 & \underline{83.50} & 77.00 & \underline{83.50} & 83.00 & 81.50 & 69.00\\*
Latency $\downarrow$ & \textbf{1.27} & 6.08 & 14.37 & 17.70 & 58.57 & 95.20 & 56.49 & 6.58\\*
ECE $\downarrow$ & 0.035 & 0.046 & 0.052 & 0.181 & 0.123 & 0.079 & \underline{0.030} & 0.143\\*
$\JEValBrierRaw\,\downarrow$ & 0.122 & 0.153 & \textbf{0.108} & 0.283 & 0.159 & 0.146 & \underline{0.118} & 0.350\\[1.5pt]
\multicolumn{9}{@{}l@{}}{\textbf{disc-law-multi}\hfill $N=200$}\\*
ACC $\uparrow$ & 47.50 & 19.00 & 45.00 & 5.00 & 12.00 & 28.50 & 42.50 & 2.50\\*
Latency $\downarrow$ & \textbf{0.89} & 5.64 & 12.24 & 16.68 & 57.83 & 93.77 & 54.37 & 6.38\\
\end{longtable}
\endgroup

\begingroup\def\rmdefault{lmr}\normalfont\fontsize{8}{9.3}\selectfont\JEValSetTableWidths{9}\renewcommand{\arraystretch}{1.0}
\begin{longtable}{@{}>{\raggedright\arraybackslash}p{\JEValCompactCol}>{\raggedleft\arraybackslash}p{\JEValCompactCol}>{\raggedleft\arraybackslash}p{\JEValCompactCol}>{\raggedleft\arraybackslash}p{\JEValCompactCol}>{\raggedleft\arraybackslash}p{\JEValCompactCol}>{\raggedleft\arraybackslash}p{\JEValCompactCol}>{\raggedleft\arraybackslash}p{\JEValCompactCol}>{\raggedleft\arraybackslash}p{\JEValCompactCol}>{\raggedleft\arraybackslash}p{\JEValCompactCol}@{}}
\caption{Law: model band 2/3. ACC (\%) $\uparrow$; benchmark-run latency (s), ECE, and $\JEValBrierRaw$ $\downarrow$. Rankings span all bands.}\label{tab:domain-law-band-2}\\
\toprule
\multicolumn{9}{@{}l@{}}{\fontsize{7.3}{8.4}\selectfont\JEValModeKey}\\[1.5pt]
\JEValHeader{\textbf{Metric}}{} & \JEValHeader{Intern-Decision\\2B}{} & \JEValHeader{Intern-Decision\\4B}{} & \JEValHeader{laya}{} & \JEValHeader{openjev-4b-v5}{} & \JEValHeader{nimble-9b}{} & \JEValHeader{nanojev}{} & \JEValHeader{lightjev}{} & \JEValHeader{jevforge-0.8b}{}\\\midrule
\endfirsthead
\multicolumn{9}{@{}l}{\textit{Table \thetable{} continued; model band 2/3}}\\\toprule
\multicolumn{9}{@{}l@{}}{\fontsize{7.3}{8.4}\selectfont\JEValModeKey}\\[1.5pt]
\JEValHeader{\textbf{Metric}}{} & \JEValHeader{Intern-Decision\\2B}{} & \JEValHeader{Intern-Decision\\4B}{} & \JEValHeader{laya}{} & \JEValHeader{openjev-4b-v5}{} & \JEValHeader{nimble-9b}{} & \JEValHeader{nanojev}{} & \JEValHeader{lightjev}{} & \JEValHeader{jevforge-0.8b}{}\\\midrule
\endhead
\midrule\multicolumn{9}{r@{}}{\fontsize{7.3}{8.4}\selectfont Continued on next page}\\\endfoot
\bottomrule\endlastfoot
\multicolumn{9}{@{}l@{}}{\textbf{cuad}\hfill $N=200$}\\*
ACC $\uparrow$ & 75.00 & 81.00 & 18.50 & \textbf{88.00} & 83.00 & 22.00 & 22.00 & 67.50\\*
Latency $\downarrow$ & 6.55 & 9.77 & 12.22 & 232.10 & 24.25 & 70.99 & 229.12 & 217.61\\*
ECE $\downarrow$ & 0.035 & \textbf{0.019} & 0.395 & 0.125 & 0.036 & 0.518 & 0.340 & 0.182\\*
$\JEValBrierRaw\,\downarrow$ & 0.233 & 0.158 & 0.731 & 0.246 & 0.133 & 0.912 & 0.606 & 0.433\\[1.5pt]
\multicolumn{9}{@{}l@{}}{\textbf{disc-law-multi}\hfill $N=200$}\\*
ACC $\uparrow$ & 5.00 & 18.00 & 1.50 & 18.50 & 33.50 & 9.00 & 15.00 & 0.00\\*
Latency $\downarrow$ & 6.32 & 9.34 & 11.47 & 229.32 & 23.61 & 36.99 & 39.31 & 218.35\\
\end{longtable}
\endgroup

\begingroup\def\rmdefault{lmr}\normalfont\fontsize{8}{9.3}\selectfont\JEValSetTableWidths{10}\renewcommand{\arraystretch}{1.0}
\begin{longtable}{@{}>{\raggedright\arraybackslash}p{\JEValCompactCol}>{\raggedleft\arraybackslash}p{\JEValCompactCol}>{\raggedleft\arraybackslash}p{\JEValCompactCol}>{\raggedleft\arraybackslash}p{\JEValCompactCol}>{\raggedleft\arraybackslash}p{\JEValCompactCol}>{\raggedleft\arraybackslash}p{\JEValCompactCol}>{\raggedleft\arraybackslash}p{\JEValCompactCol}>{\raggedleft\arraybackslash}p{\JEValCompactCol}>{\raggedleft\arraybackslash}p{\JEValCompactCol}>{\raggedleft\arraybackslash}p{\JEValCompactCol}@{}}
\caption{Law: model band 3/3. ACC (\%) $\uparrow$; benchmark-run latency (s), ECE, and $\JEValBrierRaw$ $\downarrow$. Rankings span all bands.}\label{tab:domain-law-band-3}\\
\toprule
\multicolumn{10}{@{}l@{}}{\fontsize{7.3}{8.4}\selectfont\JEValModeKey}\\[1.5pt]
\JEValHeader{\textbf{Metric}}{} & \JEValHeader{Qwen3.5-4B}{N} & \JEValHeader{Qwen3.8-27B}{N} & \JEValHeader{DeepSeek\\V4-Flash}{N} & \JEValHeader{GPT-6-Sol}{N} & \JEValHeader{GPT-6-Luna}{N} & \JEValHeader{Qwen3.5-4B}{T} & \JEValHeader{Qwen3.8-27B}{T (xhigh)} & \JEValHeader{DeepSeek\\V4-Flash}{T (high)} & \JEValHeader{GLM-5.3-Flash}{T (high)}\\\midrule
\endfirsthead
\multicolumn{10}{@{}l}{\textit{Table \thetable{} continued; model band 3/3}}\\\toprule
\multicolumn{10}{@{}l@{}}{\fontsize{7.3}{8.4}\selectfont\JEValModeKey}\\[1.5pt]
\JEValHeader{\textbf{Metric}}{} & \JEValHeader{Qwen3.5-4B}{N} & \JEValHeader{Qwen3.8-27B}{N} & \JEValHeader{DeepSeek\\V4-Flash}{N} & \JEValHeader{GPT-6-Sol}{N} & \JEValHeader{GPT-6-Luna}{N} & \JEValHeader{Qwen3.5-4B}{T} & \JEValHeader{Qwen3.8-27B}{T (xhigh)} & \JEValHeader{DeepSeek\\V4-Flash}{T (high)} & \JEValHeader{GLM-5.3-Flash}{T (high)}\\\midrule
\endhead
\midrule\multicolumn{10}{r@{}}{\fontsize{7.3}{8.4}\selectfont Continued on next page}\\\endfoot
\bottomrule\endlastfoot
\multicolumn{10}{@{}l@{}}{\textbf{cuad}\hfill $N=200$}\\*
ACC $\uparrow$ & 70.00 & 77.00 & 76.00 & 63.00 & 80.50 & 62.50 & 73.50 & 80.50 & 76.00\\*
Latency $\downarrow$ & \underline{2.56} & 13.36 & 15.48 & 13.76 & 4.72 & 43.12 & 106.46 & 50.38 & 8.51\\*
ECE $\downarrow$ & 0.196 & 0.114 & 0.142 & 0.282 & 0.092 & 0.271 & 0.146 & 0.099 & 0.099\\*
$\JEValBrierRaw\,\downarrow$ & 0.379 & 0.262 & 0.287 & 0.566 & 0.196 & 0.548 & 0.322 & 0.201 & 0.234\\[1.5pt]
\multicolumn{10}{@{}l@{}}{\textbf{disc-law-multi}\hfill $N=200$}\\*
ACC $\uparrow$ & 15.00 & 35.50 & 44.50 & \underline{70.00} & 68.50 & 34.00 & 61.50 & \textbf{73.00} & 62.00\\*
Latency $\downarrow$ & \underline{3.15} & 6.35 & 5.61 & 19.04 & 10.05 & 38.88 & 159.62 & 78.53 & 38.92\\
\end{longtable}
\endgroup

\Needspace{100pt}
\subsection{Finance}
\label{app:domain-finance}
\begingroup\def\rmdefault{lmr}\normalfont\fontsize{8}{9.3}\selectfont\JEValSetTableWidths{9}\renewcommand{\arraystretch}{1.0}
\begin{longtable}{@{}>{\raggedright\arraybackslash}p{\JEValCompactCol}>{\raggedleft\arraybackslash}p{\JEValCompactCol}>{\raggedleft\arraybackslash}p{\JEValCompactCol}>{\raggedleft\arraybackslash}p{\JEValCompactCol}>{\raggedleft\arraybackslash}p{\JEValCompactCol}>{\raggedleft\arraybackslash}p{\JEValCompactCol}>{\raggedleft\arraybackslash}p{\JEValCompactCol}>{\raggedleft\arraybackslash}p{\JEValCompactCol}>{\raggedleft\arraybackslash}p{\JEValCompactCol}@{}}
\caption{Finance: model band 1/3. ACC (\%) $\uparrow$; benchmark-run latency (s), ECE, and $\JEValBrierRaw$ $\downarrow$. Rankings span all bands.}\label{tab:domain-finance}\\
\toprule
\multicolumn{9}{@{}l@{}}{\fontsize{7.3}{8.4}\selectfont\JEValModeKey}\\[1.5pt]
\JEValHeader{\textbf{Metric}}{} & \JEValHeader{Jev}{} & \JEValHeader{SII-Decision-4B}{(ours)} & \JEValHeader{SII-Decision-27B}{(ours)} & \JEValHeader{kev-0.8b}{} & \JEValHeader{kev-4b}{} & \JEValHeader{kev-9b}{} & \JEValHeader{kev-27b}{} & \JEValHeader{Intern-Decision\\0.8B}{}\\\midrule
\endfirsthead
\multicolumn{9}{@{}l}{\textit{Table \thetable{} continued; model band 1/3}}\\\toprule
\multicolumn{9}{@{}l@{}}{\fontsize{7.3}{8.4}\selectfont\JEValModeKey}\\[1.5pt]
\JEValHeader{\textbf{Metric}}{} & \JEValHeader{Jev}{} & \JEValHeader{SII-Decision-4B}{(ours)} & \JEValHeader{SII-Decision-27B}{(ours)} & \JEValHeader{kev-0.8b}{} & \JEValHeader{kev-4b}{} & \JEValHeader{kev-9b}{} & \JEValHeader{kev-27b}{} & \JEValHeader{Intern-Decision\\0.8B}{}\\\midrule
\endhead
\midrule\multicolumn{9}{r@{}}{\fontsize{7.3}{8.4}\selectfont Continued on next page}\\\endfoot
\bottomrule\endlastfoot
\multicolumn{9}{@{}l@{}}{\textbf{finqa-yesno}\hfill $N=22$}\\*
ACC $\uparrow$ & \textbf{95.45} & 72.73 & 86.36 & 68.18 & 81.82 & 86.36 & \underline{90.91} & 54.55\\*
Latency $\downarrow$ & \textbf{0.59} & 4.94 & 11.34 & 11.29 & 42.06 & 62.85 & 36.49 & 6.61\\*
ECE $\downarrow$ & 0.051 & 0.194 & 0.098 & 0.113 & 0.082 & 0.079 & 0.056 & 0.120\\*
$\JEValBrierRaw\,\downarrow$ & \underline{0.102} & 0.263 & 0.193 & 0.336 & 0.227 & 0.162 & 0.122 & 0.439\\[1.5pt]
\multicolumn{9}{@{}l@{}}{\textbf{financebench-yesno}\hfill $N=37$}\\*
ACC $\uparrow$ & \underline{86.49} & 75.68 & 81.08 & 64.86 & 78.38 & 81.08 & \underline{86.49} & 59.46\\*
Latency $\downarrow$ & \textbf{0.68} & 5.18 & 9.39 & 12.18 & 37.46 & 71.96 & 44.21 & 6.58\\
\end{longtable}
\endgroup

\begingroup\def\rmdefault{lmr}\normalfont\fontsize{8}{9.3}\selectfont\JEValSetTableWidths{9}\renewcommand{\arraystretch}{1.0}
\begin{longtable}{@{}>{\raggedright\arraybackslash}p{\JEValCompactCol}>{\raggedleft\arraybackslash}p{\JEValCompactCol}>{\raggedleft\arraybackslash}p{\JEValCompactCol}>{\raggedleft\arraybackslash}p{\JEValCompactCol}>{\raggedleft\arraybackslash}p{\JEValCompactCol}>{\raggedleft\arraybackslash}p{\JEValCompactCol}>{\raggedleft\arraybackslash}p{\JEValCompactCol}>{\raggedleft\arraybackslash}p{\JEValCompactCol}>{\raggedleft\arraybackslash}p{\JEValCompactCol}@{}}
\caption{Finance: model band 2/3. ACC (\%) $\uparrow$; benchmark-run latency (s), ECE, and $\JEValBrierRaw$ $\downarrow$. Rankings span all bands.}\label{tab:domain-finance-band-2}\\
\toprule
\multicolumn{9}{@{}l@{}}{\fontsize{7.3}{8.4}\selectfont\JEValModeKey}\\[1.5pt]
\JEValHeader{\textbf{Metric}}{} & \JEValHeader{Intern-Decision\\2B}{} & \JEValHeader{Intern-Decision\\4B}{} & \JEValHeader{laya}{} & \JEValHeader{openjev-4b-v5}{} & \JEValHeader{nimble-9b}{} & \JEValHeader{nanojev}{} & \JEValHeader{lightjev}{} & \JEValHeader{jevforge-0.8b}{}\\\midrule
\endfirsthead
\multicolumn{9}{@{}l}{\textit{Table \thetable{} continued; model band 2/3}}\\\toprule
\multicolumn{9}{@{}l@{}}{\fontsize{7.3}{8.4}\selectfont\JEValModeKey}\\[1.5pt]
\JEValHeader{\textbf{Metric}}{} & \JEValHeader{Intern-Decision\\2B}{} & \JEValHeader{Intern-Decision\\4B}{} & \JEValHeader{laya}{} & \JEValHeader{openjev-4b-v5}{} & \JEValHeader{nimble-9b}{} & \JEValHeader{nanojev}{} & \JEValHeader{lightjev}{} & \JEValHeader{jevforge-0.8b}{}\\\midrule
\endhead
\midrule\multicolumn{9}{r@{}}{\fontsize{7.3}{8.4}\selectfont Continued on next page}\\\endfoot
\bottomrule\endlastfoot
\multicolumn{9}{@{}l@{}}{\textbf{finqa-yesno}\hfill $N=22$}\\*
ACC $\uparrow$ & 81.82 & 81.82 & 68.18 & \underline{90.91} & 86.36 & 59.09 & 54.55 & 36.36\\*
Latency $\downarrow$ & 5.37 & 8.31 & 12.49 & 99.97 & 16.88 & 25.08 & 44.75 & 68.59\\*
ECE $\downarrow$ & 0.236 & 0.135 & 0.275 & 0.156 & 0.069 & 0.103 & \textbf{0.023} & 0.589\\*
$\JEValBrierRaw\,\downarrow$ & 0.400 & 0.267 & 0.433 & 0.178 & 0.202 & 0.509 & 0.489 & 1.147\\[1.5pt]
\multicolumn{9}{@{}l@{}}{\textbf{financebench-yesno}\hfill $N=37$}\\*
ACC $\uparrow$ & 62.16 & 78.38 & 56.76 & 83.78 & 81.08 & 72.97 & 43.24 & 29.73\\*
Latency $\downarrow$ & 5.28 & 9.11 & 14.07 & 157.43 & 18.99 & 33.57 & 44.62 & 133.02\\
\end{longtable}
\endgroup

\begingroup\def\rmdefault{lmr}\normalfont\fontsize{8}{9.3}\selectfont\JEValSetTableWidths{10}\renewcommand{\arraystretch}{1.0}
\begin{longtable}{@{}>{\raggedright\arraybackslash}p{\JEValCompactCol}>{\raggedleft\arraybackslash}p{\JEValCompactCol}>{\raggedleft\arraybackslash}p{\JEValCompactCol}>{\raggedleft\arraybackslash}p{\JEValCompactCol}>{\raggedleft\arraybackslash}p{\JEValCompactCol}>{\raggedleft\arraybackslash}p{\JEValCompactCol}>{\raggedleft\arraybackslash}p{\JEValCompactCol}>{\raggedleft\arraybackslash}p{\JEValCompactCol}>{\raggedleft\arraybackslash}p{\JEValCompactCol}>{\raggedleft\arraybackslash}p{\JEValCompactCol}@{}}
\caption{Finance: model band 3/3. ACC (\%) $\uparrow$; benchmark-run latency (s), ECE, and $\JEValBrierRaw$ $\downarrow$. Rankings span all bands.}\label{tab:domain-finance-band-3}\\
\toprule
\multicolumn{10}{@{}l@{}}{\fontsize{7.3}{8.4}\selectfont\JEValModeKey}\\[1.5pt]
\JEValHeader{\textbf{Metric}}{} & \JEValHeader{Qwen3.5-4B}{N} & \JEValHeader{Qwen3.8-27B}{N} & \JEValHeader{DeepSeek\\V4-Flash}{N} & \JEValHeader{GPT-6-Sol}{N} & \JEValHeader{GPT-6-Luna}{N} & \JEValHeader{Qwen3.5-4B}{T} & \JEValHeader{Qwen3.8-27B}{T (xhigh)} & \JEValHeader{DeepSeek\\V4-Flash}{T (high)} & \JEValHeader{GLM-5.3-Flash}{T (high)}\\\midrule
\endfirsthead
\multicolumn{10}{@{}l}{\textit{Table \thetable{} continued; model band 3/3}}\\\toprule
\multicolumn{10}{@{}l@{}}{\fontsize{7.3}{8.4}\selectfont\JEValModeKey}\\[1.5pt]
\JEValHeader{\textbf{Metric}}{} & \JEValHeader{Qwen3.5-4B}{N} & \JEValHeader{Qwen3.8-27B}{N} & \JEValHeader{DeepSeek\\V4-Flash}{N} & \JEValHeader{GPT-6-Sol}{N} & \JEValHeader{GPT-6-Luna}{N} & \JEValHeader{Qwen3.5-4B}{T} & \JEValHeader{Qwen3.8-27B}{T (xhigh)} & \JEValHeader{DeepSeek\\V4-Flash}{T (high)} & \JEValHeader{GLM-5.3-Flash}{T (high)}\\\midrule
\endhead
\midrule\multicolumn{10}{r@{}}{\fontsize{7.3}{8.4}\selectfont Continued on next page}\\\endfoot
\bottomrule\endlastfoot
\multicolumn{10}{@{}l@{}}{\textbf{finqa-yesno}\hfill $N=22$}\\*
ACC $\uparrow$ & 59.09 & 86.36 & 81.82 & 81.82 & \textbf{95.45} & 77.27 & 81.82 & \textbf{95.45} & \underline{90.91}\\*
Latency $\downarrow$ & \underline{2.67} & 6.25 & 68.36 & 19.01 & 5.09 & 58.64 & 72.74 & 182.33 & 11.23\\*
ECE $\downarrow$ & 0.417 & 0.126 & 0.190 & 0.177 & \underline{0.045} & 0.210 & 0.193 & \underline{0.045} & 0.091\\*
$\JEValBrierRaw\,\downarrow$ & 0.773 & 0.261 & 0.366 & 0.358 & \textbf{0.091} & 0.439 & 0.362 & \textbf{0.091} & 0.180\\[1.5pt]
\multicolumn{10}{@{}l@{}}{\textbf{financebench-yesno}\hfill $N=37$}\\*
ACC $\uparrow$ & 72.97 & 83.78 & 67.57 & 83.78 & \textbf{94.59} & 78.38 & 83.78 & 81.08 & \textbf{94.59}\\*
Latency $\downarrow$ & \underline{2.73} & 6.10 & 67.78 & 11.22 & 8.08 & 50.30 & 97.53 & 175.09 & 12.11\\
\end{longtable}
\endgroup

\Needspace{100pt}
\subsection{Commonsense}
\label{app:domain-commonsense}
\begingroup\def\rmdefault{lmr}\normalfont\fontsize{8}{9.3}\selectfont\JEValSetTableWidths{9}\renewcommand{\arraystretch}{1.0}
\begin{longtable}{@{}>{\raggedright\arraybackslash}p{\JEValCompactCol}>{\raggedleft\arraybackslash}p{\JEValCompactCol}>{\raggedleft\arraybackslash}p{\JEValCompactCol}>{\raggedleft\arraybackslash}p{\JEValCompactCol}>{\raggedleft\arraybackslash}p{\JEValCompactCol}>{\raggedleft\arraybackslash}p{\JEValCompactCol}>{\raggedleft\arraybackslash}p{\JEValCompactCol}>{\raggedleft\arraybackslash}p{\JEValCompactCol}>{\raggedleft\arraybackslash}p{\JEValCompactCol}@{}}
\caption{Commonsense: model band 1/3. ACC (\%) $\uparrow$; benchmark-run latency (s), ECE, and $\JEValBrierRaw$ $\downarrow$. Rankings span all bands.}\label{tab:domain-commonsense}\\
\toprule
\multicolumn{9}{@{}l@{}}{\fontsize{7.3}{8.4}\selectfont\JEValModeKey}\\[1.5pt]
\JEValHeader{\textbf{Metric}}{} & \JEValHeader{Jev}{} & \JEValHeader{SII-Decision-4B}{(ours)} & \JEValHeader{SII-Decision-27B}{(ours)} & \JEValHeader{kev-0.8b}{} & \JEValHeader{kev-4b}{} & \JEValHeader{kev-9b}{} & \JEValHeader{kev-27b}{} & \JEValHeader{Intern-Decision\\0.8B}{}\\\midrule
\endfirsthead
\multicolumn{9}{@{}l}{\textit{Table \thetable{} continued; model band 1/3}}\\\toprule
\multicolumn{9}{@{}l@{}}{\fontsize{7.3}{8.4}\selectfont\JEValModeKey}\\[1.5pt]
\JEValHeader{\textbf{Metric}}{} & \JEValHeader{Jev}{} & \JEValHeader{SII-Decision-4B}{(ours)} & \JEValHeader{SII-Decision-27B}{(ours)} & \JEValHeader{kev-0.8b}{} & \JEValHeader{kev-4b}{} & \JEValHeader{kev-9b}{} & \JEValHeader{kev-27b}{} & \JEValHeader{Intern-Decision\\0.8B}{}\\\midrule
\endhead
\midrule\multicolumn{9}{r@{}}{\fontsize{7.3}{8.4}\selectfont Continued on next page}\\\endfoot
\bottomrule\endlastfoot
\multicolumn{9}{@{}l@{}}{\textbf{commonsense-qa}\hfill $N=200$}\\*
ACC $\uparrow$ & 89.00 & 82.50 & \textbf{91.00} & 56.50 & 83.50 & 80.50 & 88.50 & 54.50\\*
Latency $\downarrow$ & \textbf{0.49} & 5.61 & 12.55 & 17.44 & 57.00 & 93.18 & 54.62 & 6.54\\*
ECE $\downarrow$ & 0.043 & 0.060 & 0.072 & \underline{0.038} & 0.161 & 0.145 & 0.157 & 0.133\\*
$\JEValBrierRaw\,\downarrow$ & 0.172 & 0.244 & \textbf{0.160} & 0.560 & 0.295 & 0.287 & 0.214 & 0.615\\[1.5pt]
\multicolumn{9}{@{}l@{}}{\textbf{hellaswag}\hfill $N=200$}\\*
ACC $\uparrow$ & 93.50 & 85.00 & 93.50 & 54.00 & 83.00 & 85.00 & 91.50 & 51.00\\*
Latency $\downarrow$ & \textbf{0.69} & 5.65 & 11.70 & 16.91 & 55.51 & 92.12 & 54.15 & 6.59\\
\end{longtable}
\endgroup

\begingroup\def\rmdefault{lmr}\normalfont\fontsize{8}{9.3}\selectfont\JEValSetTableWidths{9}\renewcommand{\arraystretch}{1.0}
\begin{longtable}{@{}>{\raggedright\arraybackslash}p{\JEValCompactCol}>{\raggedleft\arraybackslash}p{\JEValCompactCol}>{\raggedleft\arraybackslash}p{\JEValCompactCol}>{\raggedleft\arraybackslash}p{\JEValCompactCol}>{\raggedleft\arraybackslash}p{\JEValCompactCol}>{\raggedleft\arraybackslash}p{\JEValCompactCol}>{\raggedleft\arraybackslash}p{\JEValCompactCol}>{\raggedleft\arraybackslash}p{\JEValCompactCol}>{\raggedleft\arraybackslash}p{\JEValCompactCol}@{}}
\caption{Commonsense: model band 2/3. ACC (\%) $\uparrow$; benchmark-run latency (s), ECE, and $\JEValBrierRaw$ $\downarrow$. Rankings span all bands.}\label{tab:domain-commonsense-band-2}\\
\toprule
\multicolumn{9}{@{}l@{}}{\fontsize{7.3}{8.4}\selectfont\JEValModeKey}\\[1.5pt]
\JEValHeader{\textbf{Metric}}{} & \JEValHeader{Intern-Decision\\2B}{} & \JEValHeader{Intern-Decision\\4B}{} & \JEValHeader{laya}{} & \JEValHeader{openjev-4b-v5}{} & \JEValHeader{nimble-9b}{} & \JEValHeader{nanojev}{} & \JEValHeader{lightjev}{} & \JEValHeader{jevforge-0.8b}{}\\\midrule
\endfirsthead
\multicolumn{9}{@{}l}{\textit{Table \thetable{} continued; model band 2/3}}\\\toprule
\multicolumn{9}{@{}l@{}}{\fontsize{7.3}{8.4}\selectfont\JEValModeKey}\\[1.5pt]
\JEValHeader{\textbf{Metric}}{} & \JEValHeader{Intern-Decision\\2B}{} & \JEValHeader{Intern-Decision\\4B}{} & \JEValHeader{laya}{} & \JEValHeader{openjev-4b-v5}{} & \JEValHeader{nimble-9b}{} & \JEValHeader{nanojev}{} & \JEValHeader{lightjev}{} & \JEValHeader{jevforge-0.8b}{}\\\midrule
\endhead
\midrule\multicolumn{9}{r@{}}{\fontsize{7.3}{8.4}\selectfont Continued on next page}\\\endfoot
\bottomrule\endlastfoot
\multicolumn{9}{@{}l@{}}{\textbf{commonsense-qa}\hfill $N=200$}\\*
ACC $\uparrow$ & 63.50 & 78.00 & 44.50 & 72.00 & 82.00 & 17.00 & 29.50 & 24.50\\*
Latency $\downarrow$ & 6.23 & 9.59 & 12.10 & 231.86 & 23.43 & 37.33 & 17.60 & 217.77\\*
ECE $\downarrow$ & 0.066 & 0.098 & 0.115 & 0.195 & 0.070 & 0.120 & 0.267 & 0.136\\*
$\JEValBrierRaw\,\downarrow$ & 0.500 & 0.332 & 0.709 & 0.441 & 0.255 & 0.815 & 0.951 & 0.824\\[1.5pt]
\multicolumn{9}{@{}l@{}}{\textbf{hellaswag}\hfill $N=200$}\\*
ACC $\uparrow$ & 61.50 & 79.00 & 19.50 & \textbf{95.50} & 89.50 & 25.50 & 33.00 & 30.50\\*
Latency $\downarrow$ & 6.16 & 9.87 & 11.45 & 231.78 & 23.75 & 37.17 & 18.20 & 219.47\\
\end{longtable}
\endgroup

\begingroup\def\rmdefault{lmr}\normalfont\fontsize{8}{9.3}\selectfont\JEValSetTableWidths{10}\renewcommand{\arraystretch}{1.0}
\begin{longtable}{@{}>{\raggedright\arraybackslash}p{\JEValCompactCol}>{\raggedleft\arraybackslash}p{\JEValCompactCol}>{\raggedleft\arraybackslash}p{\JEValCompactCol}>{\raggedleft\arraybackslash}p{\JEValCompactCol}>{\raggedleft\arraybackslash}p{\JEValCompactCol}>{\raggedleft\arraybackslash}p{\JEValCompactCol}>{\raggedleft\arraybackslash}p{\JEValCompactCol}>{\raggedleft\arraybackslash}p{\JEValCompactCol}>{\raggedleft\arraybackslash}p{\JEValCompactCol}>{\raggedleft\arraybackslash}p{\JEValCompactCol}@{}}
\caption{Commonsense: model band 3/3. ACC (\%) $\uparrow$; benchmark-run latency (s), ECE, and $\JEValBrierRaw$ $\downarrow$. Rankings span all bands.}\label{tab:domain-commonsense-band-3}\\
\toprule
\multicolumn{10}{@{}l@{}}{\fontsize{7.3}{8.4}\selectfont\JEValModeKey}\\[1.5pt]
\JEValHeader{\textbf{Metric}}{} & \JEValHeader{Qwen3.5-4B}{N} & \JEValHeader{Qwen3.8-27B}{N} & \JEValHeader{DeepSeek\\V4-Flash}{N} & \JEValHeader{GPT-6-Sol}{N} & \JEValHeader{GPT-6-Luna}{N} & \JEValHeader{Qwen3.5-4B}{T} & \JEValHeader{Qwen3.8-27B}{T (xhigh)} & \JEValHeader{DeepSeek\\V4-Flash}{T (high)} & \JEValHeader{GLM-5.3-Flash}{T (high)}\\\midrule
\endfirsthead
\multicolumn{10}{@{}l}{\textit{Table \thetable{} continued; model band 3/3}}\\\toprule
\multicolumn{10}{@{}l@{}}{\fontsize{7.3}{8.4}\selectfont\JEValModeKey}\\[1.5pt]
\JEValHeader{\textbf{Metric}}{} & \JEValHeader{Qwen3.5-4B}{N} & \JEValHeader{Qwen3.8-27B}{N} & \JEValHeader{DeepSeek\\V4-Flash}{N} & \JEValHeader{GPT-6-Sol}{N} & \JEValHeader{GPT-6-Luna}{N} & \JEValHeader{Qwen3.5-4B}{T} & \JEValHeader{Qwen3.8-27B}{T (xhigh)} & \JEValHeader{DeepSeek\\V4-Flash}{T (high)} & \JEValHeader{GLM-5.3-Flash}{T (high)}\\\midrule
\endhead
\midrule\multicolumn{10}{r@{}}{\fontsize{7.3}{8.4}\selectfont Continued on next page}\\\endfoot
\bottomrule\endlastfoot
\multicolumn{10}{@{}l@{}}{\textbf{commonsense-qa}\hfill $N=200$}\\*
ACC $\uparrow$ & 75.00 & 84.50 & 75.00 & 86.50 & 86.50 & 78.50 & \underline{90.50} & 85.50 & 85.00\\*
Latency $\downarrow$ & \underline{2.61} & 3.21 & 3.97 & 11.44 & 4.11 & 41.66 & 79.32 & 65.53 & 9.54\\*
ECE $\downarrow$ & 0.163 & 0.093 & 0.173 & 0.087 & 0.071 & 0.143 & \textbf{0.030} & 0.120 & 0.048\\*
$\JEValBrierRaw\,\downarrow$ & 0.429 & 0.275 & 0.438 & 0.227 & 0.208 & 0.393 & \underline{0.161} & 0.267 & 0.226\\[1.5pt]
\multicolumn{10}{@{}l@{}}{\textbf{hellaswag}\hfill $N=200$}\\*
ACC $\uparrow$ & 72.50 & 89.00 & 80.50 & \underline{94.00} & \textbf{95.50} & 79.50 & 89.50 & 87.00 & 90.00\\*
Latency $\downarrow$ & \underline{1.73} & 5.11 & 3.49 & 8.26 & 7.50 & 38.20 & 110.93 & 78.65 & 9.51\\
\end{longtable}
\endgroup

\Needspace{100pt}
\subsection{Agent and tool use}
\label{app:domain-agent-and-tool-use}
\begingroup\def\rmdefault{lmr}\normalfont\fontsize{8}{9.3}\selectfont\JEValSetTableWidths{9}\renewcommand{\arraystretch}{1.0}
\begin{longtable}{@{}>{\raggedright\arraybackslash}p{\JEValCompactCol}>{\raggedleft\arraybackslash}p{\JEValCompactCol}>{\raggedleft\arraybackslash}p{\JEValCompactCol}>{\raggedleft\arraybackslash}p{\JEValCompactCol}>{\raggedleft\arraybackslash}p{\JEValCompactCol}>{\raggedleft\arraybackslash}p{\JEValCompactCol}>{\raggedleft\arraybackslash}p{\JEValCompactCol}>{\raggedleft\arraybackslash}p{\JEValCompactCol}>{\raggedleft\arraybackslash}p{\JEValCompactCol}@{}}
\caption{Agent and tool use: model band 1/3. ACC (\%) $\uparrow$; benchmark-run latency (s), ECE, and $\JEValBrierRaw$ $\downarrow$. Rankings span all bands.}\label{tab:domain-agent-and-tool-use}\\
\toprule
\multicolumn{9}{@{}l@{}}{\fontsize{7.3}{8.4}\selectfont\JEValModeKey}\\[1.5pt]
\JEValHeader{\textbf{Metric}}{} & \JEValHeader{Jev}{} & \JEValHeader{SII-Decision-4B}{(ours)} & \JEValHeader{SII-Decision-27B}{(ours)} & \JEValHeader{kev-0.8b}{} & \JEValHeader{kev-4b}{} & \JEValHeader{kev-9b}{} & \JEValHeader{kev-27b}{} & \JEValHeader{Intern-Decision\\0.8B}{}\\\midrule
\endfirsthead
\multicolumn{9}{@{}l}{\textit{Table \thetable{} continued; model band 1/3}}\\\toprule
\multicolumn{9}{@{}l@{}}{\fontsize{7.3}{8.4}\selectfont\JEValModeKey}\\[1.5pt]
\JEValHeader{\textbf{Metric}}{} & \JEValHeader{Jev}{} & \JEValHeader{SII-Decision-4B}{(ours)} & \JEValHeader{SII-Decision-27B}{(ours)} & \JEValHeader{kev-0.8b}{} & \JEValHeader{kev-4b}{} & \JEValHeader{kev-9b}{} & \JEValHeader{kev-27b}{} & \JEValHeader{Intern-Decision\\0.8B}{}\\\midrule
\endhead
\midrule\multicolumn{9}{r@{}}{\fontsize{7.3}{8.4}\selectfont Continued on next page}\\\endfoot
\bottomrule\endlastfoot
\multicolumn{9}{@{}l@{}}{\textbf{agentharm}\hfill $N=97$, $N_{\mathrm{score}}=103$}\\*
ACC $\uparrow$ & 84.54 & 87.63 & 87.63 & 77.32 & 86.60 & 87.63 & \underline{89.69} & 75.26\\*
Latency $\downarrow$ & \textbf{0.79} & 5.61 & 12.26 & 18.20 & 56.38 & 95.20 & 56.37 & 6.54\\*
ECE $\downarrow$ & 0.068 & 0.077 & 0.069 & \underline{0.054} & 0.105 & 0.072 & \textbf{0.050} & 0.075\\*
$\JEValBrierRaw\,\downarrow$ & 0.201 & 0.215 & 0.200 & 0.328 & 0.239 & 0.213 & \textbf{0.158} & 0.315\\*
MAE $\downarrow$ & 0.672 & 0.853 & 0.700 & 0.919 & 0.760 & 0.742 & 0.650 & 0.988\\*
Valid score & 103/103 & 103/103 & 103/103 & 103/103 & 103/103 & 103/103 & 103/103 & 103/103\\[1.5pt]
\multicolumn{9}{@{}l@{}}{\textbf{decider}\hfill $N=200$}\\
\end{longtable}
\endgroup

\begingroup\def\rmdefault{lmr}\normalfont\fontsize{8}{9.3}\selectfont\JEValSetTableWidths{9}\renewcommand{\arraystretch}{1.0}
\begin{longtable}{@{}>{\raggedright\arraybackslash}p{\JEValCompactCol}>{\raggedleft\arraybackslash}p{\JEValCompactCol}>{\raggedleft\arraybackslash}p{\JEValCompactCol}>{\raggedleft\arraybackslash}p{\JEValCompactCol}>{\raggedleft\arraybackslash}p{\JEValCompactCol}>{\raggedleft\arraybackslash}p{\JEValCompactCol}>{\raggedleft\arraybackslash}p{\JEValCompactCol}>{\raggedleft\arraybackslash}p{\JEValCompactCol}>{\raggedleft\arraybackslash}p{\JEValCompactCol}@{}}
\caption{Agent and tool use: model band 2/3. ACC (\%) $\uparrow$; benchmark-run latency (s), ECE, and $\JEValBrierRaw$ $\downarrow$. Rankings span all bands.}\label{tab:domain-agent-and-tool-use-band-2}\\
\toprule
\multicolumn{9}{@{}l@{}}{\fontsize{7.3}{8.4}\selectfont\JEValModeKey}\\[1.5pt]
\JEValHeader{\textbf{Metric}}{} & \JEValHeader{Intern-Decision\\2B}{} & \JEValHeader{Intern-Decision\\4B}{} & \JEValHeader{laya}{} & \JEValHeader{openjev-4b-v5}{} & \JEValHeader{nimble-9b}{} & \JEValHeader{nanojev}{} & \JEValHeader{lightjev}{} & \JEValHeader{jevforge-0.8b}{}\\\midrule
\endfirsthead
\multicolumn{9}{@{}l}{\textit{Table \thetable{} continued; model band 2/3}}\\\toprule
\multicolumn{9}{@{}l@{}}{\fontsize{7.3}{8.4}\selectfont\JEValModeKey}\\[1.5pt]
\JEValHeader{\textbf{Metric}}{} & \JEValHeader{Intern-Decision\\2B}{} & \JEValHeader{Intern-Decision\\4B}{} & \JEValHeader{laya}{} & \JEValHeader{openjev-4b-v5}{} & \JEValHeader{nimble-9b}{} & \JEValHeader{nanojev}{} & \JEValHeader{lightjev}{} & \JEValHeader{jevforge-0.8b}{}\\\midrule
\endhead
\midrule\multicolumn{9}{r@{}}{\fontsize{7.3}{8.4}\selectfont Continued on next page}\\\endfoot
\bottomrule\endlastfoot
\multicolumn{9}{@{}l@{}}{\textbf{agentharm}\hfill $N=97$, $N_{\mathrm{score}}=103$}\\*
ACC $\uparrow$ & 85.57 & 73.20 & 42.27 & 82.47 & \textbf{90.72} & 52.58 & 48.45 & 47.42\\*
Latency $\downarrow$ & 6.38 & 9.94 & 12.12 & 230.13 & 23.67 & 39.07 & 19.53 & 224.29\\*
ECE $\downarrow$ & 0.074 & 0.113 & 0.328 & 0.112 & 0.078 & 0.330 & 0.111 & 0.480\\*
$\JEValBrierRaw\,\downarrow$ & 0.243 & 0.331 & 0.765 & 0.306 & \underline{0.194} & 0.711 & 0.522 & 0.956\\*
MAE $\downarrow$ & 0.770 & 0.963 & 1.308 & 0.631 & 0.510 & 1.558 & 1.384 & 1.423\\*
Valid score & 103/103 & 103/103 & 103/103 & 103/103 & 103/103 & 103/103 & 103/103 & 103/103\\[1.5pt]
\multicolumn{9}{@{}l@{}}{\textbf{decider}\hfill $N=200$}\\
\end{longtable}
\endgroup

\begingroup\def\rmdefault{lmr}\normalfont\fontsize{8}{9.3}\selectfont\JEValSetTableWidths{10}\renewcommand{\arraystretch}{1.0}
\begin{longtable}{@{}>{\raggedright\arraybackslash}p{\JEValCompactCol}>{\raggedleft\arraybackslash}p{\JEValCompactCol}>{\raggedleft\arraybackslash}p{\JEValCompactCol}>{\raggedleft\arraybackslash}p{\JEValCompactCol}>{\raggedleft\arraybackslash}p{\JEValCompactCol}>{\raggedleft\arraybackslash}p{\JEValCompactCol}>{\raggedleft\arraybackslash}p{\JEValCompactCol}>{\raggedleft\arraybackslash}p{\JEValCompactCol}>{\raggedleft\arraybackslash}p{\JEValCompactCol}>{\raggedleft\arraybackslash}p{\JEValCompactCol}@{}}
\caption{Agent and tool use: model band 3/3. ACC (\%) $\uparrow$; benchmark-run latency (s), ECE, and $\JEValBrierRaw$ $\downarrow$. Rankings span all bands.}\label{tab:domain-agent-and-tool-use-band-3}\\
\toprule
\multicolumn{10}{@{}l@{}}{\fontsize{7.3}{8.4}\selectfont\JEValModeKey}\\[1.5pt]
\JEValHeader{\textbf{Metric}}{} & \JEValHeader{Qwen3.5-4B}{N} & \JEValHeader{Qwen3.8-27B}{N} & \JEValHeader{DeepSeek\\V4-Flash}{N} & \JEValHeader{GPT-6-Sol}{N} & \JEValHeader{GPT-6-Luna}{N} & \JEValHeader{Qwen3.5-4B}{T} & \JEValHeader{Qwen3.8-27B}{T (xhigh)} & \JEValHeader{DeepSeek\\V4-Flash}{T (high)} & \JEValHeader{GLM-5.3-Flash}{T (high)}\\\midrule
\endfirsthead
\multicolumn{10}{@{}l}{\textit{Table \thetable{} continued; model band 3/3}}\\\toprule
\multicolumn{10}{@{}l@{}}{\fontsize{7.3}{8.4}\selectfont\JEValModeKey}\\[1.5pt]
\JEValHeader{\textbf{Metric}}{} & \JEValHeader{Qwen3.5-4B}{N} & \JEValHeader{Qwen3.8-27B}{N} & \JEValHeader{DeepSeek\\V4-Flash}{N} & \JEValHeader{GPT-6-Sol}{N} & \JEValHeader{GPT-6-Luna}{N} & \JEValHeader{Qwen3.5-4B}{T} & \JEValHeader{Qwen3.8-27B}{T (xhigh)} & \JEValHeader{DeepSeek\\V4-Flash}{T (high)} & \JEValHeader{GLM-5.3-Flash}{T (high)}\\\midrule
\endhead
\midrule\multicolumn{10}{r@{}}{\fontsize{7.3}{8.4}\selectfont Continued on next page}\\\endfoot
\bottomrule\endlastfoot
\multicolumn{10}{@{}l@{}}{\textbf{agentharm}\hfill $N=97$, $N_{\mathrm{score}}=103$}\\*
ACC $\uparrow$ & 61.86 & 74.23 & 78.35 & 76.29 & 71.13 & 70.10 & 84.54 & 77.32 & 85.57\\*
Latency $\downarrow$ & \underline{1.09} & 3.62 & 3.74 & 10.19 & 20.26 & 27.16 & 98.40 & 58.86 & 16.14\\*
ECE $\downarrow$ & 0.349 & 0.242 & 0.201 & 0.204 & 0.243 & 0.256 & 0.114 & 0.220 & 0.089\\*
$\JEValBrierRaw\,\downarrow$ & 0.688 & 0.484 & 0.409 & 0.417 & 0.485 & 0.501 & 0.261 & 0.437 & 0.246\\*
MAE $\downarrow$ & 0.900 & 0.960 & 0.780 & \underline{0.422} & 0.594 & 0.680 & 0.641 & 0.515 & \textbf{0.418}\\*
Valid score & 70/103 & 101/103 & 101/103 & 102/103 & 101/103 & 103/103 & 103/103 & 103/103 & 103/103\\[1.5pt]
\multicolumn{10}{@{}l@{}}{\textbf{decider}\hfill $N=200$}\\
\end{longtable}
\endgroup

\Needspace{100pt}
\subsection{Factuality and hallucination}
\label{app:domain-factuality-and-hallucination}
\begingroup\def\rmdefault{lmr}\normalfont\fontsize{8}{9.3}\selectfont\JEValSetTableWidths{9}\renewcommand{\arraystretch}{1.0}
\begin{longtable}{@{}>{\raggedright\arraybackslash}p{\JEValCompactCol}>{\raggedleft\arraybackslash}p{\JEValCompactCol}>{\raggedleft\arraybackslash}p{\JEValCompactCol}>{\raggedleft\arraybackslash}p{\JEValCompactCol}>{\raggedleft\arraybackslash}p{\JEValCompactCol}>{\raggedleft\arraybackslash}p{\JEValCompactCol}>{\raggedleft\arraybackslash}p{\JEValCompactCol}>{\raggedleft\arraybackslash}p{\JEValCompactCol}>{\raggedleft\arraybackslash}p{\JEValCompactCol}@{}}
\caption{Factuality and hallucination: model band 1/3. ACC (\%) $\uparrow$; benchmark-run latency (s), ECE, and $\JEValBrierRaw$ $\downarrow$. Rankings span all bands.}\label{tab:domain-factuality-and-hallucination}\\
\toprule
\multicolumn{9}{@{}l@{}}{\fontsize{7.3}{8.4}\selectfont\JEValModeKey}\\[1.5pt]
\JEValHeader{\textbf{Metric}}{} & \JEValHeader{Jev}{} & \JEValHeader{SII-Decision-4B}{(ours)} & \JEValHeader{SII-Decision-27B}{(ours)} & \JEValHeader{kev-0.8b}{} & \JEValHeader{kev-4b}{} & \JEValHeader{kev-9b}{} & \JEValHeader{kev-27b}{} & \JEValHeader{Intern-Decision\\0.8B}{}\\\midrule
\endfirsthead
\multicolumn{9}{@{}l}{\textit{Table \thetable{} continued; model band 1/3}}\\\toprule
\multicolumn{9}{@{}l@{}}{\fontsize{7.3}{8.4}\selectfont\JEValModeKey}\\[1.5pt]
\JEValHeader{\textbf{Metric}}{} & \JEValHeader{Jev}{} & \JEValHeader{SII-Decision-4B}{(ours)} & \JEValHeader{SII-Decision-27B}{(ours)} & \JEValHeader{kev-0.8b}{} & \JEValHeader{kev-4b}{} & \JEValHeader{kev-9b}{} & \JEValHeader{kev-27b}{} & \JEValHeader{Intern-Decision\\0.8B}{}\\\midrule
\endhead
\midrule\multicolumn{9}{r@{}}{\fontsize{7.3}{8.4}\selectfont Continued on next page}\\\endfoot
\bottomrule\endlastfoot
\multicolumn{9}{@{}l@{}}{\textbf{halueval-dialogue}\hfill $N=200$}\\*
ACC $\uparrow$ & 92.00 & 87.00 & 92.00 & 64.00 & 81.50 & 85.00 & 91.50 & 71.00\\*
Latency $\downarrow$ & \textbf{1.17} & 5.63 & 12.08 & 16.83 & 56.94 & 92.24 & 54.17 & 6.28\\*
ECE $\downarrow$ & 0.031 & 0.050 & 0.067 & 0.041 & 0.063 & 0.114 & 0.083 & 0.070\\*
$\JEValBrierRaw\,\downarrow$ & 0.093 & 0.169 & 0.136 & 0.422 & 0.274 & 0.227 & 0.124 & 0.411\\[1.5pt]
\multicolumn{9}{@{}l@{}}{\textbf{halueval-general}\hfill $N=200$}\\*
ACC $\uparrow$ & 71.00 & 80.00 & 77.50 & 78.50 & 77.00 & 79.50 & 80.00 & 73.50\\*
Latency $\downarrow$ & \textbf{1.18} & 5.62 & 12.00 & 17.02 & 57.01 & 92.21 & 53.91 & 6.33\\
\end{longtable}
\endgroup

\begingroup\def\rmdefault{lmr}\normalfont\fontsize{8}{9.3}\selectfont\JEValSetTableWidths{9}\renewcommand{\arraystretch}{1.0}
\begin{longtable}{@{}>{\raggedright\arraybackslash}p{\JEValCompactCol}>{\raggedleft\arraybackslash}p{\JEValCompactCol}>{\raggedleft\arraybackslash}p{\JEValCompactCol}>{\raggedleft\arraybackslash}p{\JEValCompactCol}>{\raggedleft\arraybackslash}p{\JEValCompactCol}>{\raggedleft\arraybackslash}p{\JEValCompactCol}>{\raggedleft\arraybackslash}p{\JEValCompactCol}>{\raggedleft\arraybackslash}p{\JEValCompactCol}>{\raggedleft\arraybackslash}p{\JEValCompactCol}@{}}
\caption{Factuality and hallucination: model band 2/3. ACC (\%) $\uparrow$; benchmark-run latency (s), ECE, and $\JEValBrierRaw$ $\downarrow$. Rankings span all bands.}\label{tab:domain-factuality-and-hallucination-band-2}\\
\toprule
\multicolumn{9}{@{}l@{}}{\fontsize{7.3}{8.4}\selectfont\JEValModeKey}\\[1.5pt]
\JEValHeader{\textbf{Metric}}{} & \JEValHeader{Intern-Decision\\2B}{} & \JEValHeader{Intern-Decision\\4B}{} & \JEValHeader{laya}{} & \JEValHeader{openjev-4b-v5}{} & \JEValHeader{nimble-9b}{} & \JEValHeader{nanojev}{} & \JEValHeader{lightjev}{} & \JEValHeader{jevforge-0.8b}{}\\\midrule
\endfirsthead
\multicolumn{9}{@{}l}{\textit{Table \thetable{} continued; model band 2/3}}\\\toprule
\multicolumn{9}{@{}l@{}}{\fontsize{7.3}{8.4}\selectfont\JEValModeKey}\\[1.5pt]
\JEValHeader{\textbf{Metric}}{} & \JEValHeader{Intern-Decision\\2B}{} & \JEValHeader{Intern-Decision\\4B}{} & \JEValHeader{laya}{} & \JEValHeader{openjev-4b-v5}{} & \JEValHeader{nimble-9b}{} & \JEValHeader{nanojev}{} & \JEValHeader{lightjev}{} & \JEValHeader{jevforge-0.8b}{}\\\midrule
\endhead
\midrule\multicolumn{9}{r@{}}{\fontsize{7.3}{8.4}\selectfont Continued on next page}\\\endfoot
\bottomrule\endlastfoot
\multicolumn{9}{@{}l@{}}{\textbf{halueval-dialogue}\hfill $N=200$}\\*
ACC $\uparrow$ & 70.00 & 87.00 & 59.00 & 84.50 & 92.00 & 47.50 & 84.00 & 38.00\\*
Latency $\downarrow$ & 5.74 & 9.79 & 12.83 & 237.61 & 23.58 & 36.33 & 23.26 & 223.89\\*
ECE $\downarrow$ & 0.052 & 0.043 & 0.031 & 0.034 & 0.059 & 0.151 & 0.236 & 0.246\\*
$\JEValBrierRaw\,\downarrow$ & 0.365 & 0.186 & 0.432 & 0.216 & 0.149 & 0.537 & 0.359 & 0.652\\[1.5pt]
\multicolumn{9}{@{}l@{}}{\textbf{halueval-general}\hfill $N=200$}\\*
ACC $\uparrow$ & 74.50 & \textbf{81.00} & 40.00 & 79.50 & \underline{80.50} & 20.00 & 47.50 & 80.00\\*
Latency $\downarrow$ & 5.76 & 9.85 & 11.85 & 238.89 & 23.16 & 36.92 & 45.94 & 221.26\\
\end{longtable}%
\endgroup

\Needspace{100pt}
\subsection{Long-context and document understanding}
\label{app:domain-long-context-and-document-understanding}
\begingroup\def\rmdefault{lmr}\normalfont\fontsize{8}{9.3}\selectfont\JEValSetTableWidths{9}\renewcommand{\arraystretch}{1.0}
%
\endgroup

\Needspace{100pt}
\subsection{Social science}
\label{app:domain-social-science}
\begingroup\def\rmdefault{lmr}\normalfont\fontsize{8}{9.3}\selectfont\JEValSetTableWidths{9}\renewcommand{\arraystretch}{1.0}
%
\endgroup

\Needspace{100pt}
\subsection{Personalization and preferences}
\label{app:domain-personalization-and-preferences}
\begingroup\def\rmdefault{lmr}\normalfont\fontsize{8}{9.3}\selectfont\JEValSetTableWidths{9}\renewcommand{\arraystretch}{1.0}
%
\endgroup

\FloatBarrier
\endgroup

\section{More Analysis Details}
\subsection{Mathematical and Probability Task Prompts}
\label{app:math-prompts}
\begin{figure}[htbp]
\centering
\includegraphics[width=0.5\columnwidth]{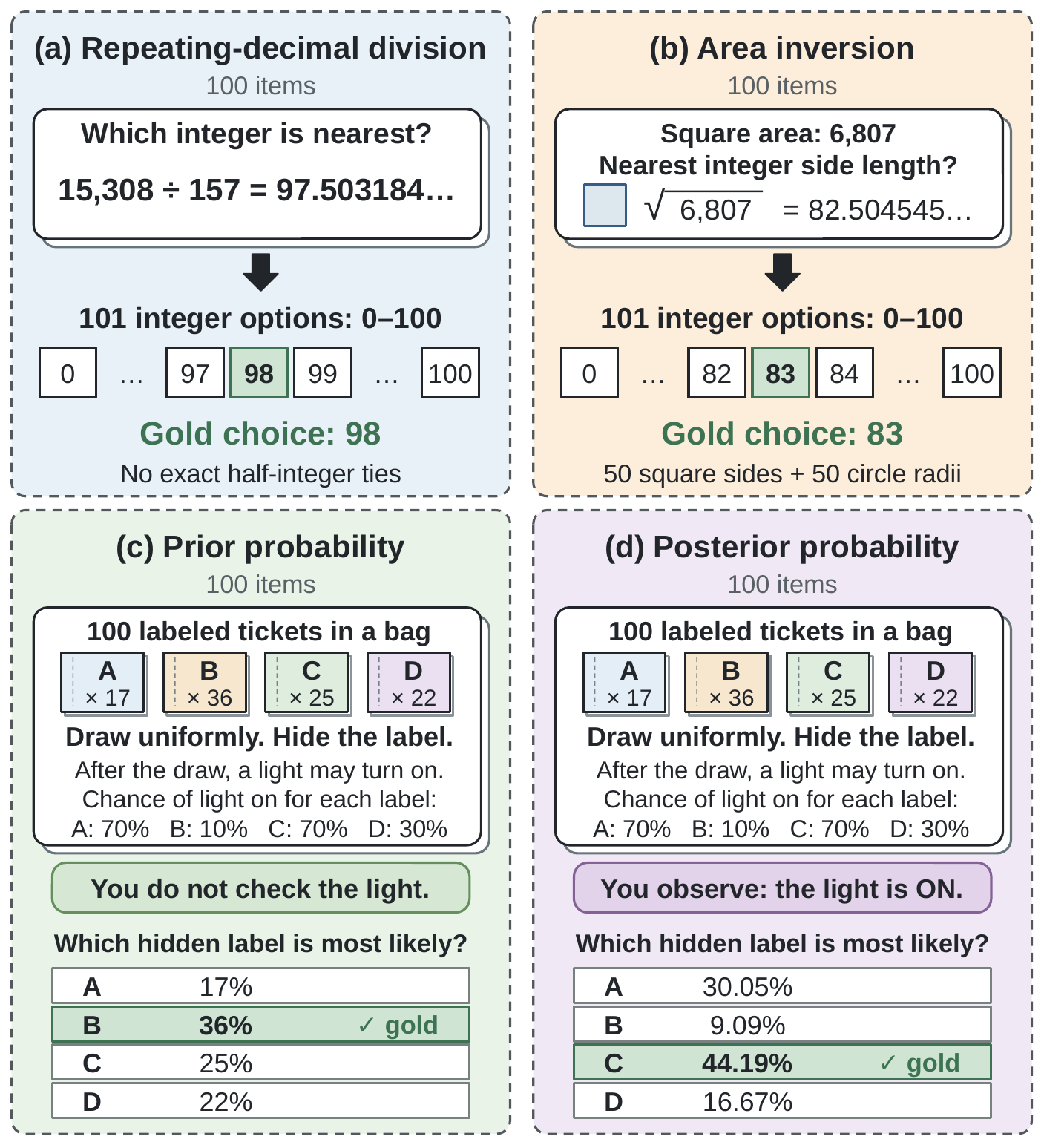}
\caption{Examples of the four tasks, with gold choices in green and true probabilities shown. Numerical options are abbreviated.}
\label{fig:math-probability-benchmark}
\end{figure}

Each row presents a complete task template for JEV (left) and LLMs (right). Braced placeholders are filled per item. Numerical options expand to all 101 integers; each probability map contains every candidate key. Field headings are presentation aids. JEV receives structured fields; LLMs receive the displayed system and user messages with a strict response JSON Schema. Area variants are alternatives within one task; posterior observations likewise select one of the two stated outcomes. Figure~\ref{fig:math-probability-benchmark} shows one example of each task: the numerical examples identify the nearest-integer gold choice, while the prior and posterior examples show the gold choice together with the analytically determined probabilities of all candidate labels.\par\medskip

\begin{promptbox}{1. Repeating-decimal division --- JEV}
\fontsize{9}{10.5}\selectfont\raggedright
\textbf{Request fields}\par\smallskip Question ID: q; type: choice.\par\medskip

\textbf{State:}\par\smallskip {\ttfamily \{\allowbreak{}"numerator":\allowbreak{} \{\allowbreak{}numerator\}\allowbreak{},\allowbreak{} "denominator":\allowbreak{} \{\allowbreak{}denominator\}\allowbreak{}\}\allowbreak{}}\par\medskip

\textbf{Task instruction:}\par\smallskip Divide numerator by denominator. Select the integer from 0 through 100 nearest to the exact quotient. No case is exactly halfway between two integers.\par\medskip

\textbf{Criteria:}\par\smallskip {\ttfamily n\_\allowbreak{}0\allowbreak{}:\allowbreak{} 0\allowbreak{}\par n\_\allowbreak{}1\allowbreak{}:\allowbreak{} 1\allowbreak{}\par ...\par n\_\allowbreak{}1\allowbreak{}0\allowbreak{}0\allowbreak{}:\allowbreak{} 1\allowbreak{}0\allowbreak{}0\allowbreak{}}\par\medskip

\textbf{Response structure:}\par\smallskip {\ttfamily \{\allowbreak{}"answers":\allowbreak{}\{\allowbreak{}"q":\allowbreak{}\{\allowbreak{}"type":\allowbreak{}"choice",\allowbreak{} "choice":\allowbreak{}"\{\allowbreak{}selected\_\allowbreak{}key\}\allowbreak{}",\allowbreak{} "confidence":\allowbreak{}\{\allowbreak{}confidence\}\allowbreak{},\allowbreak{} "probabilities":\allowbreak{}\{\allowbreak{}probability\_\allowbreak{}for\_\allowbreak{}every\_\allowbreak{}key\}\allowbreak{}\}\allowbreak{}\}\allowbreak{}\}\allowbreak{}}\par\medskip

\end{promptbox}

\begin{promptbox}{1. Repeating-decimal division --- LLM}
\fontsize{9}{10.5}\selectfont\raggedright
\textbf{System message}\par\smallskip Return only one JSON object conforming to the response JSON Schema.\par Use the original question ID under "answers".\par Include every field required for this question type.\par For choice and score questions, include every option or level exactly once\par in "probabilities", including entries with value 0.\par Output confidence and probability values as JSON numbers between 0 and 1.\par For noul questions, "noul" is the probability that the answer is true.\par Do not output Markdown, comments, explanations, or additional fields.\par\medskip

\textbf{User message}\par\smallskip
\textbf{State:}\par\smallskip {\ttfamily \{\allowbreak{}"numerator":\allowbreak{} \{\allowbreak{}numerator\}\allowbreak{},\allowbreak{} "denominator":\allowbreak{} \{\allowbreak{}denominator\}\allowbreak{}\}\allowbreak{}}\par\medskip

\textbf{Task instruction:}\par\smallskip Divide numerator by denominator. Select the integer from 0 through 100 nearest to the exact quotient. No case is exactly halfway between two integers.\par\medskip

\textbf{Options:}\par\smallskip {\ttfamily - n\_\allowbreak{}0\allowbreak{}:\allowbreak{} 0\allowbreak{}\par - n\_\allowbreak{}1\allowbreak{}:\allowbreak{} 1\allowbreak{}\par ...\par - n\_\allowbreak{}1\allowbreak{}0\allowbreak{}0\allowbreak{}:\allowbreak{} 1\allowbreak{}0\allowbreak{}0\allowbreak{}}\par\medskip

\textbf{Response structure:}\par\smallskip {\ttfamily \{\allowbreak{}"answers":\allowbreak{}\{\allowbreak{}"q":\allowbreak{}\{\allowbreak{}"type":\allowbreak{}"choice",\allowbreak{} "choice":\allowbreak{}"\{\allowbreak{}selected\_\allowbreak{}key\}\allowbreak{}",\allowbreak{} "confidence":\allowbreak{}\{\allowbreak{}confidence\}\allowbreak{},\allowbreak{} "probabilities":\allowbreak{}\{\allowbreak{}probability\_\allowbreak{}for\_\allowbreak{}every\_\allowbreak{}key\}\allowbreak{}\}\allowbreak{}\}\allowbreak{}\}\allowbreak{}}\par\medskip

\end{promptbox}

\begin{promptbox}{2. Area inversion --- JEV}
\fontsize{9}{10.5}\selectfont\raggedright
\textbf{Request fields}\par\smallskip Question ID: q; type: choice.\par\medskip

\textbf{State:}\par\smallskip {\ttfamily Square:\allowbreak{} \{\allowbreak{}"shape":\allowbreak{}"square",\allowbreak{} "area\_\allowbreak{}square\_\allowbreak{}units":\allowbreak{}\{\allowbreak{}area\}\allowbreak{}\}\allowbreak{}\par Circle:\allowbreak{} \{\allowbreak{}"shape":\allowbreak{}"circle",\allowbreak{} "area\_\allowbreak{}square\_\allowbreak{}units":\allowbreak{}\{\allowbreak{}area\}\allowbreak{},\allowbreak{} "pi":\allowbreak{}"3\allowbreak{}.1\allowbreak{}4\allowbreak{}1\allowbreak{}5\allowbreak{}9\allowbreak{}2\allowbreak{}6\allowbreak{}5\allowbreak{}3\allowbreak{}5\allowbreak{}8\allowbreak{}9\allowbreak{}7\allowbreak{}9\allowbreak{}3\allowbreak{}2\allowbreak{}3\allowbreak{}8\allowbreak{}4\allowbreak{}6\allowbreak{}2\allowbreak{}6\allowbreak{}4\allowbreak{}3\allowbreak{}3\allowbreak{}8\allowbreak{}3\allowbreak{}2\allowbreak{}7\allowbreak{}9\allowbreak{}5\allowbreak{}0\allowbreak{}2\allowbreak{}8\allowbreak{}8\allowbreak{}4\allowbreak{}1\allowbreak{}9\allowbreak{}7\allowbreak{}1\allowbreak{}6\allowbreak{}9\allowbreak{}3\allowbreak{}9\allowbreak{}9\allowbreak{}3\allowbreak{}7\allowbreak{}5\allowbreak{}1\allowbreak{}"\}\allowbreak{}}\par\medskip

\textbf{Task instruction:}\par\smallskip [Square]\par A square has the area supplied in state. Its area equals side length squared. Select the integer from 0 through 100 nearest to its side length. No case is exactly halfway between two integers.\par [Circle]\par A circle has the area supplied in state. Its area equals pi times radius squared. Use the decimal pi supplied in state as the exact constant for this problem. Select the integer from 0 through 100 nearest to its radius. No case is exactly halfway between two integers.\par\medskip

\textbf{Criteria:}\par\smallskip {\ttfamily n\_\allowbreak{}0\allowbreak{}:\allowbreak{} 0\allowbreak{}\par n\_\allowbreak{}1\allowbreak{}:\allowbreak{} 1\allowbreak{}\par ...\par n\_\allowbreak{}1\allowbreak{}0\allowbreak{}0\allowbreak{}:\allowbreak{} 1\allowbreak{}0\allowbreak{}0\allowbreak{}}\par\medskip

\textbf{Response structure:}\par\smallskip {\ttfamily \{\allowbreak{}"answers":\allowbreak{}\{\allowbreak{}"q":\allowbreak{}\{\allowbreak{}"type":\allowbreak{}"choice",\allowbreak{} "choice":\allowbreak{}"\{\allowbreak{}selected\_\allowbreak{}key\}\allowbreak{}",\allowbreak{} "confidence":\allowbreak{}\{\allowbreak{}confidence\}\allowbreak{},\allowbreak{} "probabilities":\allowbreak{}\{\allowbreak{}probability\_\allowbreak{}for\_\allowbreak{}every\_\allowbreak{}key\}\allowbreak{}\}\allowbreak{}\}\allowbreak{}\}\allowbreak{}}\par\medskip

\end{promptbox}

\begin{promptbox}{2. Area inversion --- LLM}
\fontsize{9}{10.5}\selectfont\raggedright
\textbf{System message}\par\smallskip Return only one JSON object conforming to the response JSON Schema.\par Use the original question ID under "answers".\par Include every field required for this question type.\par For choice and score questions, include every option or level exactly once\par in "probabilities", including entries with value 0.\par Output confidence and probability values as JSON numbers between 0 and 1.\par For noul questions, "noul" is the probability that the answer is true.\par Do not output Markdown, comments, explanations, or additional fields.\par\medskip

\textbf{User message}\par\smallskip
\textbf{State:}\par\smallskip {\ttfamily Square:\allowbreak{} \{\allowbreak{}"shape":\allowbreak{}"square",\allowbreak{} "area\_\allowbreak{}square\_\allowbreak{}units":\allowbreak{}\{\allowbreak{}area\}\allowbreak{}\}\allowbreak{}\par Circle:\allowbreak{} \{\allowbreak{}"shape":\allowbreak{}"circle",\allowbreak{} "area\_\allowbreak{}square\_\allowbreak{}units":\allowbreak{}\{\allowbreak{}area\}\allowbreak{},\allowbreak{} "pi":\allowbreak{}"3\allowbreak{}.1\allowbreak{}4\allowbreak{}1\allowbreak{}5\allowbreak{}9\allowbreak{}2\allowbreak{}6\allowbreak{}5\allowbreak{}3\allowbreak{}5\allowbreak{}8\allowbreak{}9\allowbreak{}7\allowbreak{}9\allowbreak{}3\allowbreak{}2\allowbreak{}3\allowbreak{}8\allowbreak{}4\allowbreak{}6\allowbreak{}2\allowbreak{}6\allowbreak{}4\allowbreak{}3\allowbreak{}3\allowbreak{}8\allowbreak{}3\allowbreak{}2\allowbreak{}7\allowbreak{}9\allowbreak{}5\allowbreak{}0\allowbreak{}2\allowbreak{}8\allowbreak{}8\allowbreak{}4\allowbreak{}1\allowbreak{}9\allowbreak{}7\allowbreak{}1\allowbreak{}6\allowbreak{}9\allowbreak{}3\allowbreak{}9\allowbreak{}9\allowbreak{}3\allowbreak{}7\allowbreak{}5\allowbreak{}1\allowbreak{}"\}\allowbreak{}}\par\medskip

\textbf{Task instruction:}\par\smallskip [Square]\par A square has the area supplied in state. Its area equals side length squared. Select the integer from 0 through 100 nearest to its side length. No case is exactly halfway between two integers.\par [Circle]\par A circle has the area supplied in state. Its area equals pi times radius squared. Use the decimal pi supplied in state as the exact constant for this problem. Select the integer from 0 through 100 nearest to its radius. No case is exactly halfway between two integers.\par\medskip

\textbf{Options:}\par\smallskip {\ttfamily - n\_\allowbreak{}0\allowbreak{}:\allowbreak{} 0\allowbreak{}\par - n\_\allowbreak{}1\allowbreak{}:\allowbreak{} 1\allowbreak{}\par ...\par - n\_\allowbreak{}1\allowbreak{}0\allowbreak{}0\allowbreak{}:\allowbreak{} 1\allowbreak{}0\allowbreak{}0\allowbreak{}}\par\medskip

\textbf{Response structure:}\par\smallskip {\ttfamily \{\allowbreak{}"answers":\allowbreak{}\{\allowbreak{}"q":\allowbreak{}\{\allowbreak{}"type":\allowbreak{}"choice",\allowbreak{} "choice":\allowbreak{}"\{\allowbreak{}selected\_\allowbreak{}key\}\allowbreak{}",\allowbreak{} "confidence":\allowbreak{}\{\allowbreak{}confidence\}\allowbreak{},\allowbreak{} "probabilities":\allowbreak{}\{\allowbreak{}probability\_\allowbreak{}for\_\allowbreak{}every\_\allowbreak{}key\}\allowbreak{}\}\allowbreak{}\}\allowbreak{}\}\allowbreak{}}\par\medskip

\end{promptbox}

\begin{promptbox}{3. Prior probability --- JEV}
\fontsize{9}{10.5}\selectfont\raggedright
\textbf{Request fields}\par\smallskip Question ID: q; type: choice.\par\medskip

\textbf{State:}\par\smallskip {\ttfamily \{\allowbreak{}"sampling":\allowbreak{}"A ticket is drawn uniformly at random from the bag. Its label is hidden.",\allowbreak{} "ticket\_\allowbreak{}counts":\allowbreak{}\{\allowbreak{}"A":\allowbreak{}\{\allowbreak{}n\_\allowbreak{}A\}\allowbreak{},\allowbreak{} "B":\allowbreak{}\{\allowbreak{}n\_\allowbreak{}B\}\allowbreak{},\allowbreak{} "C":\allowbreak{}\{\allowbreak{}n\_\allowbreak{}C\}\allowbreak{},\allowbreak{} "D":\allowbreak{}\{\allowbreak{}n\_\allowbreak{}D\}\allowbreak{}\}\allowbreak{},\allowbreak{} "clue\_\allowbreak{}likelihood":\allowbreak{}\{\allowbreak{}"P(X|\allowbreak{}label=\allowbreak{}A)":\allowbreak{}"\{\allowbreak{}likelihood\_\allowbreak{}A\}\allowbreak{}",\allowbreak{} "P(X|\allowbreak{}label=\allowbreak{}B)":\allowbreak{}"\{\allowbreak{}likelihood\_\allowbreak{}B\}\allowbreak{}",\allowbreak{} "P(X|\allowbreak{}label=\allowbreak{}C)":\allowbreak{}"\{\allowbreak{}likelihood\_\allowbreak{}C\}\allowbreak{}",\allowbreak{} "P(X|\allowbreak{}label=\allowbreak{}D)":\allowbreak{}"\{\allowbreak{}likelihood\_\allowbreak{}D\}\allowbreak{}"\}\allowbreak{},\allowbreak{} "observation":\allowbreak{}"\{\allowbreak{}observation\}\allowbreak{}"\}\allowbreak{}}\par\medskip

\textbf{Observation value:}\par\smallskip The clue was not checked. No result was observed; whether X occurred is unknown. The ticket label was not revealed.\par\medskip

\textbf{Task instruction:}\par\smallskip Which label is on the drawn ticket?\par For each option, report the probability that it is the drawn ticket's label, given the available information.\par\medskip

\textbf{Criteria:}\par\smallskip {\ttfamily A:\allowbreak{} The hidden ticket has label A.\par B:\allowbreak{} The hidden ticket has label B.\par C:\allowbreak{} The hidden ticket has label C.\par D:\allowbreak{} The hidden ticket has label D.}\par\medskip

\textbf{Response structure:}\par\smallskip {\ttfamily \{\allowbreak{}"answers":\allowbreak{}\{\allowbreak{}"q":\allowbreak{}\{\allowbreak{}"type":\allowbreak{}"choice",\allowbreak{} "choice":\allowbreak{}"\{\allowbreak{}selected\_\allowbreak{}key\}\allowbreak{}",\allowbreak{} "confidence":\allowbreak{}\{\allowbreak{}confidence\}\allowbreak{},\allowbreak{} "probabilities":\allowbreak{}\{\allowbreak{}probability\_\allowbreak{}for\_\allowbreak{}every\_\allowbreak{}key\}\allowbreak{}\}\allowbreak{}\}\allowbreak{}\}\allowbreak{}}\par\medskip

\end{promptbox}

\begin{promptbox}{3. Prior probability --- LLM}
\fontsize{9}{10.5}\selectfont\raggedright
\textbf{System message}\par\smallskip Return only one JSON object conforming to the response JSON Schema.\par Use the original question ID under "answers".\par Include every field required for this question type.\par For choice and score questions, include every option or level exactly once\par in "probabilities", including entries with value 0.\par Output confidence and probability values as JSON numbers between 0 and 1.\par For noul questions, "noul" is the probability that the answer is true.\par Do not output Markdown, comments, explanations, or additional fields.\par\medskip

\textbf{User message}\par\smallskip
\textbf{State:}\par\smallskip {\ttfamily \{\allowbreak{}"sampling":\allowbreak{}"A ticket is drawn uniformly at random from the bag. Its label is hidden.",\allowbreak{} "ticket\_\allowbreak{}counts":\allowbreak{}\{\allowbreak{}"A":\allowbreak{}\{\allowbreak{}n\_\allowbreak{}A\}\allowbreak{},\allowbreak{} "B":\allowbreak{}\{\allowbreak{}n\_\allowbreak{}B\}\allowbreak{},\allowbreak{} "C":\allowbreak{}\{\allowbreak{}n\_\allowbreak{}C\}\allowbreak{},\allowbreak{} "D":\allowbreak{}\{\allowbreak{}n\_\allowbreak{}D\}\allowbreak{}\}\allowbreak{},\allowbreak{} "clue\_\allowbreak{}likelihood":\allowbreak{}\{\allowbreak{}"P(X|\allowbreak{}label=\allowbreak{}A)":\allowbreak{}"\{\allowbreak{}likelihood\_\allowbreak{}A\}\allowbreak{}",\allowbreak{} "P(X|\allowbreak{}label=\allowbreak{}B)":\allowbreak{}"\{\allowbreak{}likelihood\_\allowbreak{}B\}\allowbreak{}",\allowbreak{} "P(X|\allowbreak{}label=\allowbreak{}C)":\allowbreak{}"\{\allowbreak{}likelihood\_\allowbreak{}C\}\allowbreak{}",\allowbreak{} "P(X|\allowbreak{}label=\allowbreak{}D)":\allowbreak{}"\{\allowbreak{}likelihood\_\allowbreak{}D\}\allowbreak{}"\}\allowbreak{},\allowbreak{} "observation":\allowbreak{}"\{\allowbreak{}observation\}\allowbreak{}"\}\allowbreak{}}\par\medskip

\textbf{Observation value:}\par\smallskip The clue was not checked. No result was observed; whether X occurred is unknown. The ticket label was not revealed.\par\medskip

\textbf{Task instruction:}\par\smallskip A ticket label was randomly selected and is hidden. Given all information in state, estimate the conditional probability that the hidden ticket has each label A, B, C, or D. The four events are mutually exclusive and exhaustive. For each label k, probabilities[k] must mean P(hidden label = k | information provided in state), not the probability that k is the most likely label. Select the label with the largest estimated probability.\par\medskip

\textbf{Options:}\par\smallskip {\ttfamily - A:\allowbreak{} The hidden ticket has label A.\par - B:\allowbreak{} The hidden ticket has label B.\par - C:\allowbreak{} The hidden ticket has label C.\par - D:\allowbreak{} The hidden ticket has label D.}\par\medskip

\textbf{Response structure:}\par\smallskip {\ttfamily \{\allowbreak{}"answers":\allowbreak{}\{\allowbreak{}"q":\allowbreak{}\{\allowbreak{}"type":\allowbreak{}"choice",\allowbreak{} "choice":\allowbreak{}"\{\allowbreak{}selected\_\allowbreak{}key\}\allowbreak{}",\allowbreak{} "confidence":\allowbreak{}\{\allowbreak{}confidence\}\allowbreak{},\allowbreak{} "probabilities":\allowbreak{}\{\allowbreak{}probability\_\allowbreak{}for\_\allowbreak{}every\_\allowbreak{}key\}\allowbreak{}\}\allowbreak{}\}\allowbreak{}\}\allowbreak{}}\par\medskip

\end{promptbox}

\begin{promptbox}{4. Posterior probability --- JEV}
\fontsize{9}{10.5}\selectfont\raggedright
\textbf{Request fields}\par\smallskip Question ID: q; type: choice.\par\medskip

\textbf{State:}\par\smallskip {\ttfamily \{\allowbreak{}"sampling":\allowbreak{}"A ticket is drawn uniformly at random from the bag. Its label is hidden.",\allowbreak{} "ticket\_\allowbreak{}counts":\allowbreak{}\{\allowbreak{}"A":\allowbreak{}\{\allowbreak{}n\_\allowbreak{}A\}\allowbreak{},\allowbreak{} "B":\allowbreak{}\{\allowbreak{}n\_\allowbreak{}B\}\allowbreak{},\allowbreak{} "C":\allowbreak{}\{\allowbreak{}n\_\allowbreak{}C\}\allowbreak{},\allowbreak{} "D":\allowbreak{}\{\allowbreak{}n\_\allowbreak{}D\}\allowbreak{}\}\allowbreak{},\allowbreak{} "clue\_\allowbreak{}likelihood":\allowbreak{}\{\allowbreak{}"P(X|\allowbreak{}label=\allowbreak{}A)":\allowbreak{}"\{\allowbreak{}likelihood\_\allowbreak{}A\}\allowbreak{}",\allowbreak{} "P(X|\allowbreak{}label=\allowbreak{}B)":\allowbreak{}"\{\allowbreak{}likelihood\_\allowbreak{}B\}\allowbreak{}",\allowbreak{} "P(X|\allowbreak{}label=\allowbreak{}C)":\allowbreak{}"\{\allowbreak{}likelihood\_\allowbreak{}C\}\allowbreak{}",\allowbreak{} "P(X|\allowbreak{}label=\allowbreak{}D)":\allowbreak{}"\{\allowbreak{}likelihood\_\allowbreak{}D\}\allowbreak{}"\}\allowbreak{},\allowbreak{} "observation":\allowbreak{}"\{\allowbreak{}observation\}\allowbreak{}"\}\allowbreak{}}\par\medskip

\textbf{Observation value (choose one):}\par\smallskip The clue was checked. The observed result was X (X=true); the ticket label was not revealed.\par The clue was checked. The observed result was not X (X=false); the ticket label was not revealed.\par\medskip

\textbf{Task instruction:}\par\smallskip Which label is on the drawn ticket?\par For each option, report the probability that it is the drawn ticket's label, given the available information.\par\medskip

\textbf{Criteria:}\par\smallskip {\ttfamily A:\allowbreak{} The hidden ticket has label A.\par B:\allowbreak{} The hidden ticket has label B.\par C:\allowbreak{} The hidden ticket has label C.\par D:\allowbreak{} The hidden ticket has label D.}\par\medskip

\textbf{Response structure:}\par\smallskip {\ttfamily \{\allowbreak{}"answers":\allowbreak{}\{\allowbreak{}"q":\allowbreak{}\{\allowbreak{}"type":\allowbreak{}"choice",\allowbreak{} "choice":\allowbreak{}"\{\allowbreak{}selected\_\allowbreak{}key\}\allowbreak{}",\allowbreak{} "confidence":\allowbreak{}\{\allowbreak{}confidence\}\allowbreak{},\allowbreak{} "probabilities":\allowbreak{}\{\allowbreak{}probability\_\allowbreak{}for\_\allowbreak{}every\_\allowbreak{}key\}\allowbreak{}\}\allowbreak{}\}\allowbreak{}\}\allowbreak{}}\par\medskip

\end{promptbox}

\begin{promptbox}{4. Posterior probability --- LLM}
\fontsize{9}{10.5}\selectfont\raggedright
\textbf{System message}\par\smallskip Return only one JSON object conforming to the response JSON Schema.\par Use the original question ID under "answers".\par Include every field required for this question type.\par For choice and score questions, include every option or level exactly once\par in "probabilities", including entries with value 0.\par Output confidence and probability values as JSON numbers between 0 and 1.\par For noul questions, "noul" is the probability that the answer is true.\par Do not output Markdown, comments, explanations, or additional fields.\par\medskip

\textbf{User message}\par\smallskip
\textbf{State:}\par\smallskip {\ttfamily \{\allowbreak{}"sampling":\allowbreak{}"A ticket is drawn uniformly at random from the bag. Its label is hidden.",\allowbreak{} "ticket\_\allowbreak{}counts":\allowbreak{}\{\allowbreak{}"A":\allowbreak{}\{\allowbreak{}n\_\allowbreak{}A\}\allowbreak{},\allowbreak{} "B":\allowbreak{}\{\allowbreak{}n\_\allowbreak{}B\}\allowbreak{},\allowbreak{} "C":\allowbreak{}\{\allowbreak{}n\_\allowbreak{}C\}\allowbreak{},\allowbreak{} "D":\allowbreak{}\{\allowbreak{}n\_\allowbreak{}D\}\allowbreak{}\}\allowbreak{},\allowbreak{} "clue\_\allowbreak{}likelihood":\allowbreak{}\{\allowbreak{}"P(X|\allowbreak{}label=\allowbreak{}A)":\allowbreak{}"\{\allowbreak{}likelihood\_\allowbreak{}A\}\allowbreak{}",\allowbreak{} "P(X|\allowbreak{}label=\allowbreak{}B)":\allowbreak{}"\{\allowbreak{}likelihood\_\allowbreak{}B\}\allowbreak{}",\allowbreak{} "P(X|\allowbreak{}label=\allowbreak{}C)":\allowbreak{}"\{\allowbreak{}likelihood\_\allowbreak{}C\}\allowbreak{}",\allowbreak{} "P(X|\allowbreak{}label=\allowbreak{}D)":\allowbreak{}"\{\allowbreak{}likelihood\_\allowbreak{}D\}\allowbreak{}"\}\allowbreak{},\allowbreak{} "observation":\allowbreak{}"\{\allowbreak{}observation\}\allowbreak{}"\}\allowbreak{}}\par\medskip

\textbf{Observation value (choose one):}\par\smallskip The clue was checked. The observed result was X (X=true); the ticket label was not revealed.\par The clue was checked. The observed result was not X (X=false); the ticket label was not revealed.\par\medskip

\textbf{Task instruction:}\par\smallskip A ticket label was randomly selected and is hidden. Given all information in state, estimate the conditional probability that the hidden ticket has each label A, B, C, or D. The four events are mutually exclusive and exhaustive. For each label k, probabilities[k] must mean P(hidden label = k | information provided in state), not the probability that k is the most likely label. Select the label with the largest estimated probability.\par\medskip

\textbf{Options:}\par\smallskip {\ttfamily - A:\allowbreak{} The hidden ticket has label A.\par - B:\allowbreak{} The hidden ticket has label B.\par - C:\allowbreak{} The hidden ticket has label C.\par - D:\allowbreak{} The hidden ticket has label D.}\par\medskip

\textbf{Response structure:}\par\smallskip {\ttfamily \{\allowbreak{}"answers":\allowbreak{}\{\allowbreak{}"q":\allowbreak{}\{\allowbreak{}"type":\allowbreak{}"choice",\allowbreak{} "choice":\allowbreak{}"\{\allowbreak{}selected\_\allowbreak{}key\}\allowbreak{}",\allowbreak{} "confidence":\allowbreak{}\{\allowbreak{}confidence\}\allowbreak{},\allowbreak{} "probabilities":\allowbreak{}\{\allowbreak{}probability\_\allowbreak{}for\_\allowbreak{}every\_\allowbreak{}key\}\allowbreak{}\}\allowbreak{}\}\allowbreak{}\}\allowbreak{}}\par\medskip

\end{promptbox}

\subsection{Agentic Workflow Evaluation}
\label{app:tau-bench-agentic}
We evaluate the original $\tau$-bench tool-calling agent and its JEV-based variant on the full Retail (115 tasks) and Airline (50 tasks) test sets. Each task is run once under both systems, yielding 330 interaction episodes in total. Both settings use the same version of $\tau$-bench and DeepSeek-V4-Flash as both the base agent model and the user simulator, with temperature set to 0. Task configurations, available tools, user simulation, environment execution, and reward verification are kept identical across the two systems. Each episode is allowed up to 30 interaction turns.

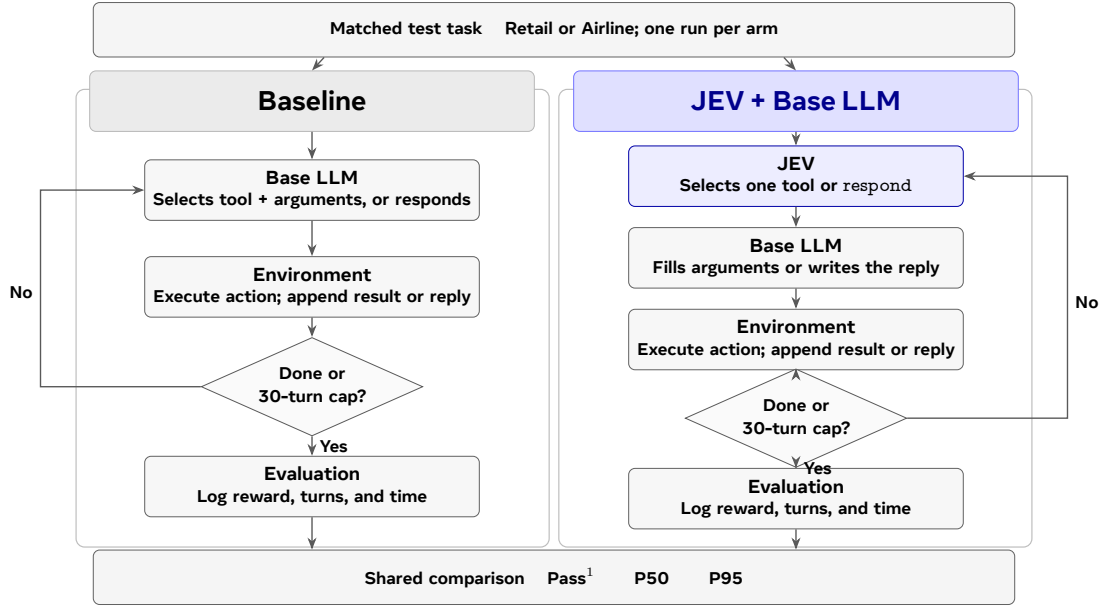
\begin{figure*}[htbp]
\centering
\scalebox{0.8}{
\begin{tikzpicture}[
    flow/.style={-{Stealth[length=2mm]},line width=0.65pt,draw=black!65},
    card/.style={draw=black!62,fill=black!3,rounded corners=3pt,
        align=center,text width=5.25cm,minimum height=10mm,inner sep=4pt,
        font=\small\sffamily},
    jevcard/.style={card,draw=blue!65!black,fill=blue!7},
    decision/.style={draw=black!65,fill=black!2,diamond,aspect=2.25,
        align=center,font=\small\bfseries\sffamily,text width=2.0cm,inner sep=1pt},
    lane/.style={draw=black!26,fill=white,rounded corners=4pt,line width=0.55pt},
    lanehead/.style={draw=black!28,fill=black!8,rounded corners=3pt,
        font=\Large\bfseries\sffamily,minimum width=7.35cm,minimum height=10mm},
    jevhead/.style={lanehead,draw=blue!55,fill=blue!12,text=blue!55!black},
    shared/.style={draw=black!62,fill=black!4,rounded corners=3pt,
        align=center,font=\small\sffamily,text width=14.95cm,
        minimum height=9mm,inner sep=4pt}
]
\node[shared] (setup) at (0,0)
    {\textbf{Matched test task} \quad Retail or Airline; one run per arm};
\draw[lane] (-7.9,-0.985) rectangle (-0.08,-8.55);
\draw[lane] (0.08,-0.985) rectangle (7.9,-8.55);
\node[lanehead] (bhead) at (-4,-1.18) {Baseline};
\node[jevhead] (jhead) at (4,-1.18) {JEV + Base LLM};
\draw[flow] ([xshift=-3.8cm]setup.south) -- (bhead.north);
\draw[flow] ([xshift=3.8cm]setup.south) -- (jhead.north);

\node[card] (bselect) at (-4,-2.65)
    {\textbf{\normalsize Base LLM}\\[-1pt]\small Selects tool + arguments, or responds};
\node[card] (benv) at (-4,-4.25)
    {\textbf{\normalsize Environment}\\[-1pt]\small Execute action; append result or reply};
\node[decision] (bcheck) at (-4,-5.9) {Done or\\30-turn cap?};
\node[card] (beval) at (-4,-7.55)
    {\textbf{\normalsize Evaluation}\\[-1pt]\small Log reward, turns, and time};
\draw[flow] (bhead.south) -- (bselect.north);
\draw[flow] (bselect) -- (benv);
\draw[flow] (benv) -- (bcheck);
\draw[flow] (bcheck) -- node[right,font=\small\sffamily] {Yes} (beval);
\draw[flow] (bcheck.west) -- ++(-2.65,0) |-
    node[pos=0.24,left,font=\small\sffamily] {No} (bselect.west);

\node[jevcard] (jselect) at (4,-2.42)
    {\textbf{\normalsize JEV}\\[-1pt]\small Selects one tool or \texttt{respond}};
\node[card] (jllm) at (4,-3.77)
    {\textbf{\normalsize Base LLM}\\[-1pt]\small Fills arguments or writes the reply};
\node[card] (jenv) at (4,-5.12)
    {\textbf{\normalsize Environment}\\[-1pt]\small Execute action; append result or reply};
\node[decision] (jcheck) at (4,-6.42) {Done or\\30-turn cap?};
\node[card] (jeval) at (4,-7.75)
    {\textbf{\normalsize Evaluation}\\[-1pt]\small Log reward, turns, and time};
\draw[flow] (jhead.south) -- (jselect.north);
\draw[flow] (jselect) -- (jllm);
\draw[flow] (jllm) -- (jenv);
\draw[flow] (jenv) -- (jcheck);
\draw[flow] (jcheck) -- node[right,font=\small\sffamily] {Yes} (jeval);
\draw[flow] (jcheck.east) -- ++(2.65,0) |-
    node[pos=0.24,right,font=\small\sffamily] {No} (jselect.east);

\node[shared] (aggregate) at (0,-9.05)
    {\textbf{Shared comparison} \quad Pass$^1$ \qquad P50 \qquad P95};
\draw[flow] (beval.south) -- (beval.south |- aggregate.north);
\draw[flow] (jeval.south) -- (jeval.south |- aggregate.north);
\end{tikzpicture}
}
\caption{Matched $\tau$-bench workflow for the baseline and JEV arms.}
\label{fig:tau-bench-agentic-workflow}
\end{figure*}

To isolate the effect of JEV on agent decision making, we replace the action-selection component of the original $\tau$-bench agent while leaving the remaining execution pipeline unchanged. In the original agent, the base LLM jointly determines the next action from the dialogue context and tool descriptions and generates the corresponding tool arguments or user-facing response. In the JEV variant, JEV first selects the next action from the currently available tools together with a \texttt{respond} option. The same base LLM then generates arguments for the selected tool, or produces the response text when \texttt{respond} is chosen. In both systems, the resulting action is executed by the same $\tau$-bench environment, which then updates the interaction state. Figure~\ref{fig:tau-bench-agentic-workflow} illustrates the two workflows.

\subsection{ElectionSim Results}
\label{app:electionsim-states}

Table~\ref{tab:electionsim} summarizes the simulation of Section~\ref{sec:electionsim}. Following ElectionSim, States and Battleground count the correctly called states among all 51 and among the 15 battleground states. MAE, RMSE, and Bias measure the error of the predicted Democratic two-party share across states; the MAE equals the per-state RMSE that ElectionSim reports, averaged over states, and a positive bias overestimates Harris.

\begin{table}[!htb]
    \centering
    \caption{Summary of the 2024 U.S. presidential election simulated in ElectionSim with voters sampled at a ratio of 1/1,000. Errors are in percentage points. Bold marks the better value in each column, with the bias closer to zero counted as better.}
    \label{tab:electionsim}
    \small
    \setlength{\tabcolsep}{5pt}
    \renewcommand{\arraystretch}{1.1}
    \begin{tabular}{@{}lccrrr@{}}
    \toprule
    \textbf{Voter model} & \textbf{States} $\uparrow$ & \textbf{Battleground} $\uparrow$ & \textbf{MAE} $\downarrow$ & \textbf{RMSE} $\downarrow$ & \textbf{Bias} \\
    \midrule
    ElectionSim (GPT-4o-mini) & 46/51 & 11/15 & \textbf{3.06} & \textbf{3.80} & \textbf{+2.31} \\
    \ourlarge{} (decision model) & \textbf{47/51} & \textbf{12/15} & 6.52 & 9.42 & +3.73 \\
    \bottomrule
    \end{tabular}
\end{table}

Table~\ref{tab:electionsim-states} lists the per-state results behind Table~\ref{tab:electionsim} and Figure~\ref{fig:electionsim-map}. Shares are Harris's percentage of the two-party vote, and errors are predicted minus actual shares in percentage points. Voters counts the simulated voters of \ourlarge{}; the archived run of the original pipeline covers 331,836 voters with nearly the same per-state counts. Following ElectionSim, the 15 battleground states are Arizona, Colorado, Florida, Georgia, Iowa, Michigan, Minnesota, Nevada, New Hampshire, North Carolina, Ohio, Pennsylvania, Texas, Virginia, and Wisconsin.

\begingroup
\footnotesize
\setlength{\tabcolsep}{5pt}
\renewcommand{\arraystretch}{1.02}
\setlength{\LTcapwidth}{\textwidth}
\begin{longtable}{@{}lcrrrr@{\hspace{12pt}}rr@{}}
\caption{Per-state results of the 2024 ElectionSim simulation with a 1/1,000 voter sample. BG marks the 15 battleground states, and $^\dagger$ marks a state called incorrectly.}
\label{tab:electionsim-states}\\
\toprule
& & & & \multicolumn{2}{c}{\textbf{\ourlarge{}}} & \multicolumn{2}{c}{\textbf{ElectionSim, GPT-4o-mini}} \\
\cmidrule(lr){5-6}\cmidrule(l){7-8}
\textbf{State} & \textbf{BG} & \textbf{Voters} & \textbf{Actual} & \textbf{Share} & \textbf{Error} & \textbf{Share} & \textbf{Error} \\
\midrule
\endfirsthead
\multicolumn{8}{@{}l}{\textit{Table \thetable{} continued}}\\
\toprule
& & & & \multicolumn{2}{c}{\textbf{\ourlarge{}}} & \multicolumn{2}{c}{\textbf{ElectionSim, GPT-4o-mini}} \\
\cmidrule(lr){5-6}\cmidrule(l){7-8}
\textbf{State} & \textbf{BG} & \textbf{Voters} & \textbf{Actual} & \textbf{Share} & \textbf{Error} & \textbf{Share} & \textbf{Error} \\
\midrule
\endhead
\midrule\multicolumn{8}{r@{}}{\footnotesize Continued on next page}\\
\endfoot
\bottomrule
\endlastfoot
Alabama &  & 5,031 & 34.5 & 32.6 & $-$1.9 & 34.4 & $-$0.2 \\
Alaska &  & 737 & 43.2 & 12.6 & $-$30.5 & 38.0 & $-$5.2 \\
Arizona & $\bullet$ & 7,159 & 47.2 & 41.4 & $-$5.8 & 46.6 & $-$0.6 \\
Arkansas &  & 3,014 & 34.3 & 35.5 & +1.2 & 33.6 & $-$0.7 \\
California &  & 39,577 & 60.6 & 66.2 & +5.6 & 64.1 & +3.4 \\
Colorado & $\bullet$ & 5,783 & 55.6 & 56.4 & +0.8 & 58.0 & +2.4 \\
Connecticut &  & 3,609 & 57.4 & 62.4 & +5.0 & 61.6 & +4.2 \\
Delaware &  & 991 & 57.5 & 71.7 & +14.3 & 64.3 & +6.8 \\
District of Columbia &  & 690 & 93.3 & 87.6 & $-$5.7 & 92.9 & $-$0.4 \\
Florida & $\bullet$ & 21,571 & 43.4 & 43.0 & $-$0.3 & 46.1 & +2.7 \\
Georgia & $\bullet$ & 10,726 & 48.9 & 45.5 & $-$3.4 & 48.4 & $-$0.5 \\
Hawaii &  & 1,461 & 61.8 & 71.7 & +9.9 & 64.1 & +2.3 \\
Idaho &  & 1,842 & 31.2 & 38.5 & +7.2 & 36.0 & +4.8 \\
Illinois &  & 12,823 & 55.4 & 59.1 & +3.7 & 57.9 & +2.5 \\
Indiana &  & 6,791 & 40.3 & 47.0 & +6.7 & 44.3 & +4.0 \\
Iowa & $\bullet$ & 3,193 & 43.3 & 47.7 & +4.4 & 47.9 & +4.6 \\
Kansas &  & 2,941 & 41.8 & 41.7 & 0.0 & 41.1 & $-$0.6 \\
Kentucky &  & 4,510 & 34.5 & 42.8 & +8.4 & 36.7 & +2.2 \\
Louisiana &  & 4,662 & 38.8 & 38.4 & $-$0.5 & 38.5 & $-$0.3 \\
Maine &  & 1,364 & 53.6 & 55.3 & +1.7 & 58.6 & +5.0 \\
Maryland &  & 6,186 & 64.4 & 65.4 & +1.0 & 64.9 & +0.5 \\
Massachusetts &  & 7,034 & 62.6 & 74.7 & +12.1 & 70.6 & +8.0 \\
Michigan & $\bullet$ & 10,085 & 49.2 & 54.7\rlap{$^\dagger$} & +5.5 & 53.4\rlap{$^\dagger$} & +4.1 \\
Minnesota & $\bullet$ & 5,710 & 52.1 & 56.8 & +4.6 & 54.5 & +2.4 \\
Mississippi &  & 2,964 & 37.8 & 38.0 & +0.1 & 40.0 & +2.2 \\
Missouri &  & 6,161 & 40.7 & 47.2 & +6.6 & 43.2 & +2.6 \\
Montana &  & 1,086 & 39.7 & 43.7 & +4.0 & 41.5 & +1.7 \\
Nebraska &  & 1,964 & 39.6 & 34.3 & $-$5.3 & 35.7 & $-$3.9 \\
Nevada & $\bullet$ & 3,109 & 48.4 & 61.5\rlap{$^\dagger$} & +13.1 & 53.3\rlap{$^\dagger$} & +4.9 \\
New Hampshire & $\bullet$ & 1,380 & 51.4 & 68.6 & +17.1 & 60.4 & +9.0 \\
New Jersey &  & 9,295 & 53.0 & 60.6 & +7.5 & 60.0 & +6.9 \\
New Mexico &  & 2,121 & 53.1 & 53.4 & +0.2 & 49.6\rlap{$^\dagger$} & $-$3.6 \\
New York &  & 20,216 & 55.9 & 64.1 & +8.2 & 61.9 & +6.0 \\
North Carolina & $\bullet$ & 10,454 & 48.3 & 48.5 & +0.2 & 49.0 & +0.7 \\
North Dakota &  & 780 & 31.3 & 21.6 & $-$9.6 & 30.1 & $-$1.2 \\
Ohio & $\bullet$ & 11,809 & 44.3 & 43.4 & $-$0.9 & 47.9 & +3.6 \\
Oklahoma &  & 3,964 & 32.5 & 31.3 & $-$1.2 & 33.9 & +1.3 \\
Oregon &  & 4,242 & 57.4 & 67.5 & +10.0 & 60.1 & +2.7 \\
Pennsylvania & $\bullet$ & 13,012 & 49.1 & 53.4\rlap{$^\dagger$} & +4.3 & 50.8\rlap{$^\dagger$} & +1.7 \\
Rhode Island &  & 1,099 & 56.9 & 60.0 & +3.0 & 58.7 & +1.8 \\
South Carolina &  & 5,125 & 41.0 & 42.8 & +1.9 & 43.5 & +2.6 \\
South Dakota &  & 888 & 35.0 & 48.5 & +13.4 & 35.0 & 0.0 \\
Tennessee &  & 6,917 & 34.9 & 36.5 & +1.6 & 37.4 & +2.4 \\
Texas & $\bullet$ & 29,184 & 43.0 & 40.2 & $-$2.8 & 43.9 & +0.9 \\
Utah &  & 3,276 & 38.9 & 37.4 & $-$1.5 & 36.9 & $-$2.0 \\
Vermont &  & 644 & 66.4 & 80.2 & +13.8 & 68.9 & +2.6 \\
Virginia & $\bullet$ & 8,655 & 52.9 & 61.5 & +8.6 & 56.1 & +3.2 \\
Washington &  & 7,716 & 59.5 & 68.2 & +8.7 & 62.8 & +3.2 \\
West Virginia &  & 1,796 & 28.6 & 41.7 & +13.0 & 36.1 & +7.5 \\
Wisconsin & $\bullet$ & 5,898 & 49.6 & 48.2 & $-$1.4 & 51.0\rlap{$^\dagger$} & +1.4 \\
Wyoming &  & 578 & 26.5 & 60.3\rlap{$^\dagger$} & +33.9 & 34.6 & +8.1 \\
\end{longtable}
\endgroup

\section{Details of \ourmethod{}}
\label{app:self-distillation}

This appendix gives the input template used by the first-token readout of Section~\ref{sec:self-distillation}, the data points of Figure~\ref{fig:self-distillation-frontier}, and the effect of the training set size on \ourmethod{}.

\subsection{Readout Template}
\label{app:self-distillation-template}

All open models in Section~\ref{sec:self-distillation} share the plain-text template below; no few-shot examples are used. The readout is taken at the token that follows the final answer prefix, over the labels of the valid options only. When the teacher reasons, the same prompt is used with thinking enabled, and the answer prefix is appended after its reasoning segment. \texttt{Noul} questions present the two outcomes as options (A) and (B); \texttt{score} questions present every rating level as an option.

\begin{promptbox}{Readout template}
\fontsize{9}{10.5}\selectfont\raggedright
{\ttfamily State:\par
\{state\}\par
Question: \{instruction\}\par
Options:\par
(A) \{option\_1\_key\}: \{option\_1\_description\}\par
(B) \{option\_2\_key\}: \{option\_2\_description\}\par
\ldots\par
The correct answer is (}
\end{promptbox}

\begin{promptbox}{Example: \texttt{noul} question}
\fontsize{9}{10.5}\selectfont\raggedright
{\ttfamily State:\par
Order \#1182. Status: shipped on 3 September. Carrier: DHL.\par
Question: Has the order been shipped? Answer strictly from the facts stated in the text.\par
Options:\par
(A) yes: The text states that this is so\par
(B) no: The text states that this is not so\par
The correct answer is (}\par\medskip
The readout returns the probability of (A) as $P(\texttt{true})$.
\end{promptbox}

\begin{promptbox}{Example: \texttt{score} question}
\fontsize{9}{10.5}\selectfont\raggedright
{\ttfamily State:\par
The icon is misaligned. Every function works.\par
Question: Rate incident impact using only reported facts. Use the highest fully supported level.\par
Options:\par
(A) 0: No function impaired; cosmetic only\par
(B) 1: One user or a nonessential function impaired, with a workaround\par
(C) 2: Many users blocked from a core function, no data loss\par
(D) 3: Confirmed irreversible data loss or physical harm\par
The correct answer is (}\par\medskip
The readout returns the expected level $0\cdot p(\mathrm{A}) + 1\cdot p(\mathrm{B}) + 2\cdot p(\mathrm{C}) + 3\cdot p(\mathrm{D})$.
\end{promptbox}

\subsection{Training Details}
\label{app:self-distillation-training}

Table~\ref{tab:self-distillation-hparams} lists the training settings of \oursmall{} and \ourlarge{}. Both scales use the same settings, and the full-rank and LoRA versions differ only in which weights are updated and in the learning rate.

\begin{table}[htbp]
\centering
\caption{Training settings for \ourmethod{}, shared by both backbones.}
\label{tab:self-distillation-hparams}
\small
\setlength{\tabcolsep}{8pt}
\renewcommand{\arraystretch}{1.08}
\begin{tabular}{@{}lll@{}}
\toprule
\textbf{Setting} & \textbf{Full-rank (main)} & \textbf{LoRA} \\
\midrule
Updated weights & attention and MLP projections & LoRA adapters on the same modules \\
LoRA rank and $\alpha$ & \textemdash & 64 and 128 \\
Learning rate & $2\times10^{-6}$ & $5\times10^{-5}$ \\
Frozen weights & \multicolumn{2}{l}{embeddings and output layer} \\
Optimizer and schedule & \multicolumn{2}{l}{AdamW, 3\% warmup, cosine decay} \\
Global batch size and epochs & \multicolumn{2}{l}{64, one epoch} \\
Precision & \multicolumn{2}{l}{BF16} \\
Teacher chains per question & \multicolumn{2}{l}{2} \\
Loss weights & \multicolumn{2}{l}{$\lambda=0.5$; $\kappa=0.05$, or $0.2$ for hard-labeled \texttt{noul}} \\
Checkpoint selection & \multicolumn{2}{l}{frozen development set} \\
\bottomrule
\end{tabular}
\end{table}

\subsection{Accuracy and Latency of Decision Models}
\label{app:self-distillation-frontier}

This section describes how the data of Figure~\ref{fig:self-distillation-frontier} are measured. Accuracy is the pooled JEVal accuracy of Section~\ref{sec:main-results}. Latency is the mean latency for inputs of up to 1K tokens, measured for every open model on a single H100 80\,GB GPU; Jev's value is that of its hosted API and includes network transport. On the same GPU, we also time our models with one request at a time over a sample of 232 JEVal questions, each timed three times, including tokenization, the forward pass, and the readout. The median latency is 58.7\,ms for \oursmall{} and 112.2\,ms for \ourlarge{}. Intern-Decision-0.8B and Intern-Decision-2B were not part of the latency test and are therefore not plotted.


\subsection{Effect of Training Set Size}
\label{app:self-distillation-scaling}

\begin{figure}[t]
  \centering
  \includegraphics[width=0.92\textwidth]{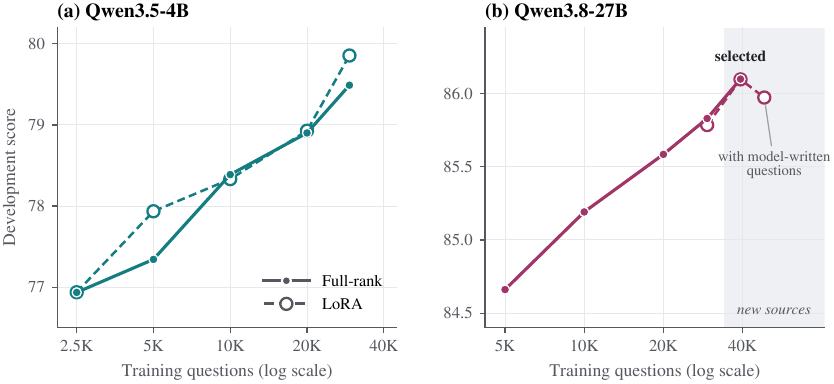}
  \caption{Development score against the number of training questions under \ourmethod{}. Filled markers with solid lines show full-rank training, and open markers with dashed lines show LoRA. Points for Qwen3.5-4B are means over two seeds. For Qwen3.8-27B, sizes up to 20,000 questions use one seed and larger sizes average two or three seeds; the shaded region marks sets that add new sources, and the largest set also contains questions written by the model itself.}
  \label{fig:data-size-curve}
\end{figure}

We measure how development performance grows with the number of training questions under \ourmethod{}, where each backbone learns from its own post-reasoning distributions. The training sets from 5,000 to 29,326 questions are nested samples of one pool, drawn with a fixed seed and stratified by capability group, so every size keeps the same mixture, and Qwen3.5-4B is also trained on a smaller set of 2,500 questions. The pool cannot grow further with this mixture, so for Qwen3.8-27B we add questions from new sources: science from QASC and ScienceQA, quantitative and logical reasoning from MathQA, AQuA, LogiQA2, and ReClor, medical and legal practice, and factual and conditional boundaries from VitaminC, ConditionalQA, and ContractNLI. This gives a set of 39,279 questions and a larger set of 48,358 questions, which also contains science questions that Qwen3.8-27B wrote itself from paper abstracts and encyclopedia articles. Both sets keep all 29,326 earlier questions but are drawn independently of each other. As before, targets for the added questions average two reasoning chains of the model and never use gold labels, and questions whose chains are truncated or invalid or disagree on the top option are left out. Table~\ref{tab:data-mixture} gives the composition of every set. Every run selects its checkpoint on a frozen development set of 7,506 questions, whose score weights nine capability groups equally and sources equally within each group, using accuracy for classification questions and one minus the normalized MAE for soft \texttt{noul} and \texttt{score} questions.

\makeatletter\setlength{\@fptop}{0pt}\makeatother
\begin{table}[!h]
\centering
\caption{Number of training questions per capability group. The sets of 5,000 to 29,326 questions are nested and keep the same mixture. The sets of 39,279 and 48,358 questions keep all 29,326 questions, add new sources, and are drawn independently of each other; \ourlarge{} is trained on the set of 39,279 questions.}
\label{tab:data-mixture}
\small
\setlength{\tabcolsep}{5.5pt}
\renewcommand{\arraystretch}{1.08}
\begin{tabular}{@{}lrrrrrr@{}}
\toprule
\textbf{Capability group} & \textbf{5,000} & \textbf{10,000} & \textbf{20,000} & \textbf{29,326} & \textbf{39,279} & \textbf{48,358} \\
\midrule
Language understanding & 1,199 & 2,398 & 4,795 & 7,031 & 7,031 & 7,031 \\
General knowledge & 938 & 1,875 & 3,751 & 5,500 & 5,500 & 5,500 \\
Commonsense & 693 & 1,387 & 2,774 & 4,068 & 4,068 & 4,068 \\
Logic & 520 & 1,040 & 2,081 & 3,051 & 3,051 & 3,051 \\
Medicine, English & 520 & 1,040 & 2,080 & 3,050 & 3,050 & 3,050 \\
Medicine, Chinese & 498 & 996 & 1,992 & 2,921 & 2,921 & 2,921 \\
Decisions, rules, and computation & 473 & 946 & 1,891 & 2,773 & 2,773 & 2,773 \\
Safety and subjective judgment & 159 & 318 & 636 & 932 & 932 & 932 \\
\midrule
\multicolumn{7}{@{}l}{\emph{New sources}} \\
Science & \textemdash & \textemdash & \textemdash & \textemdash & 3,279 & 6,014 \\
\quad of which written by the model & \textemdash & \textemdash & \textemdash & \textemdash & \textemdash & 2,731 \\
Quantitative and logical reasoning & \textemdash & \textemdash & \textemdash & \textemdash & 3,000 & 6,925 \\
Medical and legal practice & \textemdash & \textemdash & \textemdash & \textemdash & 2,500 & 4,384 \\
Factual and conditional boundaries & \textemdash & \textemdash & \textemdash & \textemdash & 1,174 & 1,709 \\
\midrule
Total & 5,000 & 10,000 & 20,000 & 29,326 & 39,279 & 48,358 \\
\bottomrule
\end{tabular}
\end{table}

Figure~\ref{fig:data-size-curve} shows that development performance rises steadily with the amount of training data at both scales. The growth is roughly linear in the logarithm of the training set size, so each doubling of the data buys a similar gain, and the gain per doubling is smaller for Qwen3.8-27B, whose scores are already higher. Full-rank and LoRA training follow nearly the same curve. For Qwen3.8-27B, the new sources carry the trend to 39,279 questions, but the larger set brings no further gain, so we train \ourlarge{} on 39,279 questions.

We attribute the stall mainly to the quality of the added questions rather than to their number. \ourmethod{} uses neither a stronger model nor gold labels, and this limits how new training questions can be made. First, questions that a model writes itself tend to fall within its comfort zone. The science questions that Qwen3.8-27B wrote turned out to be easy for it: it answered them correctly without thinking, so its teacher and student already agree and the questions carry little training signal. A useful question must be hard enough that the student fails without reasoning. Second, the model cannot reliably check whether such a hard question is itself correct, and a flawed question turns directly into a wrong target. In earlier experiments, questions generated from programmatic templates failed in yet another way: the student learned the template rather than the skill, and the gain did not transfer to real questions. Scaling \ourmethod{} further therefore depends on finding hard, well-posed questions rather than on adding more questions.

These curves describe a trend rather than a law. The development set also selects checkpoints, so absolute scores are optimistic. The smaller 27B sizes use a single seed, the expanded sets also change the mixture, and the runs on 48,358 questions stopped at a preset time limit about halfway through their epoch, so they show only that the larger set brings no early gain. The curves therefore support the overall trend but not the gain between any two neighboring sizes, and they do not support fitting a scaling law.

\end{document}